\PassOptionsToPackage{table}{xcolor}
\documentclass{lightspeed}

\definecolor{titlebg}{HTML}{EEF3FA}

\renewcommand{\title}[1]{\newcommand{\titlelist}{{\LARGE\sffamily\bfseries #1}}}
\renewcommand\authorformat[2][]{{\sffamily\bfseries\color{metabg}#2$^{#1}$}}
\renewcommand\affiliationformat[2][]{{\small\color{metabg}$^{#1}$#2}}
\renewcommand\contributionformat[2][]{{\small\color{metabg}$^{#1}$#2}}
\renewcommand\emailformat[2][]{{\footnotesize\texttt{\href{mailto:#2}{#2}}}}
\renewcommand\metadataformat[2][]{{\small{\sffamily\bfseries #1:} #2}}

\renewcommand{\mymaketitle}{%
  \begingroup
  \setlength{\parindent}{0pt}
  \setlength{\parskip}{0pt}
  \vspace*{-0.35cm}
  \begin{tcolorbox}[enhanced,colback=titlebg,colframe=titlebg,
      boxrule=0pt,arc=9pt,left=0.55cm,right=0.55cm,
      top=0.45cm,bottom=0.40cm,before skip=0pt,after skip=0.45cm]
    {\raggedright\nohyphens
      {\setstretch{1.08}\titlelist\par}
      \vspace{0.28cm}
      {\authorlist\par}
      \vspace{0.14cm}
      {\affiliationlist\par}
      \ifdefempty{\contributionlist}{}{\vspace{0.06cm}\contributionlist\par}
      \ifdefempty{\emaillist}{}{\vspace{0.10cm}{\color{metabg}\emaillist\par}}
      \vspace{0.32cm}
      {\small\setstretch{1.08}\color{metafg}\abstractlist\par}
      \ifdefempty{\headercontentlist}{}{\vspace{0.26cm}\headercontentlist}
      \ifdefempty{\metadatalist}{}{\metadatalist\par}
    }
  \end{tcolorbox}
  \endgroup
  \FloatBarrier
}

\microtypesetup{expansion=false}
\setcitestyle{square,comma,numbers,sort&compress}

\usepackage[T1]{fontenc}
\usepackage[utf8]{inputenc}
\usepackage{microtype}
\usepackage{inconsolata}

\usepackage{graphicx}
\usepackage{wrapfig}
\usepackage{needspace}
\usepackage{placeins}
\usepackage{flafter}
\usepackage{booktabs}
\usepackage{multirow}
\usepackage{amsmath}
\usepackage{amssymb}
\usepackage{url}
\usepackage{enumitem}
\usepackage{xspace}
\usepackage{listings}

\usepackage{mathtools}
\usepackage{algorithm}
\usepackage[noend]{algpseudocode}
\usepackage{pgfplots}
\usepackage{pgfplotstable}
\usepackage{fontawesome5} %
\pgfplotsset{compat=1.18}
\usepgfplotslibrary{fillbetween,groupplots}
\usetikzlibrary{calc}

\creflabelformat{equation}{#2#1#3}
\apptocmd{\beginappendix}{\crefalias{section}{appendix}\crefalias{subsection}{subappendix}}{}{}



\newlength{\colw}
\newlength{\wraptopsep}
\newlength{\wrapbotsep}
\makeatletter
\newdimen\wrap@dp      %
\newdimen\wrap@start   %
\newdimen\wrap@bottom  %
\AfterEndEnvironment{wrapfigure}{\global\wrap@dp=\dimexpr\WF@size-1.5\baselineskip-2\intextsep\relax}
\patchcmd{\WF@putfigmaybe}{\lower\intextsep}{\lower\wraptopsep}{}{\PackageError{main}{wrapfig patch failed}{}}
\def\wrap@setlines#1{\begingroup
  \dimen@=\dimexpr\wrap@bottom+\wrapbotsep+.7\baselineskip-(#1)\relax
  \ifdim\pagetotal<\wrap@start \dimen@\z@ \fi %
  \ifdim\dimen@>\z@
    \advance\dimen@\baselineskip \advance\dimen@-1sp
    \@tempcnta\dimen@ \divide\@tempcnta\baselineskip \advance\@tempcnta\@ne
  \else \@tempcnta\z@ \fi
  \global\c@WF@wrappedlines\@tempcnta \endgroup}
\def\WF@startwrapping{%
 \ifx\WF@wli\@empty
  \global\wrap@start\pagetotal
  \global\wrap@bottom=\dimexpr\pagetotal+\baselineskip+\wraptopsep+\wrap@dp\relax
  \wrap@setlines{\pagetotal+\baselineskip}%
 \else
  \setcounter{WF@wrappedlines}{\WF@wli}\global\advance\c@WF@wrappedlines\@ne
 \fi
 \ifnum\c@WF@wrappedlines>\@ne
  \let\parshape\WF@fudgeparshape \let\WF@pspars\@empty \let\WF@@par\par
  \def\@setpar##1{\def\WF@@par{##1}}\def\par{\@par}\let\@par\WF@mypar
  \xdef\WF@restoretol{\tolerance\the\tolerance}\tolerance9999
  \advance\linewidth-\WF@adjlw \advance\@totalleftmargin\WF@adjtlm
 \fi}
\def\WF@mypar{\relax
 \WF@@par
 \ifnum\@@parshape=\z@ \let\WF@pspars\@empty \fi
 \ifx\WF@wli\@empty
  \ifinner\else \wrap@setlines{\pagetotal+\parskip+\baselineskip}\fi
 \else
  \global\advance\c@WF@wrappedlines-\prevgraf
 \fi
 \prevgraf\z@
 \ifnum\c@WF@wrappedlines<\tw@ \WF@finale \fi}
\newcommand{\wrapneed}[1]{\par\penalty-100 \begingroup
  \setlength{\dimen@}{#1}%
  \ifdim\pagegoal=\maxdimen %
    \ifdim\dimen@>\vsize \hbox{}\break \fi
  \else
    \dimen@ii\pagegoal \advance\dimen@ii-\pagetotal
    \ifdim\dimen@>\dimen@ii \ifdim\dimen@ii>\z@ \vfil \fi \break \fi
  \fi \endgroup}
\newcommand{\wrapend}{\par
  \ifnum\c@WF@wrappedlines>\@ne \ifdim\pagetotal<\wrap@start\else
    \begingroup \dimen@=\dimexpr\wrap@bottom+\wrapbotsep-\pagetotal-\pagedepth\relax
    \ifdim\dimen@>\z@ \vskip\dimen@ \fi \endgroup
  \fi\fi \WFclear}
\makeatother

\definecolor{promptHead}{HTML}{34495E}
\definecolor{promptVar}{HTML}{1A2FA0}
\newtcolorbox{promptbox}[1]{enhanced, breakable, colback=black!3, colframe=promptHead,
  coltitle=white, fonttitle=\bfseries\small, title={#1}, arc=2pt, boxrule=0.6pt,
  left=4pt, right=4pt, top=3pt, bottom=3pt, fontupper=\small}
\newcommand{\promptvar}[1]{\textcolor{promptVar}{\texttt{\{#1\}}}}

\definecolor{stuV}{HTML}{1F4FE0}
\definecolor{lineO}{HTML}{EE6A1F}
\definecolor{tchV}{HTML}{E09A00}
\definecolor{dltR}{HTML}{B22222}
\definecolor{ablT}{HTML}{1A9E77}
\definecolor{simR}{HTML}{9B1B30} %
\definecolor{dltG}{HTML}{2E7D32}
\newcommand{\dlt}[1]{{\scriptsize\textcolor{dltR}{$-$#1}}}
\newcommand{\dltp}[1]{{\scriptsize\textcolor{dltG}{+#1}}}
\newcommand{\vsrow}[1]{\hspace{5pt}{\scriptsize\textcolor{black!55}{\textit{vs.\ #1}}}}
\newcommand{\dmk}[1]{\rlap{$^{#1}$}}
\makeatletter
\newcommand{\refrows}[1]{\noalign{%
  \vskip-\belowrulesep
  \begingroup\color{black!10}\hrule height\dimexpr#1\ht\@arstrutbox+#1\dp\@arstrutbox+\belowrulesep+\aboverulesep\relax\endgroup
  \vskip-\dimexpr#1\ht\@arstrutbox+#1\dp\@arstrutbox+\aboverulesep\relax}}
\makeatother

\colorlet{takeawaybg}{titlebg}
\newtcolorbox{takeawaybox}{
  enhanced,
  breakable,
  colback=takeawaybg,
  colframe=takeawaybg,
  boxrule=0pt,
  arc=0.8mm,
  left=1.2mm,
  right=1.2mm,
  top=0.9mm,
  bottom=0.9mm
}

\definecolor{decS}{HTML}{A6D5CB}
\definecolor{decM}{HTML}{C6E3DC}
\definecolor{decL}{HTML}{E2F0EC}
\definecolor{incL}{HTML}{F7E0CE}
\definecolor{incM}{HTML}{EDBB94}

\title{RoDeL: Role-Decoupled Learning for Context Distillation}

\author[1,*]{Ruiyi Yan}
\author[2]{Zhuoyuan Mao}
\author[1,\dagger]{Yugo Murawaki}
\author[3,\dagger]{Yiwen Guo}

\affiliation[1]{Kyoto University}
\affiliation[2]{LIGHTSPEED}
\affiliation[3]{Independent Researcher}
\renewcommand{\affiliationlist}{%
  \affiliationformat[1]{Kyoto University}\quad
  \affiliationformat[2]{LIGHTSPEED}\quad
  \affiliationformat[3]{Independent Researcher}%
}

\email{ruiyi@nlp.ist.i.kyoto-u.ac.jp}
\email{zhuoyuanmao@global.tencent.com}
\email{murawaki@i.kyoto-u.ac.jp}
\email{guoyiwen89@gmail.com}

\abstract{Context distillation converts the token memory of large language models (LLMs), such as a document in the context, into parametric memory stored in model parameters. It involves two roles: a teacher, which has the document in its context and answers synthetic questions about it, and a student, which answers new questions by reading the memory distilled from these answers. The two roles depend on each other, but existing methods train at most one of them, or both through shared parameters. We propose RoDeL ({\bfseries Ro}le-{\bfseries De}coupled {\bfseries L}earning), which trains the teacher and the student as two separate adapters on a frozen base model so that the two roles co-evolve. RoDeL has two variants: RoDeL-Seq updates the two adapters sequentially, alternating between the roles, and RoDeL-Sim updates them simultaneously, in the same training step.
To train the teacher, we introduce \emph{student-relayed Group Relative Policy Optimization} (SR-GRPO): each of several answer sets synthesized for a document is written into its own memory adapter and rewarded by how well the student answers after reading it. With Qwen3.5-4B and only 270 training documents, RoDeL-Seq or RoDeL-Sim performs best on each of two in-domain and eight out-of-domain datasets. On average, RoDeL-Seq leads the best baseline by 3.4 points in domain and 3.3 out of domain (relative improvements of 9.8\% and 7.4\%), and RoDeL-Sim leads it by 9.6 and 3.5 points (27.7\% and 7.9\%).
}
\date{November 2026\hfill{\sffamily\bfseries Code:} \faGithub\ \url{https://github.com/ryehr/RoDeL}}

\begin{document}
\thispagestyle{firstheader}
\maketitle
\begingroup
\renewcommand{\thefootnote}{\fnsymbol{footnote}}
\footnotetext[1]{This work was done during internship at LIGHTSPEED.}
\footnotetext[2]{Corresponding authors.}
\endgroup
\setcounter{footnote}{0}

\wrapneed{10.2cm} %
\section{Introduction}
\label{sec:intro}

\begin{wrapfigure}{r}{\colw}
\centering
\includegraphics[width=\colw]{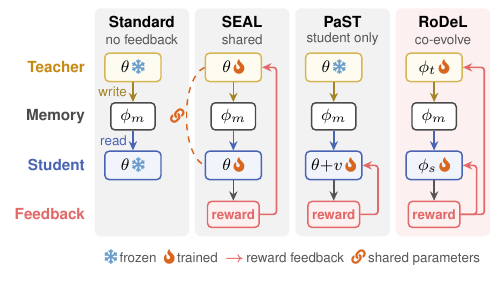}
\caption{Which roles of context distillation are trained. The teacher writes information into the memory $\phi_m$, the student answers questions by reading it, and the reward reflects how well the student answers. Standard context distillation uses the base model $\theta$ in both roles and learns no feedback. SEAL trains $\theta$, which both roles share, and PaST trains a skill vector $v$ for the student. RoDeL trains a teacher adapter $\phi_t$ and a student adapter $\phi_s$ on the frozen $\theta$, both from the student's reward, so that the two roles co-evolve.}
\label{fig:teaser}
\end{wrapfigure}

Knowledge, experiences, and preferences can be incorporated into large language models (LLMs) in two forms: \emph{token memory}, kept as tokens in the context or in a retrievable external store, and \emph{parametric memory}, stored in model parameters through training~\citep{hu2026memoryageaiagents}. Token memory occupies the context and needs prefilling at every use, which becomes costly for long-context scenarios. Token-to-parameter memory converts the former into the latter: new information is written once into model parameters, which later questions read in place of the context~\citep{su2025parametric,caccia2025training}. A common way to write such memory is context distillation~\citep{snell2022learning}, which involves two roles: a \emph{teacher}, which has the information in its context and synthesizes training data from it (e.g., answers to synthetic questions), and a \emph{student}, which answers questions by reading the memory distilled from these data. In standard context distillation, the untrained base model plays both roles, so how memories are written and read does not adapt to how well the student answers~\citep{caccia2025training,ICLR2026_4681359a}.

Recent work trains one of the roles with reinforcement learning, learning across documents how memories are written or read. For the teacher, SEAL~\citep{zweiger2025self} trains a model to generate its own fine-tuning data, rewarded by how well the model answers questions after fine-tuning on these data. For the student, PaST~\citep{tang-etal-2026-knowledge} adds a skill vector that improves how a model uses newly injected knowledge. However, the other role is either fixed, as in PaST, or shares parameters with the trained one, as in SEAL.

The two roles depend on each other: how effectively the teacher writes depends on how the student reads the memory, and how effectively the student reads depends on what the teacher writes. This raises the question of whether the two roles can co-evolve as separate learners.
Inspired by this, we propose RoDeL (Role-Decoupled Learning), which trains the teacher and the student of context distillation as two separate adapters on a frozen base model, as illustrated in \Cref{fig:teaser}. The student is trained with reinforcement learning to answer questions by reading memory adapters written by the teacher. The teacher is rewarded by how well the student answers these questions after the teacher's content has been distilled into a memory adapter. RoDeL has two variants, which differ in when the two adapters are updated. RoDeL-Seq updates them sequentially: training alternates between student updates, in which the teacher is frozen, and teacher updates, in which the student is frozen. RoDeL-Sim updates them simultaneously: every training step updates both adapters.

Our contributions are as follows. First, we propose RoDeL for context distillation, which decouples the two roles into a teacher adapter and a student adapter and trains them with reinforcement learning, sequentially in RoDeL-Seq and simultaneously in RoDeL-Sim, so that each role evolves and adapts to the current version of the other. Second, we introduce \emph{student-relayed GRPO} (SR-GRPO), which trains the teacher with Group Relative Policy Optimization (GRPO)~\citep{shao2024deepseekmathpushinglimitsmathematical} on rewards relayed by the student. For each document, the teacher synthesizes several sets of content, each written into its own memory adapter. The student's answers after reading each adapter give the reward of the corresponding set, and SR-GRPO compares these rewards across the sets of the same document. Third, with Qwen3.5-4B~\citep{qwen35} and only 270 training documents, RoDeL-Seq or RoDeL-Sim achieves the best performance on each of two in-domain and eight out-of-domain datasets. The average performance of RoDeL-Seq exceeds that of the untrained roles by 14.6 points in domain and 17.5 out of domain, and that of SEAL-style and PaST-style baselines, which each train one role, by 4.8 and 3.4 points in domain and by 3.3 and 6.2 out of domain. The average performance of RoDeL-Sim exceeds that of the untrained roles by 20.8 points in domain and 17.7 out of domain, and that of the two baselines by 11.0 and 9.6 points in domain and by 3.5 and 6.4 out of domain (\Cref{subsec:baselines}). Besides, in long-context retrieval scenarios with corpora of up to 1.37M tokens, the best performance on every dataset still belongs to one of the two variants (\Cref{subsec:retrieval}). With Gemma-4-E2B~\citep{gemmateam2026gemma4technicalreport}, a base model from another model family, the average out-of-domain performance of both variants again exceeds that of every baseline, and RoDeL-Seq also has the best average in-domain performance. Finally, with Qwen3.5-2B and Qwen3.5-0.8B, two smaller models of the Qwen3.5 family, the average performance of both variants significantly exceeds that of the untrained roles in domain and out of domain (\Cref{subsec:generality}).

\section{Preliminaries}
\label{sec:preliminaries}

\subsection{Token-to-Parameter Memory}
\label{subsec:setup}

A language model can hold the content of a document $D$ (e.g., a passage or a set of passages) in two forms: as tokens in its context, or in its parameters~\citep{hu2026memoryageaiagents}. Tokens occupy the context every time $D$ is used, whereas parameters persist and add no context at use time. Token-to-parameter memory converts the former into the latter. Before any question about $D$ is asked, $D$ is \emph{written} offline into a document-specific LoRA~\citep{hu2022lora} on a frozen base language model $\theta$, as in parametric RAG~\citep{su2025parametric} and knowledge modules~\citep{caccia2025training}. This adapter, termed the \emph{memory adapter} $\phi_m(D)$, is the parametric memory of $D$: each later question about $D$ is answered by \emph{reading} $\phi_m(D)$, with $D$ absent from the context. Each document has its own memory adapter. 
This paper studies writing through context distillation (\Cref{subsec:context_distill}). Let $\pi_{\theta\oplus\phi}$ denote the base model with adapter $\phi$ attached: each adapted weight matrix $W$ becomes $W+\Delta W_\phi$, where $\Delta W_\phi$ is the low-rank increment of $\phi$.

\paragraph{Prompts and responses.}
Let $x(q)$ be a prompt asking for an answer to $q$, and $x_T(D,q)$ the same prompt with $D$ prepended. A response $y$ consists of a reasoning trace~\citep{wei2022chain} followed by an answer $\mathrm{ans}(y)$.

\subsection{Writing and Reading via Context Distillation}
\label{subsec:context_distill}

Context distillation~\citep{snell2022learning} carries out both operations with two roles: a \emph{teacher}, which has $D$ in its context and synthesizes training data for the memory adapter, and a \emph{student}, which answers questions by reading the memory adapter. In its standard form, the base model plays both roles (self-distillation). Writing $D$ takes two steps: synthesis, in which the teacher generates the training data, and distillation, in which $\phi_m(D)$ is trained on them.

\paragraph{Synthesis.}
Synthesis can be prompted in various ways, e.g., by rewriting $D$ or by generating conversations about it~\citep{su2025parametric,ICLR2026_4681359a}. Here, the teacher is prompted with \emph{synthetic questions} $\tilde Q_D=\{\tilde q_1,\dots,\tilde q_n\}$, generated offline from $D$ alone by a question-generation model. With $D$ in its context, it answers each of them,
\begin{equation}
\label{eq:synthesis}
y_j\sim\pi_{\theta}\big(\cdot\mid x_T(D,\tilde q_j)\big),\quad j=1,\dots,n,
\end{equation}
yielding an \emph{answer set} $S=\{(\tilde q_j,y_j)\}_{j=1}^{n}$, which pairs each synthetic question with the teacher's response.

\paragraph{Distillation.}
A new memory adapter $\phi_m$, initialized to a zero increment and attached to the frozen $\theta$, is trained to imitate the teacher on its responses with $D$ removed from the prompt. At position $l$ of response $y_j$, let
\begin{equation*}
\begin{aligned}
p^{T}_{j,l}&=\pi_{\theta}\big(\cdot\mid x_T(D,\tilde q_j),\,y_{j,<l}\big),\\
p^{\phi_m}_{j,l}&=\pi_{\theta\oplus\phi_m}\big(\cdot\mid x(\tilde q_j),\,y_{j,<l}\big)
\end{aligned}
\end{equation*}
be the next-token distributions of the teacher, which sees $D$, and of the model with $\phi_m$ attached, which does not. The distillation loss averages a token-level loss $\ell_{j,l}$ over all $N=\sum_{j}|y_j|$ response tokens:
\begin{equation}
\label{eq:distill_loss}
\mathcal L_{\mathrm{KD}}(\phi_m;S)=\frac{1}{N}\sum_{j=1}^{n}\sum_{l=1}^{|y_j|}\ell_{j,l}.
\end{equation}
Common choices of $\ell_{j,l}$ include the cross-entropy on the teacher's sampled token, $-\log p^{\phi_m}_{j,l}(y_{j,l})$, as in sequence-level distillation~\citep{kim2016sequence}, and the forward KL divergence to the teacher's full distribution, $\mathrm{KL}\big(p^{T}_{j,l}\,\|\,p^{\phi_m}_{j,l}\big)$ \citep{hinton2015distilling,kim2016sequence}. The trained adapter is denoted by $\mathrm{Distill}(S)$, leaving its dependence on the teacher's distributions implicit, and $\phi_m(D)=\mathrm{Distill}(S)$.

\paragraph{Reading.}
To answer a question $q$ about $D$, the student, i.e., the base model with $\phi_m(D)$ attached, sees only $x(q)$, samples $\hat y\sim\pi_{\theta\oplus\phi_m(D)}(\cdot\mid x(q))$, and returns $\mathrm{ans}(\hat y)$ as its answer.

\subsection{Related Work}
\label{subsec:related}

Methods that move context into parameters typically train each memory either on the context itself~\citep{wang2024with} or on content derived from it by a prompted model: the model's own outputs given the context~\citep{snell2022learning}, rewrites and question--answer pairs~\citep{su2025parametric}, text relating extracted entities~\citep{ICLR2025_6dcf277e}, synthetic conversations~\citep{ICLR2026_4681359a}, or knowledge elicited and selected from the context~\citep{cao-etal-2025-infiniteicl}. Other methods additionally learn, across documents, how memories are written or read. SEAL~\citep{zweiger2025self} trains a model with reinforcement learning to generate its own fine-tuning data, rewarding the data by the performance of the fine-tuned model. PaST~\citep{tang-etal-2026-knowledge} adds a skill vector, obtained from reinforcement learning on a source domain, to a model fine-tuned on new data, improving how it uses the new knowledge. Besides, \citet{zheng2026contextdistillationlatentmemory} manage a bank of per-context adapters, retrieving and routing each query to a suitable one.

Viewed through the two roles in \Cref{subsec:context_distill}, methods that only train each memory leave both the teacher, which writes, and the student, which reads, untrained. SEAL and PaST each train only one: SEAL trains the teacher, whose updates also change the student that shares its parameters, and PaST trains the student but not the teacher. These limitations motivate training both roles with independent parameter modules, each for its own role.

\section{RoDeL}
\label{sec:method}

\begin{figure}[!htb]
\begin{minipage}[t]{\colw}
\centering
\includegraphics[width=\colw]{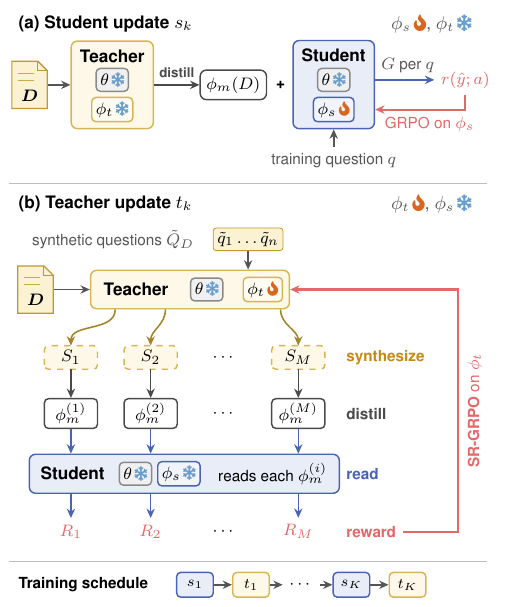}
\caption{Overview of RoDeL-Seq. (a)~Student update $s_k$: the frozen teacher writes each training document $D$ into a memory adapter $\phi_m(D)$, and $\phi_s$ is trained with GRPO on the rewards $r(\hat y;a)$ of the student's responses to training questions. (b)~Teacher update $t_k$: the teacher synthesizes $M$ answer sets $S_1,\dots,S_M$ for $D$, each distilled into its own memory adapter $\phi_m^{(i)}$ and read by the frozen student; the set-level rewards $R_i$ (\Cref{eq:set_reward}) train $\phi_t$ with SR-GRPO.}
\label{fig:overview}
\end{minipage}\hfill
\begin{minipage}[t]{\colw}
\centering
\includegraphics[width=\colw]{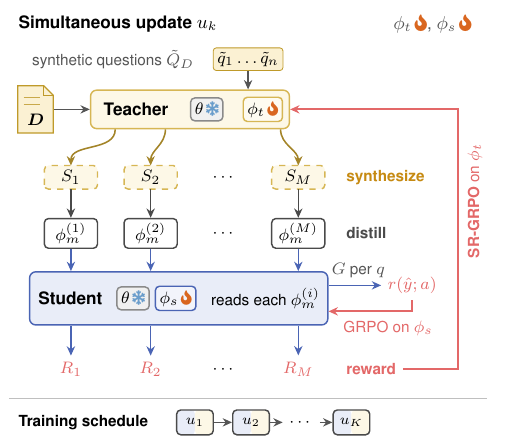}
\caption{Overview of RoDeL-Sim. In a simultaneous update $u_k$, the teacher synthesizes $M$ answer sets $S_1,\dots,S_M$ for $D$, each distilled into its own memory adapter $\phi_m^{(i)}$ and read by the student; the student's sampled responses train $\phi_s$ with GRPO, and the set-level rewards $R_i$ (\Cref{eq:set_reward}) of its greedy responses train $\phi_t$ with SR-GRPO.}
\label{fig:sim}
\end{minipage}
\end{figure}

RoDeL (Role-Decoupled Learning) trains both roles of context distillation. On the frozen base model $\theta$, the teacher and the student each have their own LoRA, the \emph{teacher adapter} $\phi_t$ and the \emph{student adapter} $\phi_s$, and both are trained with reinforcement learning from how well questions are answered by reading the memory. Each role has its own objective, detailed in \Cref{subsec:student,subsec:teacher}. RoDeL has two variants, which differ in when the two adapters are updated (\Cref{subsec:variants}). RoDeL-Seq updates them sequentially, alternating between student updates and teacher updates (\Cref{fig:overview}). RoDeL-Sim updates them simultaneously, training both adapters in the same step (\Cref{fig:sim}). RoDeL-Seq lets the roles be updated at different times or places (e.g., a teacher that writes offline and a student that reads online), whereas RoDeL-Sim trains both in one loop.

\subsection{Data and Adapters}
\label{subsec:adapters}

The training data of RoDeL train the role adapters $\phi_t$ and $\phi_s$. They consist of triples $(D,q,a)\in\mathcal T$, each formed by a document $D$, a \emph{training question} $q$ about $D$, and its gold answer $a$. A document may appear in several triples with different question--answer pairs; let $\mathcal D$ be the set of training documents and $\mathcal P_D=\{(q,a):(D,q,a)\in\mathcal T\}$ the question--answer pairs of $D$. Training questions are posed only to the student, and their gold answers serve to compute rewards. Synthetic questions, in contrast, serve to write memory: each memory adapter $\phi_m(D)$ is distilled from the teacher's answers to $\tilde Q_D$ (\Cref{subsec:context_distill}).  The teacher is prompted with $D$ and synthetic questions.

The teacher is $\theta\oplus\phi_t$: with $D$ in its context, it synthesizes $y\sim\pi_{\theta\oplus\phi_t}(\cdot\mid x_T(D,\tilde q))$ for each synthetic question $\tilde q\in\tilde Q_D$. The memory adapter is distilled on the frozen $\theta$ alone, without $\phi_t$ or $\phi_s$ attached. The student is $\theta\oplus\phi_m(D)\oplus\phi_s$, where attaching two adapters adds both increments, $W+\Delta W_{\phi_m(D)}+\Delta W_{\phi_s}$.

\looseness=-1 A response $y$ is \emph{valid}, $\mathrm{valid}(y)=1$, if it finishes its reasoning trace and gives a non-empty answer before generation is truncated. $\mathrm{Synth}(\phi_t,D)$ denotes an answer set for $D$ synthesized by the teacher, and $\mathrm{Distill}(S)$ applies \Cref{eq:distill_loss} with the token-level cross-entropy to the valid responses $S^+=\{(\tilde q_j,y_j)\in S:\mathrm{valid}(y_j)=1\}$.

\subsection{Student Objective}
\label{subsec:student}

The student adapter $\phi_s$ learns to answer training questions by reading memory adapters. Training passes over the training documents $\mathcal D$ in mini-batches $\mathcal B\subset\mathcal D$. For each document $D\in\mathcal B$, the student reads a memory adapter $\phi_m(D)$ written by the teacher; when and how these memory adapters are written depends on the variant (\Cref{subsec:variants}). For each $(q,a)\in\mathcal P_D$, the student samples $G$ responses without $D$ in its context, $\hat y_{q,g}\sim\pi_{\theta\oplus\phi_m(D)\oplus\phi_s}(\cdot\mid x(q))$. A student response $\hat y$ to a training question with gold answer $a$ receives the reward
\begin{equation}
\label{eq:reward}
r(\hat y;a)=\begin{cases}\mathrm{F1}(\mathrm{ans}(\hat y),a) & \text{if valid},\\ -c & \text{otherwise},\end{cases}
\end{equation}
where $\mathrm{F1}$ is the token-level F1 score and $c \geq 0$ is a fixed penalty. With $r_{q,g}=r(\hat y_{q,g};a)$, the sampled responses receive group-normalized advantages
\begin{equation}
\label{eq:student_adv}
\hat A_{q,g}=\frac{r_{q,g}-\operatorname{mean}_{g'}r_{q,g'}}{\operatorname{std}_{g'}r_{q,g'}+\epsilon}.
\end{equation}
Here $\epsilon$ is a small constant for numerical stability. Following GRPO~\citep{shao2024deepseekmathpushinglimitsmathematical}, $\phi_s$ minimizes
\begin{equation}
\label{eq:student_loss}
\mathcal L_s=-\frac{1}{|\mathcal B|}\sum_{D\in\mathcal B}\frac{1}{|\mathcal P_D|\,G}\sum_{(q,a)\in\mathcal P_D}\sum_{g=1}^{G}\frac{\hat A_{q,g}}{|\hat y_{q,g}|}\log\pi_{\theta\oplus\phi_m(D)\oplus\phi_s}\big(\hat y_{q,g}\mid x(q)\big),
\end{equation}
where $\log\pi(y\mid x)=\sum_l\log\pi(y_l\mid x,y_{<l})$ and gradients flow only into $\phi_s$.

\subsection{Teacher Objective}
\label{subsec:teacher}

The teacher adapter $\phi_t$ is trained with \emph{student-relayed GRPO} (SR-GRPO). For each document $D\in\mathcal B$, the teacher independently synthesizes $M$ answer sets $S_i=\mathrm{Synth}(\phi_t,D)=\{(\tilde q_j,y_{i,j})\}_{j=1}^{n}$, $i=1,\dots,M$, and each is distilled into its own memory adapter $\phi_m^{(i)}=\mathrm{Distill}(S_i)$. Reading $\phi_m^{(i)}$, the student answers every training question of $D$ with greedy decoding, $\hat y_{i,q}=\operatorname{greedy}\pi_{\theta\oplus\phi_m^{(i)}\oplus\phi_s}(\cdot\mid x(q))$. The reward of $S_i$ is the student's mean reward minus a penalty on invalid responses,
\begin{equation}
\label{eq:set_reward}
R_i=\frac{1}{|\mathcal P_D|}\sum_{(q,a)\in\mathcal P_D}r(\hat y_{i,q};a)-\lambda b_i,
\end{equation}
where $b_i$ is the fraction of the teacher's responses in $S_i$ that are invalid, and $\lambda>0$ weights this penalty.

The reward is defined after a whole answer set has been written and read. The $M$ answer sets of a document therefore form a group, and each receives the set-level advantage
\begin{equation}
\label{eq:set_adv}
A_i=\frac{R_i-\bar R}{\sigma_R+\epsilon},
\end{equation}
where $\bar R$ and $\sigma_R$ are the mean and standard deviation of $R_1,\dots,R_M$. Every response in $S_i$ is weighted by $A_i$:
\begin{equation}
\label{eq:teacher_loss}
\mathcal L_t=-\frac{1}{|\mathcal B|}\sum_{D\in\mathcal B}\frac{1}{M}\sum_{i=1}^{M}\frac{A_i}{n}\sum_{j=1}^{n}\frac{1}{|y_{i,j}|}\log\pi_{\theta\oplus\phi_t}\big(y_{i,j}\mid x_T(D,\tilde q_j)\big),
\end{equation}
where gradients flow into $\phi_t$ alone. The signal reaches the teacher only as relayed by the student. Its reward thus measures how well it \textit{teaches} the student: how well the student answers after the teacher's own content has been written into memory.

\subsection{Two Variants: RoDeL-Seq and RoDeL-Sim}
\label{subsec:variants}

Both variants start $\phi_t$ and $\phi_s$ from zero increments and train them for $K$ rounds. They differ in how a round schedules the two objectives.

\paragraph{RoDeL-Seq.} Round $k$ performs a student update $s_k$ followed by a teacher update $t_k$:
\begin{equation*}
s_1\to t_1\to s_2\to t_2\to\cdots\to s_K\to t_K.
\end{equation*}
A student update trains $\phi_s$ on $\mathcal L_s$ with the teacher frozen. It first writes a memory adapter for every training document with the current teacher, $\phi_m(D)=\mathrm{Distill}(\mathrm{Synth}(\phi_t,D))$, exactly as at inference. These adapters stay fixed during the update, which passes once over $\mathcal D$ and takes one gradient step per batch. A teacher update trains $\phi_t$ on $\mathcal L_t$ with the student frozen, again passing once over $\mathcal D$ with one gradient step per batch. Each update trains its own adapter, continuing from that adapter's previous update. The memory adapters for $s_k$ are written by the teacher after $t_{k-1}$, and the student frozen in $t_k$ is the one after $s_k$. Each update adapts one role to the latest version of the other. Alternating the two updates lets the roles adapt to each other and improve together over rounds. \Cref{fig:overview} illustrates the two updates, and \Cref{alg:rodel} summarizes training.

\paragraph{RoDeL-Sim.} Round $k$ performs one simultaneous update $u_k$, which passes once over $\mathcal D$ and trains both adapters in every batch. For each document $D\in\mathcal B$, the teacher synthesizes $M$ answer sets, each distilled into its own memory adapter $\phi_m^{(i)}$, as in \Cref{subsec:teacher}. These memory adapters serve both objectives. Reading each $\phi_m^{(i)}$, the student samples $G$ responses to every training question of $D$, $\hat y^{(i)}_{q,g}\sim\pi_{\theta\oplus\phi_m^{(i)}\oplus\phi_s}(\cdot\mid x(q))$. The $G$ responses that share a question and a memory adapter form a group and receive the advantages of \Cref{eq:student_adv}, and $\mathcal L_s$ (\Cref{eq:student_loss}) is averaged over the $M$ memory adapters of each document. With the same memory adapters, the student's greedy responses give the set-level rewards $R_i$ (\Cref{eq:set_reward}) and thus $\mathcal L_t$ (\Cref{eq:teacher_loss}). Each batch takes one gradient step on $\phi_s$ and one on $\phi_t$, and both adapters continue from the previous round. Each batch thus adapts both roles to the current version of each other. \Cref{fig:sim} illustrates one simultaneous update, and \Cref{alg:sim} summarizes training.

In both variants, each adapter has made $k$ passes over $\mathcal D$ after round $k$.

\begin{figure}[!htb]
\algrenewcommand\algorithmicindent{1.2em}
\begin{minipage}[t]{0.485\textwidth}
\vspace{0pt}
\begin{algorithm}[H]
\caption{RoDeL-Seq training.}
\label{alg:rodel}
\small
\begin{algorithmic}[1]
\Require frozen $\theta$, training documents $\mathcal D$, rounds $K$
\State $\phi_t\gets0$, $\phi_s\gets0$
\For{$k=1,\dots,K$}
  \State $\phi_m(D)\gets\mathrm{Distill}(\mathrm{Synth}(\phi_t,D))$, $\forall D\in\mathcal D$
  \ForAll{batches $\mathcal B\subset\mathcal D$} \Comment{student update $s_k$}
    \State sample $\hat y_{q,1:G}$ for all $(q,a)\in\mathcal P_D$, $D\in\mathcal B$
    \State compute $\hat A_{q,g}$ (\Cref{eq:student_adv})
    \State update $\phi_s$ on $\mathcal L_s$ (\Cref{eq:student_loss})
  \EndFor
  \ForAll{batches $\mathcal B\subset\mathcal D$} \Comment{teacher update $t_k$}
    \ForAll{$D\in\mathcal B$, $i=1,\dots,M$}
      \State $S_i\gets\mathrm{Synth}(\phi_t,D)$
      \State $\phi_m^{(i)}\gets\mathrm{Distill}(S_i)$
      \State greedy student gives $R_i$ (\Cref{eq:set_reward})
    \EndFor
    \State compute $A_i$ (\Cref{eq:set_adv})
    \State update $\phi_t$ on $\mathcal L_t$ (\Cref{eq:teacher_loss})
  \EndFor
\EndFor
\State \Return $\phi_t,\phi_s$
\end{algorithmic}
\end{algorithm}
\end{minipage}\hfill
\begin{minipage}[t]{0.485\textwidth}
\vspace{0pt}
\begin{algorithm}[H]
\caption{RoDeL-Sim training.}
\label{alg:sim}
\small
\begin{algorithmic}[1]
\Require frozen $\theta$, training documents $\mathcal D$, rounds $K$
\State $\phi_t\gets0$, $\phi_s\gets0$
\For{$k=1,\dots,K$}
  \ForAll{batches $\mathcal B\subset\mathcal D$} \Comment{update $u_k$}
    \ForAll{$D\in\mathcal B$, $i=1,\dots,M$}
      \State $S_i\gets\mathrm{Synth}(\phi_t,D)$
      \State $\phi_m^{(i)}\gets\mathrm{Distill}(S_i)$
      \State sample $\hat y^{(i)}_{q,1:G}$ for all $(q,a)\in\mathcal P_D$
      \State greedy student gives $R_i$ (\Cref{eq:set_reward})
    \EndFor
    \State compute $\hat A^{(i)}_{q,g}$ (\Cref{eq:student_adv})
    \State update $\phi_s$ on $\mathcal L_s$ (\Cref{eq:student_loss})
    \State compute $A_i$ (\Cref{eq:set_adv})
    \State update $\phi_t$ on $\mathcal L_t$ (\Cref{eq:teacher_loss})
  \EndFor
\EndFor
\State \Return $\phi_t,\phi_s$
\end{algorithmic}
\end{algorithm}
\end{minipage}
\end{figure}

\subsection{Inference}
\label{subsec:inference}

Inference is the same for both variants. Given a new document $D_{\text{new}}$ and its synthetic questions, generated in the same way as for training documents, the final teacher writes its memory adapter once, $\phi_m(D_{\text{new}})=\mathrm{Distill}(\mathrm{Synth}(\phi_t,D_{\text{new}}))$. Each test question $q$ about $D_{\text{new}}$ is then answered by the student $\theta\oplus\phi_m(D_{\text{new}})\oplus\phi_s$ from $x(q)$, which returns $\mathrm{ans}(\hat y)$.

\section{Experiments}
\label{sec:experiments}

\subsection{Experimental Setup}
\label{subsec:exp_setup}

\paragraph{Datasets.}
Training documents come from the training splits of HotpotQA~\citep{yang-etal-2018-hotpotqa} and MuSiQue~\citep{trivedi-etal-2022-musique}. Each document consists of the passages provided with one question of these datasets, so $|\mathcal P_D|=1$. The training set $\mathcal D$ contains 180 HotpotQA and 90 MuSiQue documents. Synthetic questions are generated by Qwen3-4B~\citep{yang2025qwen3technicalreport} from each document, averaging 37.7 per HotpotQA and 48.2 per MuSiQue training document.

Evaluation covers two in-domain and eight out-of-domain datasets. In domain, HotpotQA and MuSiQue each contribute 200 test questions with their passages. Out of domain, documents vary in genre and length: paragraphs from SQuAD~\citep{rajpurkar-etal-2016-squad}, Wikipedia passages from 2WikiMultihopQA (2Wiki)~\citep{ho-etal-2020-constructing}, Wikipedia pages from Natural Questions (NQ)~\citep{kwiatkowski-etal-2019-natural} and TriviaQA~\citep{joshi-etal-2017-triviaqa}, NLP papers from Qasper~\citep{dasigi-etal-2021-dataset}, multi-domain documents from MultiFieldQA in LongBench~\citep{bai-etal-2024-longbench}, books and movie scripts from NarrativeQA~\citep{kocisky-etal-2018-narrativeqa}, and multi-session conversations from LoCoMo~\citep{maharana-etal-2024-evaluating}, with median lengths from 161 to 24k tokens. Each test question is answered by reading the memory adapter written for its document. A response is scored by the token-level F1 of its answer, maximized over gold-answer aliases. \Cref{app:data} details the construction and statistics of all datasets.

\paragraph{Baselines.}
\textit{PRAG-style} represents the methods in \Cref{subsec:related} that train each memory but leave both roles untrained~\citep{wang2024with,snell2022learning,ICLR2025_6dcf277e,ICLR2026_4681359a,cao-etal-2025-infiniteicl}: as in parametric RAG~\citep{su2025parametric}, the base model writes each memory adapter and reads it. It is also the starting point of RoDeL. \textit{SEAL-style} follows SEAL~\citep{zweiger2025self} in training one adapter shared by both roles. \textit{PaST-style} follows PaST~\citep{tang-etal-2026-knowledge} in training the student. The untrained base model writes the memory adapters, as in PRAG-style, and $\phi_s$ is trained by student updates (\Cref{subsec:student}). It runs for as many steps as the student updates of RoDeL-Seq. 
\Cref{app:baselines} lists how these baselines differ from the original methods.

\paragraph{Implementation details.}
In the main experiments, the frozen base model $\theta$ is Qwen3.5-4B~\citep{qwen35} in its thinking mode. Generation is truncated at 1024 tokens. The teacher and student adapters are rank-64 LoRAs and each memory adapter is a rank-16 LoRA, all applied to every linear layer of $\theta$. Both variants of RoDeL run $K=3$ rounds with $M=G=4$ and $c=\lambda=2$. $\mathrm{Distill}$ trains each memory adapter for five epochs. Both role adapters are trained with AdamW~\citep{loshchilov2019decoupled} at a learning rate of $10^{-5}$. Responses used for reinforcement learning are sampled at temperature 1.0. \Cref{app:impl} lists detailed hyperparameters and prompts.

\subsection{RoDeL across Training Updates}
\label{subsec:dynamics}

\begin{figure}[!htb]
\begin{minipage}[t]{\colw}
\centering
\pgfplotstableread{
x  id    idsd  ood   oodsd
0  23.4  1.82  30.1  0.81
1  24.5  1.92  32.0  0.84
2  26.5  1.99  33.2  0.82
3  32.9  2.06  40.6  0.80
4  37.3  2.06  47.6  0.82
5  38.0  2.11  46.4  0.81
6  36.9  2.05  44.9  0.81
}\dyntab
\begin{tikzpicture}
\begin{axis}[
  width=\colw, height=5.2cm,
  xmin=-0.5, xmax=6.5, ymin=14, ymax=58,
  xtick={0,...,6},
  xticklabels={Frozen,\textcolor{stuV}{$s_1$},\textcolor{tchV}{$t_1$},\textcolor{stuV}{$s_2$},\textcolor{tchV}{$t_2$},\textcolor{stuV}{$s_3$},\textcolor{tchV}{$t_3$}},
  ytick={20,30,40,50},
  ylabel={Average F1},
  axis lines=left, axis line style={-, black!60}, tick style={black!60},
  tick label style={font=\small}, label style={font=\small},
  ymajorgrids, grid style={black!22, dashed},
  set layers, clip=false,
  legend style={font=\small, at={(0.5,1.03)}, anchor=south, draw=none, fill=none, legend columns=2, /tikz/every even column/.append style={column sep=10pt}},
  legend cell align=left,
]
\begin{pgfonlayer}{axis background}
\draw[black!25] (axis cs:0.5,14) -- (axis cs:0.5,58);
\draw[black!25] (axis cs:2.5,14) -- (axis cs:2.5,58);
\draw[black!25] (axis cs:4.5,14) -- (axis cs:4.5,58);
\end{pgfonlayer}
\node[font=\scriptsize, text=black!55, anchor=north] at (axis cs:1.5,58) {Round 1};
\node[font=\scriptsize, text=black!55, anchor=north] at (axis cs:3.5,58) {Round 2};
\node[font=\scriptsize, text=black!55, anchor=north] at (axis cs:5.5,58) {Round 3};
\addplot[name path=idU, draw=none, forget plot] table[x=x, y expr=\thisrow{id}+\thisrow{idsd}] {\dyntab};
\addplot[name path=idL, draw=none, forget plot] table[x=x, y expr=\thisrow{id}-\thisrow{idsd}] {\dyntab};
\addplot[lineO, fill opacity=0.15, draw=none, forget plot] fill between[of=idU and idL];
\addplot[name path=oodU, draw=none, forget plot] table[x=x, y expr=\thisrow{ood}+\thisrow{oodsd}] {\dyntab};
\addplot[name path=oodL, draw=none, forget plot] table[x=x, y expr=\thisrow{ood}-\thisrow{oodsd}] {\dyntab};
\addplot[lineO, fill opacity=0.15, draw=none, forget plot] fill between[of=oodU and oodL];
\addplot[lineO, line width=1.2pt, mark=*, mark size=2.3pt, mark options={fill=lineO, draw=white, line width=0.6pt},
  nodes near coords, point meta=explicit symbolic, every node near coord/.append style={font=\scriptsize, text=black!70, anchor=north, yshift=-4pt}]
  table[x=x, y=id, meta=id] {\dyntab};
\addlegendentry{In-domain}
\addplot[lineO, line width=1.2pt, mark=square*, mark size=2.1pt, mark options={fill=lineO, draw=white, line width=0.6pt},
  nodes near coords, point meta=explicit symbolic, every node near coord/.append style={font=\scriptsize, text=black!70, anchor=south, yshift=2pt}]
  table[x=x, y=ood, meta=ood] {\dyntab};
\addlegendentry{Out-of-domain}
\end{axis}
\end{tikzpicture}
\caption{Average F1 of RoDeL-Seq with Qwen3.5-4B after each update, from the frozen roles (PRAG-style) to the end of training. Blue and gold labels mark student and teacher updates. Shaded bands show $\pm 1$ standard deviation of a test-set bootstrap (\Cref{app:results}).}
\label{fig:dynamics}
\end{minipage}\hfill
\begin{minipage}[t]{\colw}
\centering
\pgfplotstableread{
x  id    idsd  ood   oodsd
0  23.4  1.82  30.1  0.81
1  25.2  1.88  36.7  0.81
2  40.1  2.19  47.8  0.78
3  44.2  2.11  47.2  0.77
}\simtab
\begin{tikzpicture}
\begin{axis}[
  width=\colw, height=5.2cm,
  xmin=-0.5, xmax=3.5, ymin=14, ymax=58,
  xtick={0,...,3},
  xticklabels={Frozen,$u_1$,$u_2$,$u_3$},
  ytick={20,30,40,50},
  ylabel={Average F1},
  axis lines=left, axis line style={-, black!60}, tick style={black!60},
  tick label style={font=\small}, label style={font=\small},
  ymajorgrids, grid style={black!22, dashed},
  clip=false,
  legend style={font=\small, at={(0.5,1.03)}, anchor=south, draw=none, fill=none, legend columns=2, /tikz/every even column/.append style={column sep=10pt}},
  legend cell align=left,
]
\addplot[name path=idU, draw=none, forget plot] table[x=x, y expr=\thisrow{id}+\thisrow{idsd}] {\simtab};
\addplot[name path=idL, draw=none, forget plot] table[x=x, y expr=\thisrow{id}-\thisrow{idsd}] {\simtab};
\addplot[lineO, fill opacity=0.15, draw=none, forget plot] fill between[of=idU and idL];
\addplot[name path=oodU, draw=none, forget plot] table[x=x, y expr=\thisrow{ood}+\thisrow{oodsd}] {\simtab};
\addplot[name path=oodL, draw=none, forget plot] table[x=x, y expr=\thisrow{ood}-\thisrow{oodsd}] {\simtab};
\addplot[lineO, fill opacity=0.15, draw=none, forget plot] fill between[of=oodU and oodL];
\addplot[lineO, line width=1.2pt, mark=*, mark size=2.3pt, mark options={fill=lineO, draw=white, line width=0.6pt}]
  table[x=x, y=id] {\simtab};
\addlegendentry{In-domain}
\addplot[lineO, line width=1.2pt, mark=square*, mark size=2.1pt, mark options={fill=lineO, draw=white, line width=0.6pt}]
  table[x=x, y=ood] {\simtab};
\addlegendentry{Out-of-domain}
\node[font=\scriptsize, text=black!70, anchor=north, yshift=-4pt] at (axis cs:0,23.4) {23.4};
\node[font=\scriptsize, text=black!70, anchor=north, yshift=-4pt] at (axis cs:1,25.2) {25.2};
\node[font=\scriptsize, text=black!70, anchor=north west, yshift=-2pt, xshift=1pt] at (axis cs:2,40.1) {40.1};
\node[font=\scriptsize, text=black!70, anchor=north, yshift=-4pt] at (axis cs:3,44.2) {44.2};
\node[font=\scriptsize, text=black!70, anchor=south east, yshift=1pt, xshift=1pt] at (axis cs:0,30.1) {30.1};
\node[font=\scriptsize, text=black!70, anchor=south east, yshift=1pt, xshift=-1pt] at (axis cs:1,36.7) {36.7};
\node[font=\scriptsize, text=black!70, anchor=south, yshift=2pt] at (axis cs:2,47.8) {47.8};
\node[font=\scriptsize, text=black!70, anchor=south, yshift=2pt] at (axis cs:3,47.2) {47.2};
\end{axis}
\end{tikzpicture}
\caption{Average F1 of RoDeL-Sim with Qwen3.5-4B after each simultaneous update. Shaded bands are as in \Cref{fig:dynamics}.}
\label{fig:avgsim}
\end{minipage}
\end{figure}

\Cref{fig:dynamics,fig:avgsim} follow RoDeL-Seq and RoDeL-Sim from the frozen roles to the end of training on average, and \Cref{fig:perset4b,fig:persetsim} on each dataset. Each variant is evaluated after every update, with memory adapters written by the current teacher and read by the current student, as at inference. Before any update, both roles are the base model (the first point is PRAG-style). \Cref{app:results} gives the significance tests.

With RoDeL-Seq, training both roles raises the average F1 from 23.4 to 36.9 in domain and from 30.1 to 44.9 out of domain. After $t_3$, RoDeL-Seq scores higher than PRAG-style on all ten datasets. The largest gains are on LoCoMo ($+27.6$), Qasper ($+21.2$), and MuSiQue ($+17.6$).

With RoDeL-Sim, the average F1 rises from 23.4 to 44.2 in domain and from 30.1 to 47.2 out of domain. After $u_3$, RoDeL-Sim scores significantly higher than PRAG-style on both averages and, after Holm correction, on all ten datasets. The largest gains are again on LoCoMo ($+28.4$), Qasper ($+23.3$), and MuSiQue ($+23.0$).

\begin{figure}[!htb]
\centering
\begin{tikzpicture}
\begin{groupplot}[
  group style={group size=5 by 2, horizontal sep=0.55cm, vertical sep=0.95cm},
  scale only axis, width=2.5cm, height=2.0cm,
  xmin=-0.5, xmax=6.5,
  xtick={0,...,6},
  xticklabels={Fr.,\textcolor{stuV}{$s_1$},\textcolor{tchV}{$t_1$},\textcolor{stuV}{$s_2$},\textcolor{tchV}{$t_2$},\textcolor{stuV}{$s_3$},\textcolor{tchV}{$t_3$}},
  axis lines=left, axis line style={-, black!60}, tick style={black!60}, major tick length=2pt,
  tick label style={font=\tiny, inner sep=1pt},
  title style={font=\footnotesize, yshift=-4pt},
  ymajorgrids, grid style={black!22, dashed},
  extra x ticks={0.5,2.5,4.5}, extra x tick labels={},
  extra x tick style={grid=major, major grid style={black!25, solid}, tick style={draw=none}},
  clip=false,
]
\nextgroupplot[title={HotpotQA (ID)}, ymin=30, ymax=60, ylabel={F1}, ylabel style={font=\scriptsize, yshift=-3pt}]
\addplot[name path=U, draw=none, forget plot] table[x=x, y expr=\thisrow{m}+\thisrow{sd}] {
x  m  sd
0  39.5  3.26
1  37.9  3.27
2  38.5  3.27
3  42.7  3.19
4  52.4  3.22
5  48.3  3.22
6  48.9  3.22
};
\addplot[name path=L, draw=none, forget plot] table[x=x, y expr=\thisrow{m}-\thisrow{sd}] {
x  m  sd
0  39.5  3.26
1  37.9  3.27
2  38.5  3.27
3  42.7  3.19
4  52.4  3.22
5  48.3  3.22
6  48.9  3.22
};
\addplot[lineO, fill opacity=0.15, draw=none, forget plot] fill between[of=U and L];
\addplot[lineO, line width=0.9pt, mark=*, mark size=1.4pt, mark options={fill=lineO, draw=white, line width=0.35pt}] table[x=x, y=m] {
x  m  sd
0  39.5  3.26
1  37.9  3.27
2  38.5  3.27
3  42.7  3.19
4  52.4  3.22
5  48.3  3.22
6  48.9  3.22
};
\node[font=\tiny, text=black!70, anchor=south, yshift=1.6pt, inner sep=0.5pt] at (axis cs:0,39.5) {39.5};
\node[font=\tiny, text=black!70, anchor=north, yshift=-1.6pt, inner sep=0.5pt] at (axis cs:1,37.9) {37.9};
\node[font=\tiny, text=black!70, anchor=south, yshift=1.6pt, inner sep=0.5pt] at (axis cs:2,38.5) {38.5};
\node[font=\tiny, text=black!70, anchor=south, yshift=1.6pt, inner sep=0.5pt] at (axis cs:3,42.7) {42.7};
\node[font=\tiny, text=black!70, anchor=south, yshift=1.6pt, inner sep=0.5pt] at (axis cs:4,52.4) {52.4};
\node[font=\tiny, text=black!70, anchor=north, yshift=-1.6pt, inner sep=0.5pt] at (axis cs:5,48.3) {48.3};
\node[font=\tiny, text=black!70, anchor=south, yshift=1.6pt, inner sep=0.5pt] at (axis cs:6,48.9) {48.9};
\nextgroupplot[title={MuSiQue (ID)}, ymin=0, ymax=40]
\addplot[name path=U, draw=none, forget plot] table[x=x, y expr=\thisrow{m}+\thisrow{sd}] {
x  m  sd
0  7.3  1.67
1  11.1  2.02
2  14.4  2.31
3  23.1  2.68
4  22.2  2.56
5  27.7  2.77
6  24.9  2.54
};
\addplot[name path=L, draw=none, forget plot] table[x=x, y expr=\thisrow{m}-\thisrow{sd}] {
x  m  sd
0  7.3  1.67
1  11.1  2.02
2  14.4  2.31
3  23.1  2.68
4  22.2  2.56
5  27.7  2.77
6  24.9  2.54
};
\addplot[lineO, fill opacity=0.15, draw=none, forget plot] fill between[of=U and L];
\addplot[lineO, line width=0.9pt, mark=*, mark size=1.4pt, mark options={fill=lineO, draw=white, line width=0.35pt}] table[x=x, y=m] {
x  m  sd
0  7.3  1.67
1  11.1  2.02
2  14.4  2.31
3  23.1  2.68
4  22.2  2.56
5  27.7  2.77
6  24.9  2.54
};
\node[font=\tiny, text=black!70, anchor=north, yshift=-1.6pt, inner sep=0.5pt] at (axis cs:0,7.3) {7.3};
\node[font=\tiny, text=black!70, anchor=south, yshift=1.6pt, inner sep=0.5pt] at (axis cs:1,11.1) {11.1};
\node[font=\tiny, text=black!70, anchor=north, yshift=-1.6pt, inner sep=0.5pt] at (axis cs:2,14.4) {14.4};
\node[font=\tiny, text=black!70, anchor=south, yshift=1.6pt, inner sep=0.5pt] at (axis cs:3,23.1) {23.1};
\node[font=\tiny, text=black!70, anchor=north, yshift=-1.6pt, inner sep=0.5pt] at (axis cs:4,22.2) {22.2};
\node[font=\tiny, text=black!70, anchor=south, yshift=1.6pt, inner sep=0.5pt] at (axis cs:5,27.7) {27.7};
\node[font=\tiny, text=black!70, anchor=north, yshift=-1.6pt, inner sep=0.5pt] at (axis cs:6,24.9) {24.9};
\nextgroupplot[title={SQuAD}, ymin=45, ymax=80]
\addplot[name path=U, draw=none, forget plot] table[x=x, y expr=\thisrow{m}+\thisrow{sd}] {
x  m  sd
0  53.6  1.86
1  56.1  1.76
2  57.5  1.74
3  62.0  1.73
4  71.6  1.54
5  71.4  1.45
6  67.6  1.36
};
\addplot[name path=L, draw=none, forget plot] table[x=x, y expr=\thisrow{m}-\thisrow{sd}] {
x  m  sd
0  53.6  1.86
1  56.1  1.76
2  57.5  1.74
3  62.0  1.73
4  71.6  1.54
5  71.4  1.45
6  67.6  1.36
};
\addplot[lineO, fill opacity=0.15, draw=none, forget plot] fill between[of=U and L];
\addplot[lineO, line width=0.9pt, mark=*, mark size=1.4pt, mark options={fill=lineO, draw=white, line width=0.35pt}] table[x=x, y=m] {
x  m  sd
0  53.6  1.86
1  56.1  1.76
2  57.5  1.74
3  62.0  1.73
4  71.6  1.54
5  71.4  1.45
6  67.6  1.36
};
\node[font=\tiny, text=black!70, anchor=north, yshift=-1.6pt, inner sep=0.5pt] at (axis cs:0,53.6) {53.6};
\node[font=\tiny, text=black!70, anchor=south, yshift=1.6pt, inner sep=0.5pt] at (axis cs:1,56.1) {56.1};
\node[font=\tiny, text=black!70, anchor=north, yshift=-1.6pt, inner sep=0.5pt] at (axis cs:2,57.5) {57.5};
\node[font=\tiny, text=black!70, anchor=north, yshift=-1.6pt, inner sep=0.5pt] at (axis cs:3,62.0) {62.0};
\node[font=\tiny, text=black!70, anchor=south, yshift=1.6pt, inner sep=0.5pt] at (axis cs:4,71.6) {71.6};
\node[font=\tiny, text=black!70, anchor=north, yshift=-1.6pt, inner sep=0.5pt] at (axis cs:5,71.4) {71.4};
\node[font=\tiny, text=black!70, anchor=north, yshift=-1.6pt, inner sep=0.5pt] at (axis cs:6,67.6) {67.6};
\nextgroupplot[title={2Wiki}, ymin=35, ymax=65]
\addplot[name path=U, draw=none, forget plot] table[x=x, y expr=\thisrow{m}+\thisrow{sd}] {
x  m  sd
0  47.0  3.44
1  42.8  3.39
2  49.3  3.42
3  55.9  3.30
4  56.8  3.18
5  57.4  3.17
6  48.7  3.25
};
\addplot[name path=L, draw=none, forget plot] table[x=x, y expr=\thisrow{m}-\thisrow{sd}] {
x  m  sd
0  47.0  3.44
1  42.8  3.39
2  49.3  3.42
3  55.9  3.30
4  56.8  3.18
5  57.4  3.17
6  48.7  3.25
};
\addplot[lineO, fill opacity=0.15, draw=none, forget plot] fill between[of=U and L];
\addplot[lineO, line width=0.9pt, mark=*, mark size=1.4pt, mark options={fill=lineO, draw=white, line width=0.35pt}] table[x=x, y=m] {
x  m  sd
0  47.0  3.44
1  42.8  3.39
2  49.3  3.42
3  55.9  3.30
4  56.8  3.18
5  57.4  3.17
6  48.7  3.25
};
\node[font=\tiny, text=black!70, anchor=south, yshift=1.6pt, inner sep=0.5pt] at (axis cs:0,47.0) {47.0};
\node[font=\tiny, text=black!70, anchor=north, yshift=-1.6pt, inner sep=0.5pt] at (axis cs:1,42.8) {42.8};
\node[font=\tiny, text=black!70, anchor=north, yshift=-1.6pt, inner sep=0.5pt] at (axis cs:2,49.3) {49.3};
\node[font=\tiny, text=black!70, anchor=south, yshift=1.6pt, inner sep=0.5pt] at (axis cs:3,55.9) {55.9};
\node[font=\tiny, text=black!70, anchor=north, yshift=-1.6pt, inner sep=0.5pt] at (axis cs:4,56.8) {56.8};
\node[font=\tiny, text=black!70, anchor=south, yshift=1.6pt, inner sep=0.5pt] at (axis cs:5,57.4) {57.4};
\node[font=\tiny, text=black!70, anchor=north, yshift=-1.6pt, inner sep=0.5pt] at (axis cs:6,48.7) {48.7};
\nextgroupplot[title={NQ}, ymin=30, ymax=70]
\addplot[name path=U, draw=none, forget plot] table[x=x, y expr=\thisrow{m}+\thisrow{sd}] {
x  m  sd
0  40.6  3.25
1  43.0  3.21
2  41.7  3.27
3  55.9  3.05
4  57.7  3.02
5  55.2  3.03
6  53.9  2.91
};
\addplot[name path=L, draw=none, forget plot] table[x=x, y expr=\thisrow{m}-\thisrow{sd}] {
x  m  sd
0  40.6  3.25
1  43.0  3.21
2  41.7  3.27
3  55.9  3.05
4  57.7  3.02
5  55.2  3.03
6  53.9  2.91
};
\addplot[lineO, fill opacity=0.15, draw=none, forget plot] fill between[of=U and L];
\addplot[lineO, line width=0.9pt, mark=*, mark size=1.4pt, mark options={fill=lineO, draw=white, line width=0.35pt}] table[x=x, y=m] {
x  m  sd
0  40.6  3.25
1  43.0  3.21
2  41.7  3.27
3  55.9  3.05
4  57.7  3.02
5  55.2  3.03
6  53.9  2.91
};
\node[font=\tiny, text=black!70, anchor=north, yshift=-1.6pt, inner sep=0.5pt] at (axis cs:0,40.6) {40.6};
\node[font=\tiny, text=black!70, anchor=south, yshift=1.6pt, inner sep=0.5pt] at (axis cs:1,43.0) {43.0};
\node[font=\tiny, text=black!70, anchor=north, yshift=-1.6pt, inner sep=0.5pt] at (axis cs:2,41.7) {41.7};
\node[font=\tiny, text=black!70, anchor=south, yshift=1.6pt, inner sep=0.5pt] at (axis cs:3,55.9) {55.9};
\node[font=\tiny, text=black!70, anchor=north, yshift=-1.6pt, inner sep=0.5pt] at (axis cs:4,57.7) {57.7};
\node[font=\tiny, text=black!70, anchor=north, yshift=-1.6pt, inner sep=0.5pt] at (axis cs:5,55.2) {55.2};
\node[font=\tiny, text=black!70, anchor=south, yshift=1.6pt, inner sep=0.5pt] at (axis cs:6,53.9) {53.9};
\nextgroupplot[title={TriviaQA}, ymin=40, ymax=70, ylabel={F1}, ylabel style={font=\scriptsize, yshift=-3pt}]
\addplot[name path=U, draw=none, forget plot] table[x=x, y expr=\thisrow{m}+\thisrow{sd}] {
x  m  sd
0  47.9  2.50
1  50.5  2.49
2  52.2  2.30
3  57.6  2.42
4  61.9  2.24
5  59.4  2.36
6  57.5  2.25
};
\addplot[name path=L, draw=none, forget plot] table[x=x, y expr=\thisrow{m}-\thisrow{sd}] {
x  m  sd
0  47.9  2.50
1  50.5  2.49
2  52.2  2.30
3  57.6  2.42
4  61.9  2.24
5  59.4  2.36
6  57.5  2.25
};
\addplot[lineO, fill opacity=0.15, draw=none, forget plot] fill between[of=U and L];
\addplot[lineO, line width=0.9pt, mark=*, mark size=1.4pt, mark options={fill=lineO, draw=white, line width=0.35pt}] table[x=x, y=m] {
x  m  sd
0  47.9  2.50
1  50.5  2.49
2  52.2  2.30
3  57.6  2.42
4  61.9  2.24
5  59.4  2.36
6  57.5  2.25
};
\node[font=\tiny, text=black!70, anchor=north, yshift=-1.6pt, inner sep=0.5pt] at (axis cs:0,47.9) {47.9};
\node[font=\tiny, text=black!70, anchor=south, yshift=1.6pt, inner sep=0.5pt] at (axis cs:1,50.5) {50.5};
\node[font=\tiny, text=black!70, anchor=north, yshift=-1.6pt, inner sep=0.5pt] at (axis cs:2,52.2) {52.2};
\node[font=\tiny, text=black!70, anchor=south, yshift=1.6pt, inner sep=0.5pt] at (axis cs:3,57.6) {57.6};
\node[font=\tiny, text=black!70, anchor=south, yshift=1.6pt, inner sep=0.5pt] at (axis cs:4,61.9) {61.9};
\node[font=\tiny, text=black!70, anchor=north, yshift=-1.6pt, inner sep=0.5pt] at (axis cs:5,59.4) {59.4};
\node[font=\tiny, text=black!70, anchor=north, yshift=-1.6pt, inner sep=0.5pt] at (axis cs:6,57.5) {57.5};
\nextgroupplot[title={Qasper}, ymin=10, ymax=55]
\addplot[name path=U, draw=none, forget plot] table[x=x, y expr=\thisrow{m}+\thisrow{sd}] {
x  m  sd
0  20.8  1.79
1  24.1  1.86
2  25.3  1.83
3  32.1  1.75
4  44.1  1.89
5  42.0  1.65
6  42.1  1.80
};
\addplot[name path=L, draw=none, forget plot] table[x=x, y expr=\thisrow{m}-\thisrow{sd}] {
x  m  sd
0  20.8  1.79
1  24.1  1.86
2  25.3  1.83
3  32.1  1.75
4  44.1  1.89
5  42.0  1.65
6  42.1  1.80
};
\addplot[lineO, fill opacity=0.15, draw=none, forget plot] fill between[of=U and L];
\addplot[lineO, line width=0.9pt, mark=*, mark size=1.4pt, mark options={fill=lineO, draw=white, line width=0.35pt}] table[x=x, y=m] {
x  m  sd
0  20.8  1.79
1  24.1  1.86
2  25.3  1.83
3  32.1  1.75
4  44.1  1.89
5  42.0  1.65
6  42.1  1.80
};
\node[font=\tiny, text=black!70, anchor=north, yshift=-1.6pt, inner sep=0.5pt] at (axis cs:0,20.8) {20.8};
\node[font=\tiny, text=black!70, anchor=south, yshift=1.6pt, inner sep=0.5pt] at (axis cs:1,24.1) {24.1};
\node[font=\tiny, text=black!70, anchor=north, yshift=-1.6pt, inner sep=0.5pt] at (axis cs:2,25.3) {25.3};
\node[font=\tiny, text=black!70, anchor=north, yshift=-1.6pt, inner sep=0.5pt] at (axis cs:3,32.1) {32.1};
\node[font=\tiny, text=black!70, anchor=south, yshift=1.6pt, inner sep=0.5pt] at (axis cs:4,44.1) {44.1};
\node[font=\tiny, text=black!70, anchor=north, yshift=-1.6pt, inner sep=0.5pt] at (axis cs:5,42.0) {42.0};
\node[font=\tiny, text=black!70, anchor=south, yshift=1.6pt, inner sep=0.5pt] at (axis cs:6,42.1) {42.1};
\nextgroupplot[title={MultiFieldQA}, ymin=5, ymax=40]
\addplot[name path=U, draw=none, forget plot] table[x=x, y expr=\thisrow{m}+\thisrow{sd}] {
x  m  sd
0  13.5  2.58
1  18.0  3.08
2  15.6  2.87
3  22.6  2.75
4  30.5  3.27
5  29.3  3.26
6  28.5  3.21
};
\addplot[name path=L, draw=none, forget plot] table[x=x, y expr=\thisrow{m}-\thisrow{sd}] {
x  m  sd
0  13.5  2.58
1  18.0  3.08
2  15.6  2.87
3  22.6  2.75
4  30.5  3.27
5  29.3  3.26
6  28.5  3.21
};
\addplot[lineO, fill opacity=0.15, draw=none, forget plot] fill between[of=U and L];
\addplot[lineO, line width=0.9pt, mark=*, mark size=1.4pt, mark options={fill=lineO, draw=white, line width=0.35pt}] table[x=x, y=m] {
x  m  sd
0  13.5  2.58
1  18.0  3.08
2  15.6  2.87
3  22.6  2.75
4  30.5  3.27
5  29.3  3.26
6  28.5  3.21
};
\node[font=\tiny, text=black!70, anchor=north, yshift=-1.6pt, inner sep=0.5pt] at (axis cs:0,13.5) {13.5};
\node[font=\tiny, text=black!70, anchor=south, yshift=1.6pt, inner sep=0.5pt] at (axis cs:1,18.0) {18.0};
\node[font=\tiny, text=black!70, anchor=north, yshift=-1.6pt, inner sep=0.5pt] at (axis cs:2,15.6) {15.6};
\node[font=\tiny, text=black!70, anchor=north, yshift=-1.6pt, inner sep=0.5pt] at (axis cs:3,22.6) {22.6};
\node[font=\tiny, text=black!70, anchor=south, yshift=1.6pt, inner sep=0.5pt] at (axis cs:4,30.5) {30.5};
\node[font=\tiny, text=black!70, anchor=north, yshift=-1.6pt, inner sep=0.5pt] at (axis cs:5,29.3) {29.3};
\node[font=\tiny, text=black!70, anchor=south, yshift=1.6pt, inner sep=0.5pt] at (axis cs:6,28.5) {28.5};
\nextgroupplot[title={NarrativeQA}, ymin=0, ymax=30]
\addplot[name path=U, draw=none, forget plot] table[x=x, y expr=\thisrow{m}+\thisrow{sd}] {
x  m  sd
0  9.1  0.95
1  11.0  0.95
2  11.8  1.17
3  18.9  1.36
4  24.0  1.25
5  23.3  1.24
6  24.6  1.32
};
\addplot[name path=L, draw=none, forget plot] table[x=x, y expr=\thisrow{m}-\thisrow{sd}] {
x  m  sd
0  9.1  0.95
1  11.0  0.95
2  11.8  1.17
3  18.9  1.36
4  24.0  1.25
5  23.3  1.24
6  24.6  1.32
};
\addplot[lineO, fill opacity=0.15, draw=none, forget plot] fill between[of=U and L];
\addplot[lineO, line width=0.9pt, mark=*, mark size=1.4pt, mark options={fill=lineO, draw=white, line width=0.35pt}] table[x=x, y=m] {
x  m  sd
0  9.1  0.95
1  11.0  0.95
2  11.8  1.17
3  18.9  1.36
4  24.0  1.25
5  23.3  1.24
6  24.6  1.32
};
\node[font=\tiny, text=black!70, anchor=north, yshift=-1.6pt, inner sep=0.5pt] at (axis cs:0,9.1) {9.1};
\node[font=\tiny, text=black!70, anchor=south, yshift=1.6pt, inner sep=0.5pt] at (axis cs:1,11.0) {11.0};
\node[font=\tiny, text=black!70, anchor=north, yshift=-1.6pt, inner sep=0.5pt] at (axis cs:2,11.8) {11.8};
\node[font=\tiny, text=black!70, anchor=south, yshift=1.6pt, inner sep=0.5pt] at (axis cs:3,18.9) {18.9};
\node[font=\tiny, text=black!70, anchor=south, yshift=1.6pt, inner sep=0.5pt] at (axis cs:4,24.0) {24.0};
\node[font=\tiny, text=black!70, anchor=north, yshift=-1.6pt, inner sep=0.5pt] at (axis cs:5,23.3) {23.3};
\node[font=\tiny, text=black!70, anchor=south, yshift=1.6pt, inner sep=0.5pt] at (axis cs:6,24.6) {24.6};
\nextgroupplot[title={LoCoMo}, ymin=0, ymax=45]
\addplot[name path=U, draw=none, forget plot] table[x=x, y expr=\thisrow{m}+\thisrow{sd}] {
x  m  sd
0  8.5  0.64
1  10.5  0.68
2  11.8  0.72
3  19.4  0.80
4  34.1  0.95
5  33.4  0.97
6  36.2  0.97
};
\addplot[name path=L, draw=none, forget plot] table[x=x, y expr=\thisrow{m}-\thisrow{sd}] {
x  m  sd
0  8.5  0.64
1  10.5  0.68
2  11.8  0.72
3  19.4  0.80
4  34.1  0.95
5  33.4  0.97
6  36.2  0.97
};
\addplot[lineO, fill opacity=0.15, draw=none, forget plot] fill between[of=U and L];
\addplot[lineO, line width=0.9pt, mark=*, mark size=1.4pt, mark options={fill=lineO, draw=white, line width=0.35pt}] table[x=x, y=m] {
x  m  sd
0  8.5  0.64
1  10.5  0.68
2  11.8  0.72
3  19.4  0.80
4  34.1  0.95
5  33.4  0.97
6  36.2  0.97
};
\node[font=\tiny, text=black!70, anchor=north, yshift=-1.6pt, inner sep=0.5pt] at (axis cs:0,8.5) {8.5};
\node[font=\tiny, text=black!70, anchor=south, yshift=1.6pt, inner sep=0.5pt] at (axis cs:1,10.5) {10.5};
\node[font=\tiny, text=black!70, anchor=north, yshift=-1.6pt, inner sep=0.5pt] at (axis cs:2,11.8) {11.8};
\node[font=\tiny, text=black!70, anchor=north, yshift=-1.6pt, inner sep=0.5pt] at (axis cs:3,19.4) {19.4};
\node[font=\tiny, text=black!70, anchor=south, yshift=1.6pt, inner sep=0.5pt] at (axis cs:4,34.1) {34.1};
\node[font=\tiny, text=black!70, anchor=north, yshift=-1.6pt, inner sep=0.5pt] at (axis cs:5,33.4) {33.4};
\node[font=\tiny, text=black!70, anchor=south, yshift=1.6pt, inner sep=0.5pt] at (axis cs:6,36.2) {36.2};
\end{groupplot}
\end{tikzpicture}
\caption{F1 of RoDeL-Seq with Qwen3.5-4B on each dataset after each update. HotpotQA and MuSiQue (ID) are in-domain and the others out-of-domain. Fr.: frozen roles (PRAG-style). Shaded bands show $\pm 1$ standard deviation of a test-set bootstrap.}
\label{fig:perset4b}
\end{figure}

\begin{figure}[!htb]
\centering
\begin{tikzpicture}
\begin{groupplot}[
  group style={group size=5 by 2, horizontal sep=0.55cm, vertical sep=0.95cm},
  scale only axis, width=2.5cm, height=2.0cm,
  xmin=-0.5, xmax=3.5,
  xtick={0,...,3},
  xticklabels={Fr.,$u_1$,$u_2$,$u_3$},
  axis lines=left, axis line style={-, black!60}, tick style={black!60}, major tick length=2pt,
  tick label style={font=\tiny, inner sep=1pt},
  title style={font=\footnotesize, yshift=-4pt},
  ymajorgrids, grid style={black!22, dashed},
  clip=false,
]
\nextgroupplot[title={HotpotQA (ID)}, ymin=30, ymax=65, ylabel={F1}, ylabel style={font=\scriptsize, yshift=-3pt}]
\addplot[name path=U, draw=none, forget plot] table[x=x, y expr=\thisrow{m}+\thisrow{sd}] {
x  m  sd
0  39.5  3.26
1  38.1  3.20
2  48.4  3.21
3  58.1  3.17
};
\addplot[name path=L, draw=none, forget plot] table[x=x, y expr=\thisrow{m}-\thisrow{sd}] {
x  m  sd
0  39.5  3.26
1  38.1  3.20
2  48.4  3.21
3  58.1  3.17
};
\addplot[lineO, fill opacity=0.15, draw=none, forget plot] fill between[of=U and L];
\addplot[lineO, line width=0.9pt, mark=*, mark size=1.4pt, mark options={fill=lineO, draw=white, line width=0.35pt}] table[x=x, y=m] {
x  m  sd
0  39.5  3.26
1  38.1  3.20
2  48.4  3.21
3  58.1  3.17
};
\node[font=\tiny, text=black!70, anchor=south, yshift=1.6pt, inner sep=0.5pt] at (axis cs:0,39.5) {39.5};
\node[font=\tiny, text=black!70, anchor=north, yshift=-1.6pt, inner sep=0.5pt] at (axis cs:1,38.1) {38.1};
\node[font=\tiny, text=black!70, anchor=north west, yshift=-1.6pt, inner sep=0.5pt] at (axis cs:2,48.4) {48.4};
\node[font=\tiny, text=black!70, anchor=south, yshift=1.6pt, inner sep=0.5pt] at (axis cs:3,58.1) {58.1};
\nextgroupplot[title={MuSiQue (ID)}, ymin=0, ymax=40]
\addplot[name path=U, draw=none, forget plot] table[x=x, y expr=\thisrow{m}+\thisrow{sd}] {
x  m  sd
0  7.3  1.67
1  12.3  2.01
2  31.9  3.00
3  30.3  2.80
};
\addplot[name path=L, draw=none, forget plot] table[x=x, y expr=\thisrow{m}-\thisrow{sd}] {
x  m  sd
0  7.3  1.67
1  12.3  2.01
2  31.9  3.00
3  30.3  2.80
};
\addplot[lineO, fill opacity=0.15, draw=none, forget plot] fill between[of=U and L];
\addplot[lineO, line width=0.9pt, mark=*, mark size=1.4pt, mark options={fill=lineO, draw=white, line width=0.35pt}] table[x=x, y=m] {
x  m  sd
0  7.3  1.67
1  12.3  2.01
2  31.9  3.00
3  30.3  2.80
};
\node[font=\tiny, text=black!70, anchor=north, yshift=-1.6pt, inner sep=0.5pt] at (axis cs:0,7.3) {7.3};
\node[font=\tiny, text=black!70, anchor=north west, yshift=-1.6pt, inner sep=0.5pt] at (axis cs:1,12.3) {12.3};
\node[font=\tiny, text=black!70, anchor=south, yshift=1.6pt, inner sep=0.5pt] at (axis cs:2,31.9) {31.9};
\node[font=\tiny, text=black!70, anchor=north, yshift=-1.6pt, inner sep=0.5pt] at (axis cs:3,30.3) {30.3};
\nextgroupplot[title={SQuAD}, ymin=45, ymax=80]
\addplot[name path=U, draw=none, forget plot] table[x=x, y expr=\thisrow{m}+\thisrow{sd}] {
x  m  sd
0  53.6  1.86
1  55.4  1.72
2  72.4  1.27
3  69.8  1.33
};
\addplot[name path=L, draw=none, forget plot] table[x=x, y expr=\thisrow{m}-\thisrow{sd}] {
x  m  sd
0  53.6  1.86
1  55.4  1.72
2  72.4  1.27
3  69.8  1.33
};
\addplot[lineO, fill opacity=0.15, draw=none, forget plot] fill between[of=U and L];
\addplot[lineO, line width=0.9pt, mark=*, mark size=1.4pt, mark options={fill=lineO, draw=white, line width=0.35pt}] table[x=x, y=m] {
x  m  sd
0  53.6  1.86
1  55.4  1.72
2  72.4  1.27
3  69.8  1.33
};
\node[font=\tiny, text=black!70, anchor=north, yshift=-1.6pt, inner sep=0.5pt] at (axis cs:0,53.6) {53.6};
\node[font=\tiny, text=black!70, anchor=north west, yshift=-1.6pt, inner sep=0.5pt] at (axis cs:1,55.4) {55.4};
\node[font=\tiny, text=black!70, anchor=south, yshift=1.6pt, inner sep=0.5pt] at (axis cs:2,72.4) {72.4};
\node[font=\tiny, text=black!70, anchor=north, yshift=-1.6pt, inner sep=0.5pt] at (axis cs:3,69.8) {69.8};
\nextgroupplot[title={2Wiki}, ymin=25, ymax=65]
\addplot[name path=U, draw=none, forget plot] table[x=x, y expr=\thisrow{m}+\thisrow{sd}] {
x  m  sd
0  47.0  3.44
1  35.2  3.19
2  59.1  3.20
3  55.5  3.19
};
\addplot[name path=L, draw=none, forget plot] table[x=x, y expr=\thisrow{m}-\thisrow{sd}] {
x  m  sd
0  47.0  3.44
1  35.2  3.19
2  59.1  3.20
3  55.5  3.19
};
\addplot[lineO, fill opacity=0.15, draw=none, forget plot] fill between[of=U and L];
\addplot[lineO, line width=0.9pt, mark=*, mark size=1.4pt, mark options={fill=lineO, draw=white, line width=0.35pt}] table[x=x, y=m] {
x  m  sd
0  47.0  3.44
1  35.2  3.19
2  59.1  3.20
3  55.5  3.19
};
\node[font=\tiny, text=black!70, anchor=south, yshift=1.6pt, inner sep=0.5pt] at (axis cs:0,47.0) {47.0};
\node[font=\tiny, text=black!70, anchor=north, yshift=-1.6pt, inner sep=0.5pt] at (axis cs:1,35.2) {35.2};
\node[font=\tiny, text=black!70, anchor=south, yshift=1.6pt, inner sep=0.5pt] at (axis cs:2,59.1) {59.1};
\node[font=\tiny, text=black!70, anchor=north, yshift=-1.6pt, inner sep=0.5pt] at (axis cs:3,55.5) {55.5};
\nextgroupplot[title={NQ}, ymin=30, ymax=70]
\addplot[name path=U, draw=none, forget plot] table[x=x, y expr=\thisrow{m}+\thisrow{sd}] {
x  m  sd
0  40.6  3.25
1  53.3  3.00
2  63.8  2.85
3  58.8  2.81
};
\addplot[name path=L, draw=none, forget plot] table[x=x, y expr=\thisrow{m}-\thisrow{sd}] {
x  m  sd
0  40.6  3.25
1  53.3  3.00
2  63.8  2.85
3  58.8  2.81
};
\addplot[lineO, fill opacity=0.15, draw=none, forget plot] fill between[of=U and L];
\addplot[lineO, line width=0.9pt, mark=*, mark size=1.4pt, mark options={fill=lineO, draw=white, line width=0.35pt}] table[x=x, y=m] {
x  m  sd
0  40.6  3.25
1  53.3  3.00
2  63.8  2.85
3  58.8  2.81
};
\node[font=\tiny, text=black!70, anchor=north, yshift=-1.6pt, inner sep=0.5pt] at (axis cs:0,40.6) {40.6};
\node[font=\tiny, text=black!70, anchor=south east, yshift=1.6pt, inner sep=0.5pt] at (axis cs:1,53.3) {53.3};
\node[font=\tiny, text=black!70, anchor=south, yshift=1.6pt, inner sep=0.5pt] at (axis cs:2,63.8) {63.8};
\node[font=\tiny, text=black!70, anchor=north, yshift=-1.6pt, inner sep=0.5pt] at (axis cs:3,58.8) {58.8};
\nextgroupplot[title={TriviaQA}, ymin=40, ymax=70, ylabel={F1}, ylabel style={font=\scriptsize, yshift=-3pt}]
\addplot[name path=U, draw=none, forget plot] table[x=x, y expr=\thisrow{m}+\thisrow{sd}] {
x  m  sd
0  47.9  2.50
1  55.1  2.27
2  59.3  2.44
3  61.3  2.23
};
\addplot[name path=L, draw=none, forget plot] table[x=x, y expr=\thisrow{m}-\thisrow{sd}] {
x  m  sd
0  47.9  2.50
1  55.1  2.27
2  59.3  2.44
3  61.3  2.23
};
\addplot[lineO, fill opacity=0.15, draw=none, forget plot] fill between[of=U and L];
\addplot[lineO, line width=0.9pt, mark=*, mark size=1.4pt, mark options={fill=lineO, draw=white, line width=0.35pt}] table[x=x, y=m] {
x  m  sd
0  47.9  2.50
1  55.1  2.27
2  59.3  2.44
3  61.3  2.23
};
\node[font=\tiny, text=black!70, anchor=north, yshift=-1.6pt, inner sep=0.5pt] at (axis cs:0,47.9) {47.9};
\node[font=\tiny, text=black!70, anchor=south east, yshift=1.6pt, inner sep=0.5pt] at (axis cs:1,55.1) {55.1};
\node[font=\tiny, text=black!70, anchor=north west, yshift=-1.6pt, inner sep=0.5pt] at (axis cs:2,59.3) {59.3};
\node[font=\tiny, text=black!70, anchor=south, yshift=1.6pt, inner sep=0.5pt] at (axis cs:3,61.3) {61.3};
\nextgroupplot[title={Qasper}, ymin=10, ymax=55]
\addplot[name path=U, draw=none, forget plot] table[x=x, y expr=\thisrow{m}+\thisrow{sd}] {
x  m  sd
0  20.8  1.79
1  35.9  1.76
2  41.3  1.76
3  44.1  1.74
};
\addplot[name path=L, draw=none, forget plot] table[x=x, y expr=\thisrow{m}-\thisrow{sd}] {
x  m  sd
0  20.8  1.79
1  35.9  1.76
2  41.3  1.76
3  44.1  1.74
};
\addplot[lineO, fill opacity=0.15, draw=none, forget plot] fill between[of=U and L];
\addplot[lineO, line width=0.9pt, mark=*, mark size=1.4pt, mark options={fill=lineO, draw=white, line width=0.35pt}] table[x=x, y=m] {
x  m  sd
0  20.8  1.79
1  35.9  1.76
2  41.3  1.76
3  44.1  1.74
};
\node[font=\tiny, text=black!70, anchor=north, yshift=-1.6pt, inner sep=0.5pt] at (axis cs:0,20.8) {20.8};
\node[font=\tiny, text=black!70, anchor=south east, yshift=1.6pt, inner sep=0.5pt] at (axis cs:1,35.9) {35.9};
\node[font=\tiny, text=black!70, anchor=north west, yshift=-1.6pt, inner sep=0.5pt] at (axis cs:2,41.3) {41.3};
\node[font=\tiny, text=black!70, anchor=south, yshift=1.6pt, inner sep=0.5pt] at (axis cs:3,44.1) {44.1};
\nextgroupplot[title={MultiFieldQA}, ymin=5, ymax=40]
\addplot[name path=U, draw=none, forget plot] table[x=x, y expr=\thisrow{m}+\thisrow{sd}] {
x  m  sd
0  13.5  2.58
1  21.3  2.96
2  30.2  2.90
3  27.3  2.90
};
\addplot[name path=L, draw=none, forget plot] table[x=x, y expr=\thisrow{m}-\thisrow{sd}] {
x  m  sd
0  13.5  2.58
1  21.3  2.96
2  30.2  2.90
3  27.3  2.90
};
\addplot[lineO, fill opacity=0.15, draw=none, forget plot] fill between[of=U and L];
\addplot[lineO, line width=0.9pt, mark=*, mark size=1.4pt, mark options={fill=lineO, draw=white, line width=0.35pt}] table[x=x, y=m] {
x  m  sd
0  13.5  2.58
1  21.3  2.96
2  30.2  2.90
3  27.3  2.90
};
\node[font=\tiny, text=black!70, anchor=north, yshift=-1.6pt, inner sep=0.5pt] at (axis cs:0,13.5) {13.5};
\node[font=\tiny, text=black!70, anchor=south east, yshift=1.6pt, inner sep=0.5pt] at (axis cs:1,21.3) {21.3};
\node[font=\tiny, text=black!70, anchor=south, yshift=1.6pt, inner sep=0.5pt] at (axis cs:2,30.2) {30.2};
\node[font=\tiny, text=black!70, anchor=north, yshift=-1.6pt, inner sep=0.5pt] at (axis cs:3,27.3) {27.3};
\nextgroupplot[title={NarrativeQA}, ymin=0, ymax=30]
\addplot[name path=U, draw=none, forget plot] table[x=x, y expr=\thisrow{m}+\thisrow{sd}] {
x  m  sd
0  9.1  0.95
1  17.9  1.37
2  21.9  1.33
3  23.7  1.22
};
\addplot[name path=L, draw=none, forget plot] table[x=x, y expr=\thisrow{m}-\thisrow{sd}] {
x  m  sd
0  9.1  0.95
1  17.9  1.37
2  21.9  1.33
3  23.7  1.22
};
\addplot[lineO, fill opacity=0.15, draw=none, forget plot] fill between[of=U and L];
\addplot[lineO, line width=0.9pt, mark=*, mark size=1.4pt, mark options={fill=lineO, draw=white, line width=0.35pt}] table[x=x, y=m] {
x  m  sd
0  9.1  0.95
1  17.9  1.37
2  21.9  1.33
3  23.7  1.22
};
\node[font=\tiny, text=black!70, anchor=north, yshift=-1.6pt, inner sep=0.5pt] at (axis cs:0,9.1) {9.1};
\node[font=\tiny, text=black!70, anchor=south east, yshift=1.6pt, inner sep=0.5pt] at (axis cs:1,17.9) {17.9};
\node[font=\tiny, text=black!70, anchor=north west, yshift=-1.6pt, inner sep=0.5pt] at (axis cs:2,21.9) {21.9};
\node[font=\tiny, text=black!70, anchor=south, yshift=1.6pt, inner sep=0.5pt] at (axis cs:3,23.7) {23.7};
\nextgroupplot[title={LoCoMo}, ymin=0, ymax=45]
\addplot[name path=U, draw=none, forget plot] table[x=x, y expr=\thisrow{m}+\thisrow{sd}] {
x  m  sd
0  8.5  0.64
1  19.5  0.83
2  34.7  0.96
3  36.9  0.94
};
\addplot[name path=L, draw=none, forget plot] table[x=x, y expr=\thisrow{m}-\thisrow{sd}] {
x  m  sd
0  8.5  0.64
1  19.5  0.83
2  34.7  0.96
3  36.9  0.94
};
\addplot[lineO, fill opacity=0.15, draw=none, forget plot] fill between[of=U and L];
\addplot[lineO, line width=0.9pt, mark=*, mark size=1.4pt, mark options={fill=lineO, draw=white, line width=0.35pt}] table[x=x, y=m] {
x  m  sd
0  8.5  0.64
1  19.5  0.83
2  34.7  0.96
3  36.9  0.94
};
\node[font=\tiny, text=black!70, anchor=north, yshift=-1.6pt, inner sep=0.5pt] at (axis cs:0,8.5) {8.5};
\node[font=\tiny, text=black!70, anchor=north west, yshift=-1.6pt, inner sep=0.5pt] at (axis cs:1,19.5) {19.5};
\node[font=\tiny, text=black!70, anchor=north west, yshift=-1.6pt, inner sep=0.5pt] at (axis cs:2,34.7) {34.7};
\node[font=\tiny, text=black!70, anchor=south, yshift=1.6pt, inner sep=0.5pt] at (axis cs:3,36.9) {36.9};
\end{groupplot}
\end{tikzpicture}
\caption{F1 of RoDeL-Sim with Qwen3.5-4B on each dataset after each simultaneous update. Datasets, Fr., and shaded bands are as in \Cref{fig:perset4b}.}
\label{fig:persetsim}
\end{figure}

\subsection{Comparison against Baselines}
\label{subsec:baselines}

\Cref{tab:main} compares the methods at their best points, chosen separately for the in-domain and out-of-domain averages. RoDeL-Seq, SEAL-style, and PaST-style each train for six epochs, and RoDeL-Sim for three simultaneous updates. The in-domain and out-of-domain best points are after epochs 5 and 4 for RoDeL-Seq ($s_3$ and $t_2$), after $u_3$ and $u_2$ for RoDeL-Sim, after epochs 5 and 6 for SEAL-style, and after epoch 6 for PaST-style. PRAG-style has a single model. Both variants have a higher average F1 than every baseline, in domain (38.0 for RoDeL-Seq and 44.2 for RoDeL-Sim) and out of domain (47.6 and 47.8). On every dataset, the highest F1 among the five methods belongs to one of the two variants. Their comparison with PRAG-style, the starting point of training, is given in \Cref{subsec:dynamics}.

\begin{table}[!htb]
\centering
\footnotesize
\setlength{\tabcolsep}{3.1pt}
\caption{F1 of each method at its best point with Qwen3.5-4B. For each method, the in-domain columns use its checkpoint with the highest in-domain average, and the out-of-domain columns its checkpoint with the highest out-of-domain average. The two numbers below each baseline score are its differences from RoDeL-Seq and from RoDeL-Sim, in red when the variant is higher and in green when the baseline is higher; $^*$ marks a difference with $p<0.05$, and $^\dagger$ one that remains significant after Holm correction across the datasets of the group; averages carry only $^*$ (\Cref{app:results}). The rows on a gray background are references with the base model $\theta$ and no adapter: No memory answers from the question alone, and In-context answers with the whole document in its context (\Cref{subsec:efficiency}). MFQA: MultiFieldQA; NarQA: NarrativeQA. Bold: best in each column among the five methods.}
\label{tab:main}
\begin{tabular}{@{}l rr!{\hspace{3pt}}r@{\hspace{10pt}} rr!{\hspace{3pt}}rrrrrr!{\hspace{3pt}}r@{\hspace{6pt}}}
\toprule
 & \multicolumn{3}{c}{In-domain} & \multicolumn{9}{c}{Out-of-domain} \\
\cmidrule(lr){2-4}\cmidrule(l){5-13}
Method & HotpotQA & MuSiQue & Avg. & SQuAD & 2Wiki & NQ & TriviaQA & Qasper & MFQA & NarQA & LoCoMo & Avg. \\
\midrule
\refrows{2}
No memory & 13.3 & 0.6 & 7.0 & 10.9 & 7.2 & 15.4 & 48.2 & 5.6 & 4.3 & 0.7 & 0.7 & 11.6 \\
In-context & 62.8 & 30.6 & 46.7 & 73.6 & 65.5 & 43.1 & 47.6 & 26.6 & 35.5 & 19.8 & 28.7 & 42.5 \\
\midrule
\textit{PRAG-style} & 39.5 & 7.3 & 23.4 & 53.6 & 47.0 & 40.6 & 47.9 & 20.8 & 13.5 & 9.1 & 8.5 & 30.1 \\[-1.5pt]
\vsrow{Seq} & \dlt{8.8}\dmk{{*\dagger}} & \dlt{20.4}\dmk{{*\dagger}} & \dlt{14.6}\dmk{*} & \dlt{18.0}\dmk{{*\dagger}} & \dlt{9.8}\dmk{{*\dagger}} & \dlt{17.1}\dmk{{*\dagger}} & \dlt{14.0}\dmk{{*\dagger}} & \dlt{23.3}\dmk{{*\dagger}} & \dlt{17.0}\dmk{{*\dagger}} & \dlt{14.9}\dmk{{*\dagger}} & \dlt{25.6}\dmk{{*\dagger}} & \dlt{17.5}\dmk{*} \\[-1.5pt]
\vsrow{Sim} & \dlt{18.6}\dmk{{*\dagger}} & \dlt{23.0}\dmk{{*\dagger}} & \dlt{20.8}\dmk{*} & \dlt{18.8}\dmk{{*\dagger}} & \dlt{12.1}\dmk{{*\dagger}} & \dlt{23.2}\dmk{{*\dagger}} & \dlt{11.4}\dmk{{*\dagger}} & \dlt{20.5}\dmk{{*\dagger}} & \dlt{16.7}\dmk{{*\dagger}} & \dlt{12.8}\dmk{{*\dagger}} & \dlt{26.2}\dmk{{*\dagger}} & \dlt{17.7}\dmk{*} \\[2pt]
\textit{SEAL-style} & 44.3 & 22.1 & 33.2 & 70.0 & 51.2 & 54.2 & 56.3 & 38.5 & 30.1 & 22.4 & 31.6 & 44.3 \\[-1.5pt]
\vsrow{Seq} & \dlt{4.0} & \dlt{5.6} & \dlt{4.8}\dmk{*} & \dlt{1.6} & \dlt{5.6} & \dlt{3.5} & \dlt{5.6}\dmk{{*\dagger}} & \dlt{5.6}\dmk{{*\dagger}} & \dlt{0.4} & \dlt{1.6} & \dlt{2.5}\dmk{*} & \dlt{3.3}\dmk{*} \\[-1.5pt]
\vsrow{Sim} & \dlt{13.8}\dmk{{*\dagger}} & \dlt{8.2}\dmk{{*\dagger}} & \dlt{11.0}\dmk{*} & \dlt{2.4} & \dlt{7.9}\dmk{*} & \dlt{9.6}\dmk{{*\dagger}} & \dlt{3.0} & \dlt{2.8} & \dlt{0.1} & \dltp{0.5} & \dlt{3.1}\dmk{{*\dagger}} & \dlt{3.5}\dmk{*} \\[2pt]
\textit{PaST-style} & 48.0 & 21.1 & 34.6 & 63.8 & 52.3 & 55.3 & 60.0 & 34.1 & 23.1 & 19.9 & 22.7 & 41.4 \\[-1.5pt]
\vsrow{Seq} & \dlt{0.3} & \dlt{6.6}\dmk{*} & \dlt{3.4} & \dlt{7.8}\dmk{{*\dagger}} & \dlt{4.5} & \dlt{2.4} & \dlt{1.9} & \dlt{10.0}\dmk{{*\dagger}} & \dlt{7.4}\dmk{*} & \dlt{4.1}\dmk{{*\dagger}} & \dlt{11.4}\dmk{{*\dagger}} & \dlt{6.2}\dmk{*} \\[-1.5pt]
\vsrow{Sim} & \dlt{10.1}\dmk{{*\dagger}} & \dlt{9.2}\dmk{{*\dagger}} & \dlt{9.6}\dmk{*} & \dlt{8.6}\dmk{{*\dagger}} & \dlt{6.8}\dmk{*} & \dlt{8.5}\dmk{{*\dagger}} & \dltp{0.7} & \dlt{7.2}\dmk{{*\dagger}} & \dlt{7.1}\dmk{*} & \dlt{2.0}\dmk{*} & \dlt{12.0}\dmk{{*\dagger}} & \dlt{6.4}\dmk{*} \\
\midrule
RoDeL-Seq & 48.3 & 27.7 & 38.0 & 71.6 & 56.8 & 57.7 & \textbf{61.9} & \textbf{44.1} & \textbf{30.5} & \textbf{24.0} & 34.1 & 47.6 \\
RoDeL-Sim & \textbf{58.1} & \textbf{30.3} & \textbf{44.2} & \textbf{72.4} & \textbf{59.1} & \textbf{63.8} & 59.3 & 41.3 & 30.2 & 21.9 & \textbf{34.7} & \textbf{47.8} \\
\bottomrule
\end{tabular}
\end{table}

Against SEAL-style, RoDeL-Seq leads by 4.8 points in domain and 3.3 out of domain, and RoDeL-Sim by 11.0 and 3.5, all with $p<0.05$. Against PaST-style, RoDeL-Seq leads by 3.4 points in domain and 6.2 out of domain, and RoDeL-Sim by 9.6 and 6.4; these leads have $p<0.05$ except the 3.4-point one. For both variants, the out-of-domain lead over PaST-style is largest on LoCoMo (11.4 and 12.0).

\subsection{Long-Context Retrieval Scenarios}
\label{subsec:retrieval}

In the comparison of \Cref{subsec:baselines}, each test question reads the memory adapter written for its document. When the knowledge spans many documents, the memory adapter to read needs to be selected by retrieval. To evaluate this scenario, the documents of each test set form one corpus, from 35K tokens on SQuAD to 1.37M on NQ (\Cref{tab:retrieval}). Each question reads the memory adapter of the top-1 document retrieved by Qwen3-Embedding-8B~\citep{zhang2025qwen3embedding}. All methods share these retrieval results and keep their checkpoints from \Cref{tab:main}.

\begin{table}[!htb]
\centering
\footnotesize
\setlength{\tabcolsep}{2.9pt}
\caption{F1 of each method with retrieval and Qwen3.5-4B. Each question reads the memory adapter of the top-1 document that Qwen3-Embedding-8B retrieves from the documents of its test set, and each method uses its checkpoints in \Cref{tab:main}. Tokens: total number of tokens in the documents of each test set, counted with the Qwen3.5 tokenizer. R@1: percentage of questions whose top-1 document is their own. The two numbers below each baseline score are its differences from RoDeL-Seq and from RoDeL-Sim, with colors, $^*$, and $^\dagger$ as in \Cref{tab:main}; the gray row below each variant (vs.\ oracle) is its difference from its F1 under oracle retrieval in \Cref{tab:main}. The row on a gray background is a reference: In-context is the base model $\theta$ with the retrieved top-1 document in its context. Bold: best in each column among the five methods.}
\label{tab:retrieval}
\begin{tabular}{@{}l rr!{\hspace{3pt}}r@{\hspace{10pt}} rr!{\hspace{3pt}}rrrrrr!{\hspace{3pt}}r@{\hspace{6pt}}}
\toprule
 & \multicolumn{3}{c}{In-domain} & \multicolumn{9}{c}{Out-of-domain} \\
\cmidrule(lr){2-4}\cmidrule(l){5-13}
Method & HotpotQA & MuSiQue & Avg. & SQuAD & 2Wiki & NQ & TriviaQA & Qasper & MFQA & NarQA & LoCoMo & Avg. \\
\midrule
\textcolor{black!55}{\textit{Tokens}} & \textcolor{black!55}{\textit{271K}} & \textcolor{black!55}{\textit{491K}} & \textcolor{black!55}{\textit{--}} & \textcolor{black!55}{\textit{35K}} & \textcolor{black!55}{\textit{200K}} & \textcolor{black!55}{\textit{1.37M}} & \textcolor{black!55}{\textit{1.25M}} & \textcolor{black!55}{\textit{983K}} & \textcolor{black!55}{\textit{591K}} & \textcolor{black!55}{\textit{867K}} & \textcolor{black!55}{\textit{228K}} & \textcolor{black!55}{\textit{--}} \\
\textcolor{black!55}{\textit{R@1 (\%)}} & \textcolor{black!55}{\textit{99.0}} & \textcolor{black!55}{\textit{86.5}} & \textcolor{black!55}{\textit{--}} & \textcolor{black!55}{\textit{94.1}} & \textcolor{black!55}{\textit{93.5}} & \textcolor{black!55}{\textit{99.5}} & \textcolor{black!55}{\textit{88.3}} & \textcolor{black!55}{\textit{32.2}} & \textcolor{black!55}{\textit{75.2}} & \textcolor{black!55}{\textit{78.1}} & \textcolor{black!55}{\textit{94.7}} & \textcolor{black!55}{\textit{--}} \\
\midrule
\refrows{1}
In-context & 62.0 & 28.9 & 45.5 & 70.0 & 63.0 & 42.6 & 47.2 & 14.6 & 31.9 & 16.0 & 26.8 & 39.0 \\
\midrule
\textit{PRAG-style} & 39.7 & 7.4 & 23.6 & 51.5 & 45.0 & 40.6 & 46.1 & 12.9 & 11.0 & 7.6 & 8.1 & 27.9 \\[-1.5pt]
\vsrow{Seq} & \dlt{8.4}\dmk{{*\dagger}} & \dlt{20.1}\dmk{{*\dagger}} & \dlt{14.2}\dmk{*} & \dlt{17.4}\dmk{{*\dagger}} & \dlt{10.0}\dmk{{*\dagger}} & \dlt{17.1}\dmk{{*\dagger}} & \dlt{14.5}\dmk{{*\dagger}} & \dlt{11.7}\dmk{{*\dagger}} & \dlt{18.2}\dmk{{*\dagger}} & \dlt{13.1}\dmk{{*\dagger}} & \dlt{24.4}\dmk{{*\dagger}} & \dlt{15.8}\dmk{*} \\[-1.5pt]
\vsrow{Sim} & \dlt{18.1}\dmk{{*\dagger}} & \dlt{22.2}\dmk{{*\dagger}} & \dlt{20.1}\dmk{*} & \dlt{18.4}\dmk{{*\dagger}} & \dlt{12.7}\dmk{{*\dagger}} & \dlt{23.2}\dmk{{*\dagger}} & \dlt{11.3}\dmk{{*\dagger}} & \dlt{11.2}\dmk{{*\dagger}} & \dlt{19.1}\dmk{{*\dagger}} & \dlt{11.1}\dmk{{*\dagger}} & \dlt{25.2}\dmk{{*\dagger}} & \dlt{16.5}\dmk{*} \\[2pt]
\textit{SEAL-style} & 43.8 & 21.3 & 32.6 & 67.3 & 50.2 & 54.2 & 55.3 & 23.6 & 27.1 & 18.6 & 30.2 & 40.8 \\[-1.5pt]
\vsrow{Seq} & \dlt{4.3} & \dlt{6.2}\dmk{{*\dagger}} & \dlt{5.2}\dmk{*} & \dlt{1.6} & \dlt{4.8} & \dlt{3.5} & \dlt{5.3}\dmk{{*\dagger}} & \dlt{1.0} & \dlt{2.1} & \dlt{2.1}\dmk{*} & \dlt{2.3}\dmk{*} & \dlt{2.9}\dmk{*} \\[-1.5pt]
\vsrow{Sim} & \dlt{14.0}\dmk{{*\dagger}} & \dlt{8.3}\dmk{{*\dagger}} & \dlt{11.1}\dmk{*} & \dlt{2.6} & \dlt{7.5}\dmk{*} & \dlt{9.6}\dmk{{*\dagger}} & \dlt{2.1} & \dlt{0.5} & \dlt{3.0} & \dlt{0.1} & \dlt{3.1}\dmk{{*\dagger}} & \dlt{3.6}\dmk{*} \\[2pt]
\textit{PaST-style} & 48.0 & 20.8 & 34.4 & 61.2 & 51.0 & 55.3 & 57.8 & 23.0 & 19.8 & 16.4 & 22.0 & 38.3 \\[-1.5pt]
\vsrow{Seq} & \dlt{0.1} & \dlt{6.7}\dmk{{*\dagger}} & \dlt{3.4} & \dlt{7.7}\dmk{{*\dagger}} & \dlt{4.0} & \dlt{2.4} & \dlt{2.8} & \dlt{1.6} & \dlt{9.4}\dmk{{*\dagger}} & \dlt{4.3}\dmk{{*\dagger}} & \dlt{10.5}\dmk{{*\dagger}} & \dlt{5.4}\dmk{*} \\[-1.5pt]
\vsrow{Sim} & \dlt{9.8}\dmk{{*\dagger}} & \dlt{8.8}\dmk{{*\dagger}} & \dlt{9.3}\dmk{*} & \dlt{8.7}\dmk{{*\dagger}} & \dlt{6.7}\dmk{*} & \dlt{8.5}\dmk{{*\dagger}} & \dltp{0.4} & \dlt{1.1} & \dlt{10.3}\dmk{{*\dagger}} & \dlt{2.3}\dmk{*} & \dlt{11.3}\dmk{{*\dagger}} & \dlt{6.1}\dmk{*} \\
\midrule
RoDeL-Seq & 48.1 & 27.5 & 37.8 & 68.9 & 55.0 & 57.7 & \textbf{60.6} & \textbf{24.6} & 29.2 & \textbf{20.7} & 32.5 & 43.7 \\[-1.5pt]
{\scriptsize\textcolor{black!55}{\textit{vs.\ oracle}}} & {\scriptsize\textcolor{black!55}{\textit{$-$0.2}}} & {\scriptsize\textcolor{black!55}{\textit{$-$0.2}}} & {\scriptsize\textcolor{black!55}{\textit{$-$0.2}}} & {\scriptsize\textcolor{black!55}{\textit{$-$2.7}}} & {\scriptsize\textcolor{black!55}{\textit{$-$1.8}}} & {\scriptsize\textcolor{black!55}{\textit{0.0}}} & {\scriptsize\textcolor{black!55}{\textit{$-$1.3}}} & {\scriptsize\textcolor{black!55}{\textit{$-$19.5}}} & {\scriptsize\textcolor{black!55}{\textit{$-$1.3}}} & {\scriptsize\textcolor{black!55}{\textit{$-$3.3}}} & {\scriptsize\textcolor{black!55}{\textit{$-$1.6}}} & {\scriptsize\textcolor{black!55}{\textit{$-$3.9}}} \\[2pt]
RoDeL-Sim & \textbf{57.8} & \textbf{29.6} & \textbf{43.7} & \textbf{69.9} & \textbf{57.7} & \textbf{63.8} & 57.4 & 24.1 & \textbf{30.1} & 18.7 & \textbf{33.3} & \textbf{44.4} \\[-1.5pt]
{\scriptsize\textcolor{black!55}{\textit{vs.\ oracle}}} & {\scriptsize\textcolor{black!55}{\textit{$-$0.3}}} & {\scriptsize\textcolor{black!55}{\textit{$-$0.7}}} & {\scriptsize\textcolor{black!55}{\textit{$-$0.5}}} & {\scriptsize\textcolor{black!55}{\textit{$-$2.5}}} & {\scriptsize\textcolor{black!55}{\textit{$-$1.4}}} & {\scriptsize\textcolor{black!55}{\textit{0.0}}} & {\scriptsize\textcolor{black!55}{\textit{$-$1.9}}} & {\scriptsize\textcolor{black!55}{\textit{$-$17.2}}} & {\scriptsize\textcolor{black!55}{\textit{$-$0.1}}} & {\scriptsize\textcolor{black!55}{\textit{$-$3.2}}} & {\scriptsize\textcolor{black!55}{\textit{$-$1.4}}} & {\scriptsize\textcolor{black!55}{\textit{$-$3.4}}} \\
\bottomrule
\end{tabular}
\end{table}

\Cref{tab:retrieval} reports the top-1 retrieval accuracy (R@1) on each dataset and the F1 of each method with the retrieved memory adapters.
Both variants still have a higher average F1 than every baseline: 37.8 in domain and 43.7 out of domain for RoDeL-Seq, and 43.7 and 44.4 for RoDeL-Sim. On every dataset, the highest F1 among the five methods again belongs to one of the two variants. Out of domain, the leads over PRAG-style, SEAL-style, and PaST-style are 15.8, 2.9, and 5.4 points for RoDeL-Seq and 16.5, 3.6, and 6.1 points for RoDeL-Sim, all with $p<0.05$. In domain, every lead has $p<0.05$ except that of RoDeL-Seq over PaST-style (3.4 points), as in \Cref{tab:main}.

With retrieval, both variants stay close to their F1 under \textit{oracle} retrieval in \Cref{tab:main}, where each question reads the memory adapter of its own document. The average falls by 0.2 points in domain and by 3.9 out of domain for RoDeL-Seq, and by 0.5 and 3.4 for RoDeL-Sim. The gap stays small on corpora of over one million tokens. Most of the out-of-domain gap comes from Qasper, where the top-1 document is correct for only 32.2\% of the questions.

\subsection{Ablation Study}
\label{subsec:ablation}

Two single-role variants test whether both roles need training. Like RoDeL-Seq, each of them trains for six epochs with one update per epoch, but all six updates train the same role. The student-only variant trains $\phi_s$, and the base model writes the memory adapters. The teacher-only variant trains $\phi_t$, and the base model reads the memory adapters, both in the teacher reward and at test time. \Cref{fig:abl_perset,fig:abl_avg} show the F1 after each update. They also show RoDeL-Sim. Each of its updates $u_k$ trains both adapters, which takes RoDeL-Seq the two epochs $s_k$ and $t_k$. After $u_k$, each adapter has had $k$ updates, as in RoDeL-Seq after $t_k$ (\Cref{subsec:variants}), so the F1 after $u_k$ is plotted at epoch $2k$.

\begin{figure}[!htb]
\centering
\begin{tikzpicture}
\begin{groupplot}[
  group style={group size=5 by 2, horizontal sep=0.55cm, vertical sep=0.95cm},
  scale only axis, width=2.5cm, height=2.0cm,
  xmin=-0.5, xmax=6.5,
  xtick={0,...,6},
  xticklabels={Fr.,1,2,3,4,5,6},
  axis lines=left, axis line style={-, black!60}, tick style={black!60}, major tick length=2pt,
  tick label style={font=\tiny, inner sep=1pt},
  title style={font=\footnotesize, yshift=-4pt},
  ymajorgrids, grid style={black!22, dashed},
  extra x ticks={0.5}, extra x tick labels={},
  extra x tick style={grid=major, major grid style={black!25, solid}, tick style={draw=none}},
  clip=false,
]
\nextgroupplot[title={HotpotQA (ID)}, ymin=20, ymax=65, ylabel={F1}, ylabel style={font=\scriptsize, yshift=-3pt}, legend to name=abllegend, legend columns=4, reverse legend, legend style={font=\scriptsize, draw=none, /tikz/every even column/.append style={column sep=10pt}}]
\addplot[name path=UTONLY, draw=none, forget plot] table[x=x, y expr=\thisrow{m}+\thisrow{sd}] {
x  m  sd
0  39.50  3.26
1  35.10  3.20
2  26.40  2.60
3  44.50  3.30
4  48.30  3.20
5  44.40  3.20
6  33.00  3.00
};
\addplot[name path=LTONLY, draw=none, forget plot] table[x=x, y expr=\thisrow{m}-\thisrow{sd}] {
x  m  sd
0  39.50  3.26
1  35.10  3.20
2  26.40  2.60
3  44.50  3.30
4  48.30  3.20
5  44.40  3.20
6  33.00  3.00
};
\addplot[ablT, fill opacity=0.10, draw=none, forget plot] fill between[of=UTONLY and LTONLY];
\addplot[name path=USONLY, draw=none, forget plot] table[x=x, y expr=\thisrow{m}+\thisrow{sd}] {
x  m  sd
0  39.50  3.26
1  37.90  3.27
2  44.70  3.20
3  45.60  3.20
4  45.90  3.20
5  46.30  3.20
6  48.80  3.20
};
\addplot[name path=LSONLY, draw=none, forget plot] table[x=x, y expr=\thisrow{m}-\thisrow{sd}] {
x  m  sd
0  39.50  3.26
1  37.90  3.27
2  44.70  3.20
3  45.60  3.20
4  45.90  3.20
5  46.30  3.20
6  48.80  3.20
};
\addplot[stuV, fill opacity=0.10, draw=none, forget plot] fill between[of=USONLY and LSONLY];
\addplot[name path=URoDeL, draw=none, forget plot] table[x=x, y expr=\thisrow{m}+\thisrow{sd}] {
x  m  sd
0  39.50  3.26
1  37.90  3.27
2  38.50  3.27
3  42.70  3.19
4  52.40  3.22
5  48.30  3.22
6  48.90  3.22
};
\addplot[name path=LRoDeL, draw=none, forget plot] table[x=x, y expr=\thisrow{m}-\thisrow{sd}] {
x  m  sd
0  39.50  3.26
1  37.90  3.27
2  38.50  3.27
3  42.70  3.19
4  52.40  3.22
5  48.30  3.22
6  48.90  3.22
};
\addplot[lineO, fill opacity=0.10, draw=none, forget plot] fill between[of=URoDeL and LRoDeL];
\addplot[name path=USim, draw=none, forget plot] table[x=x, y expr=\thisrow{m}+\thisrow{sd}] {
x  m  sd
0  39.50  3.26
2  38.10  3.20
4  48.40  3.21
6  58.10  3.17
};
\addplot[name path=LSim, draw=none, forget plot] table[x=x, y expr=\thisrow{m}-\thisrow{sd}] {
x  m  sd
0  39.50  3.26
2  38.10  3.20
4  48.40  3.21
6  58.10  3.17
};
\addplot[simR, fill opacity=0.10, draw=none, forget plot] fill between[of=USim and LSim];
\addplot[ablT, line width=0.8pt, mark=triangle*, mark size=2.0pt, mark options={fill=ablT, draw=white, line width=0.3pt}] table[x=x, y=m] {
x  m
0  39.50
1  35.10
2  26.40
3  44.50
4  48.30
5  44.40
6  33.00
};
\addlegendentry{Teacher only}
\addplot[stuV, line width=0.8pt, mark=square*, mark size=1.5pt, mark options={fill=stuV, draw=white, line width=0.3pt}] table[x=x, y=m] {
x  m
0  39.50
1  37.90
2  44.70
3  45.60
4  45.90
5  46.30
6  48.80
};
\addlegendentry{Student only}
\addplot[simR, line width=0.8pt, mark=diamond*, mark size=1.9pt, mark options={fill=simR, draw=white, line width=0.3pt}] table[x=x, y=m] {
x  m
0  39.50
2  38.10
4  48.40
6  58.10
};
\addlegendentry{RoDeL-Sim}
\addplot[lineO, line width=0.8pt, mark=*, mark size=1.6pt, mark options={fill=lineO, draw=white, line width=0.3pt}] table[x=x, y=m] {
x  m
0  39.50
1  37.90
2  38.50
3  42.70
4  52.40
5  48.30
6  48.90
};
\addlegendentry{RoDeL-Seq}
\nextgroupplot[title={MuSiQue (ID)}, ymin=0, ymax=35]
\addplot[name path=UTONLY, draw=none, forget plot] table[x=x, y expr=\thisrow{m}+\thisrow{sd}] {
x  m  sd
0  7.30  1.67
1  4.90  1.50
2  13.60  1.90
3  9.70  1.90
4  16.60  2.30
5  13.80  2.20
6  15.40  2.00
};
\addplot[name path=LTONLY, draw=none, forget plot] table[x=x, y expr=\thisrow{m}-\thisrow{sd}] {
x  m  sd
0  7.30  1.67
1  4.90  1.50
2  13.60  1.90
3  9.70  1.90
4  16.60  2.30
5  13.80  2.20
6  15.40  2.00
};
\addplot[ablT, fill opacity=0.10, draw=none, forget plot] fill between[of=UTONLY and LTONLY];
\addplot[name path=USONLY, draw=none, forget plot] table[x=x, y expr=\thisrow{m}+\thisrow{sd}] {
x  m  sd
0  7.30  1.67
1  11.10  2.02
2  19.00  2.50
3  22.20  2.60
4  21.90  2.60
5  21.90  2.60
6  20.40  2.50
};
\addplot[name path=LSONLY, draw=none, forget plot] table[x=x, y expr=\thisrow{m}-\thisrow{sd}] {
x  m  sd
0  7.30  1.67
1  11.10  2.02
2  19.00  2.50
3  22.20  2.60
4  21.90  2.60
5  21.90  2.60
6  20.40  2.50
};
\addplot[stuV, fill opacity=0.10, draw=none, forget plot] fill between[of=USONLY and LSONLY];
\addplot[name path=URoDeL, draw=none, forget plot] table[x=x, y expr=\thisrow{m}+\thisrow{sd}] {
x  m  sd
0  7.30  1.67
1  11.10  2.02
2  14.40  2.31
3  23.10  2.68
4  22.20  2.56
5  27.70  2.77
6  24.90  2.54
};
\addplot[name path=LRoDeL, draw=none, forget plot] table[x=x, y expr=\thisrow{m}-\thisrow{sd}] {
x  m  sd
0  7.30  1.67
1  11.10  2.02
2  14.40  2.31
3  23.10  2.68
4  22.20  2.56
5  27.70  2.77
6  24.90  2.54
};
\addplot[lineO, fill opacity=0.10, draw=none, forget plot] fill between[of=URoDeL and LRoDeL];
\addplot[name path=USim, draw=none, forget plot] table[x=x, y expr=\thisrow{m}+\thisrow{sd}] {
x  m  sd
0  7.30  1.67
2  12.30  2.01
4  31.90  3.00
6  30.30  2.80
};
\addplot[name path=LSim, draw=none, forget plot] table[x=x, y expr=\thisrow{m}-\thisrow{sd}] {
x  m  sd
0  7.30  1.67
2  12.30  2.01
4  31.90  3.00
6  30.30  2.80
};
\addplot[simR, fill opacity=0.10, draw=none, forget plot] fill between[of=USim and LSim];
\addplot[ablT, line width=0.8pt, mark=triangle*, mark size=2.0pt, mark options={fill=ablT, draw=white, line width=0.3pt}] table[x=x, y=m] {
x  m
0  7.30
1  4.90
2  13.60
3  9.70
4  16.60
5  13.80
6  15.40
};
\addplot[stuV, line width=0.8pt, mark=square*, mark size=1.5pt, mark options={fill=stuV, draw=white, line width=0.3pt}] table[x=x, y=m] {
x  m
0  7.30
1  11.10
2  19.00
3  22.20
4  21.90
5  21.90
6  20.40
};
\addplot[simR, line width=0.8pt, mark=diamond*, mark size=1.9pt, mark options={fill=simR, draw=white, line width=0.3pt}] table[x=x, y=m] {
x  m
0  7.30
2  12.30
4  31.90
6  30.30
};
\addplot[lineO, line width=0.8pt, mark=*, mark size=1.6pt, mark options={fill=lineO, draw=white, line width=0.3pt}] table[x=x, y=m] {
x  m
0  7.30
1  11.10
2  14.40
3  23.10
4  22.20
5  27.70
6  24.90
};
\nextgroupplot[title={SQuAD}, ymin=45, ymax=85]
\addplot[name path=UTONLY, draw=none, forget plot] table[x=x, y expr=\thisrow{m}+\thisrow{sd}] {
x  m  sd
0  53.60  1.86
1  53.30  2.00
2  50.40  1.50
3  73.30  1.50
4  71.90  1.40
5  79.50  1.40
6  52.20  1.50
};
\addplot[name path=LTONLY, draw=none, forget plot] table[x=x, y expr=\thisrow{m}-\thisrow{sd}] {
x  m  sd
0  53.60  1.86
1  53.30  2.00
2  50.40  1.50
3  73.30  1.50
4  71.90  1.40
5  79.50  1.40
6  52.20  1.50
};
\addplot[ablT, fill opacity=0.10, draw=none, forget plot] fill between[of=UTONLY and LTONLY];
\addplot[name path=USONLY, draw=none, forget plot] table[x=x, y expr=\thisrow{m}+\thisrow{sd}] {
x  m  sd
0  53.60  1.86
1  56.10  1.76
2  61.00  1.60
3  61.50  1.70
4  61.80  1.70
5  62.00  1.70
6  63.10  1.70
};
\addplot[name path=LSONLY, draw=none, forget plot] table[x=x, y expr=\thisrow{m}-\thisrow{sd}] {
x  m  sd
0  53.60  1.86
1  56.10  1.76
2  61.00  1.60
3  61.50  1.70
4  61.80  1.70
5  62.00  1.70
6  63.10  1.70
};
\addplot[stuV, fill opacity=0.10, draw=none, forget plot] fill between[of=USONLY and LSONLY];
\addplot[name path=URoDeL, draw=none, forget plot] table[x=x, y expr=\thisrow{m}+\thisrow{sd}] {
x  m  sd
0  53.60  1.86
1  56.10  1.76
2  57.50  1.74
3  62.00  1.73
4  71.60  1.54
5  71.40  1.45
6  67.60  1.36
};
\addplot[name path=LRoDeL, draw=none, forget plot] table[x=x, y expr=\thisrow{m}-\thisrow{sd}] {
x  m  sd
0  53.60  1.86
1  56.10  1.76
2  57.50  1.74
3  62.00  1.73
4  71.60  1.54
5  71.40  1.45
6  67.60  1.36
};
\addplot[lineO, fill opacity=0.10, draw=none, forget plot] fill between[of=URoDeL and LRoDeL];
\addplot[name path=USim, draw=none, forget plot] table[x=x, y expr=\thisrow{m}+\thisrow{sd}] {
x  m  sd
0  53.60  1.86
2  55.40  1.72
4  72.40  1.27
6  69.80  1.33
};
\addplot[name path=LSim, draw=none, forget plot] table[x=x, y expr=\thisrow{m}-\thisrow{sd}] {
x  m  sd
0  53.60  1.86
2  55.40  1.72
4  72.40  1.27
6  69.80  1.33
};
\addplot[simR, fill opacity=0.10, draw=none, forget plot] fill between[of=USim and LSim];
\addplot[ablT, line width=0.8pt, mark=triangle*, mark size=2.0pt, mark options={fill=ablT, draw=white, line width=0.3pt}] table[x=x, y=m] {
x  m
0  53.60
1  53.30
2  50.40
3  73.30
4  71.90
5  79.50
6  52.20
};
\addplot[stuV, line width=0.8pt, mark=square*, mark size=1.5pt, mark options={fill=stuV, draw=white, line width=0.3pt}] table[x=x, y=m] {
x  m
0  53.60
1  56.10
2  61.00
3  61.50
4  61.80
5  62.00
6  63.10
};
\addplot[simR, line width=0.8pt, mark=diamond*, mark size=1.9pt, mark options={fill=simR, draw=white, line width=0.3pt}] table[x=x, y=m] {
x  m
0  53.60
2  55.40
4  72.40
6  69.80
};
\addplot[lineO, line width=0.8pt, mark=*, mark size=1.6pt, mark options={fill=lineO, draw=white, line width=0.3pt}] table[x=x, y=m] {
x  m
0  53.60
1  56.10
2  57.50
3  62.00
4  71.60
5  71.40
6  67.60
};
\nextgroupplot[title={2Wiki}, ymin=25, ymax=65]
\addplot[name path=UTONLY, draw=none, forget plot] table[x=x, y expr=\thisrow{m}+\thisrow{sd}] {
x  m  sd
0  47.00  3.44
1  38.50  3.30
2  30.00  2.60
3  47.10  3.40
4  46.70  3.30
5  55.70  3.20
6  36.50  3.10
};
\addplot[name path=LTONLY, draw=none, forget plot] table[x=x, y expr=\thisrow{m}-\thisrow{sd}] {
x  m  sd
0  47.00  3.44
1  38.50  3.30
2  30.00  2.60
3  47.10  3.40
4  46.70  3.30
5  55.70  3.20
6  36.50  3.10
};
\addplot[ablT, fill opacity=0.10, draw=none, forget plot] fill between[of=UTONLY and LTONLY];
\addplot[name path=USONLY, draw=none, forget plot] table[x=x, y expr=\thisrow{m}+\thisrow{sd}] {
x  m  sd
0  47.00  3.44
1  42.80  3.39
2  47.40  3.30
3  48.60  3.30
4  48.60  3.30
5  50.40  3.40
6  52.00  3.30
};
\addplot[name path=LSONLY, draw=none, forget plot] table[x=x, y expr=\thisrow{m}-\thisrow{sd}] {
x  m  sd
0  47.00  3.44
1  42.80  3.39
2  47.40  3.30
3  48.60  3.30
4  48.60  3.30
5  50.40  3.40
6  52.00  3.30
};
\addplot[stuV, fill opacity=0.10, draw=none, forget plot] fill between[of=USONLY and LSONLY];
\addplot[name path=URoDeL, draw=none, forget plot] table[x=x, y expr=\thisrow{m}+\thisrow{sd}] {
x  m  sd
0  47.00  3.44
1  42.80  3.39
2  49.30  3.42
3  55.90  3.30
4  56.80  3.18
5  57.40  3.17
6  48.70  3.25
};
\addplot[name path=LRoDeL, draw=none, forget plot] table[x=x, y expr=\thisrow{m}-\thisrow{sd}] {
x  m  sd
0  47.00  3.44
1  42.80  3.39
2  49.30  3.42
3  55.90  3.30
4  56.80  3.18
5  57.40  3.17
6  48.70  3.25
};
\addplot[lineO, fill opacity=0.10, draw=none, forget plot] fill between[of=URoDeL and LRoDeL];
\addplot[name path=USim, draw=none, forget plot] table[x=x, y expr=\thisrow{m}+\thisrow{sd}] {
x  m  sd
0  47.00  3.44
2  35.20  3.19
4  59.10  3.20
6  55.50  3.19
};
\addplot[name path=LSim, draw=none, forget plot] table[x=x, y expr=\thisrow{m}-\thisrow{sd}] {
x  m  sd
0  47.00  3.44
2  35.20  3.19
4  59.10  3.20
6  55.50  3.19
};
\addplot[simR, fill opacity=0.10, draw=none, forget plot] fill between[of=USim and LSim];
\addplot[ablT, line width=0.8pt, mark=triangle*, mark size=2.0pt, mark options={fill=ablT, draw=white, line width=0.3pt}] table[x=x, y=m] {
x  m
0  47.00
1  38.50
2  30.00
3  47.10
4  46.70
5  55.70
6  36.50
};
\addplot[stuV, line width=0.8pt, mark=square*, mark size=1.5pt, mark options={fill=stuV, draw=white, line width=0.3pt}] table[x=x, y=m] {
x  m
0  47.00
1  42.80
2  47.40
3  48.60
4  48.60
5  50.40
6  52.00
};
\addplot[simR, line width=0.8pt, mark=diamond*, mark size=1.9pt, mark options={fill=simR, draw=white, line width=0.3pt}] table[x=x, y=m] {
x  m
0  47.00
2  35.20
4  59.10
6  55.50
};
\addplot[lineO, line width=0.8pt, mark=*, mark size=1.6pt, mark options={fill=lineO, draw=white, line width=0.3pt}] table[x=x, y=m] {
x  m
0  47.00
1  42.80
2  49.30
3  55.90
4  56.80
5  57.40
6  48.70
};
\nextgroupplot[title={NQ}, ymin=30, ymax=70]
\addplot[name path=UTONLY, draw=none, forget plot] table[x=x, y expr=\thisrow{m}+\thisrow{sd}] {
x  m  sd
0  40.60  3.25
1  38.00  3.20
2  45.90  2.90
3  49.70  3.10
4  50.00  2.90
5  57.00  3.00
6  50.00  3.00
};
\addplot[name path=LTONLY, draw=none, forget plot] table[x=x, y expr=\thisrow{m}-\thisrow{sd}] {
x  m  sd
0  40.60  3.25
1  38.00  3.20
2  45.90  2.90
3  49.70  3.10
4  50.00  2.90
5  57.00  3.00
6  50.00  3.00
};
\addplot[ablT, fill opacity=0.10, draw=none, forget plot] fill between[of=UTONLY and LTONLY];
\addplot[name path=USONLY, draw=none, forget plot] table[x=x, y expr=\thisrow{m}+\thisrow{sd}] {
x  m  sd
0  40.60  3.25
1  43.00  3.21
2  52.40  3.00
3  54.10  3.00
4  51.40  3.20
5  53.10  3.10
6  55.50  3.10
};
\addplot[name path=LSONLY, draw=none, forget plot] table[x=x, y expr=\thisrow{m}-\thisrow{sd}] {
x  m  sd
0  40.60  3.25
1  43.00  3.21
2  52.40  3.00
3  54.10  3.00
4  51.40  3.20
5  53.10  3.10
6  55.50  3.10
};
\addplot[stuV, fill opacity=0.10, draw=none, forget plot] fill between[of=USONLY and LSONLY];
\addplot[name path=URoDeL, draw=none, forget plot] table[x=x, y expr=\thisrow{m}+\thisrow{sd}] {
x  m  sd
0  40.60  3.25
1  43.00  3.21
2  41.70  3.27
3  55.90  3.05
4  57.70  3.02
5  55.20  3.03
6  53.90  2.91
};
\addplot[name path=LRoDeL, draw=none, forget plot] table[x=x, y expr=\thisrow{m}-\thisrow{sd}] {
x  m  sd
0  40.60  3.25
1  43.00  3.21
2  41.70  3.27
3  55.90  3.05
4  57.70  3.02
5  55.20  3.03
6  53.90  2.91
};
\addplot[lineO, fill opacity=0.10, draw=none, forget plot] fill between[of=URoDeL and LRoDeL];
\addplot[name path=USim, draw=none, forget plot] table[x=x, y expr=\thisrow{m}+\thisrow{sd}] {
x  m  sd
0  40.60  3.25
2  53.30  3.00
4  63.80  2.85
6  58.80  2.81
};
\addplot[name path=LSim, draw=none, forget plot] table[x=x, y expr=\thisrow{m}-\thisrow{sd}] {
x  m  sd
0  40.60  3.25
2  53.30  3.00
4  63.80  2.85
6  58.80  2.81
};
\addplot[simR, fill opacity=0.10, draw=none, forget plot] fill between[of=USim and LSim];
\addplot[ablT, line width=0.8pt, mark=triangle*, mark size=2.0pt, mark options={fill=ablT, draw=white, line width=0.3pt}] table[x=x, y=m] {
x  m
0  40.60
1  38.00
2  45.90
3  49.70
4  50.00
5  57.00
6  50.00
};
\addplot[stuV, line width=0.8pt, mark=square*, mark size=1.5pt, mark options={fill=stuV, draw=white, line width=0.3pt}] table[x=x, y=m] {
x  m
0  40.60
1  43.00
2  52.40
3  54.10
4  51.40
5  53.10
6  55.50
};
\addplot[simR, line width=0.8pt, mark=diamond*, mark size=1.9pt, mark options={fill=simR, draw=white, line width=0.3pt}] table[x=x, y=m] {
x  m
0  40.60
2  53.30
4  63.80
6  58.80
};
\addplot[lineO, line width=0.8pt, mark=*, mark size=1.6pt, mark options={fill=lineO, draw=white, line width=0.3pt}] table[x=x, y=m] {
x  m
0  40.60
1  43.00
2  41.70
3  55.90
4  57.70
5  55.20
6  53.90
};
\nextgroupplot[title={TriviaQA}, ymin=40, ymax=70, ylabel={F1}, ylabel style={font=\scriptsize, yshift=-3pt}]
\addplot[name path=UTONLY, draw=none, forget plot] table[x=x, y expr=\thisrow{m}+\thisrow{sd}] {
x  m  sd
0  47.90  2.50
1  45.40  2.50
2  46.60  2.00
3  49.40  2.30
4  53.00  2.20
5  56.40  2.40
6  52.10  2.20
};
\addplot[name path=LTONLY, draw=none, forget plot] table[x=x, y expr=\thisrow{m}-\thisrow{sd}] {
x  m  sd
0  47.90  2.50
1  45.40  2.50
2  46.60  2.00
3  49.40  2.30
4  53.00  2.20
5  56.40  2.40
6  52.10  2.20
};
\addplot[ablT, fill opacity=0.10, draw=none, forget plot] fill between[of=UTONLY and LTONLY];
\addplot[name path=USONLY, draw=none, forget plot] table[x=x, y expr=\thisrow{m}+\thisrow{sd}] {
x  m  sd
0  47.90  2.50
1  50.50  2.49
2  57.70  2.40
3  62.00  2.40
4  59.20  2.50
5  58.70  2.40
6  60.50  2.60
};
\addplot[name path=LSONLY, draw=none, forget plot] table[x=x, y expr=\thisrow{m}-\thisrow{sd}] {
x  m  sd
0  47.90  2.50
1  50.50  2.49
2  57.70  2.40
3  62.00  2.40
4  59.20  2.50
5  58.70  2.40
6  60.50  2.60
};
\addplot[stuV, fill opacity=0.10, draw=none, forget plot] fill between[of=USONLY and LSONLY];
\addplot[name path=URoDeL, draw=none, forget plot] table[x=x, y expr=\thisrow{m}+\thisrow{sd}] {
x  m  sd
0  47.90  2.50
1  50.50  2.49
2  52.20  2.30
3  57.60  2.42
4  61.90  2.24
5  59.40  2.36
6  57.50  2.25
};
\addplot[name path=LRoDeL, draw=none, forget plot] table[x=x, y expr=\thisrow{m}-\thisrow{sd}] {
x  m  sd
0  47.90  2.50
1  50.50  2.49
2  52.20  2.30
3  57.60  2.42
4  61.90  2.24
5  59.40  2.36
6  57.50  2.25
};
\addplot[lineO, fill opacity=0.10, draw=none, forget plot] fill between[of=URoDeL and LRoDeL];
\addplot[name path=USim, draw=none, forget plot] table[x=x, y expr=\thisrow{m}+\thisrow{sd}] {
x  m  sd
0  47.90  2.50
2  55.10  2.27
4  59.30  2.44
6  61.30  2.23
};
\addplot[name path=LSim, draw=none, forget plot] table[x=x, y expr=\thisrow{m}-\thisrow{sd}] {
x  m  sd
0  47.90  2.50
2  55.10  2.27
4  59.30  2.44
6  61.30  2.23
};
\addplot[simR, fill opacity=0.10, draw=none, forget plot] fill between[of=USim and LSim];
\addplot[ablT, line width=0.8pt, mark=triangle*, mark size=2.0pt, mark options={fill=ablT, draw=white, line width=0.3pt}] table[x=x, y=m] {
x  m
0  47.90
1  45.40
2  46.60
3  49.40
4  53.00
5  56.40
6  52.10
};
\addplot[stuV, line width=0.8pt, mark=square*, mark size=1.5pt, mark options={fill=stuV, draw=white, line width=0.3pt}] table[x=x, y=m] {
x  m
0  47.90
1  50.50
2  57.70
3  62.00
4  59.20
5  58.70
6  60.50
};
\addplot[simR, line width=0.8pt, mark=diamond*, mark size=1.9pt, mark options={fill=simR, draw=white, line width=0.3pt}] table[x=x, y=m] {
x  m
0  47.90
2  55.10
4  59.30
6  61.30
};
\addplot[lineO, line width=0.8pt, mark=*, mark size=1.6pt, mark options={fill=lineO, draw=white, line width=0.3pt}] table[x=x, y=m] {
x  m
0  47.90
1  50.50
2  52.20
3  57.60
4  61.90
5  59.40
6  57.50
};
\nextgroupplot[title={Qasper}, ymin=10, ymax=50]
\addplot[name path=UTONLY, draw=none, forget plot] table[x=x, y expr=\thisrow{m}+\thisrow{sd}] {
x  m  sd
0  20.80  1.79
1  17.60  1.60
2  33.70  1.60
3  28.80  1.70
4  30.30  1.40
5  40.20  1.80
6  39.30  1.70
};
\addplot[name path=LTONLY, draw=none, forget plot] table[x=x, y expr=\thisrow{m}-\thisrow{sd}] {
x  m  sd
0  20.80  1.79
1  17.60  1.60
2  33.70  1.60
3  28.80  1.70
4  30.30  1.40
5  40.20  1.80
6  39.30  1.70
};
\addplot[ablT, fill opacity=0.10, draw=none, forget plot] fill between[of=UTONLY and LTONLY];
\addplot[name path=USONLY, draw=none, forget plot] table[x=x, y expr=\thisrow{m}+\thisrow{sd}] {
x  m  sd
0  20.80  1.79
1  24.10  1.86
2  32.70  1.80
3  32.50  1.80
4  30.90  1.90
5  32.20  1.90
6  32.80  1.80
};
\addplot[name path=LSONLY, draw=none, forget plot] table[x=x, y expr=\thisrow{m}-\thisrow{sd}] {
x  m  sd
0  20.80  1.79
1  24.10  1.86
2  32.70  1.80
3  32.50  1.80
4  30.90  1.90
5  32.20  1.90
6  32.80  1.80
};
\addplot[stuV, fill opacity=0.10, draw=none, forget plot] fill between[of=USONLY and LSONLY];
\addplot[name path=URoDeL, draw=none, forget plot] table[x=x, y expr=\thisrow{m}+\thisrow{sd}] {
x  m  sd
0  20.80  1.79
1  24.10  1.86
2  25.30  1.83
3  32.10  1.75
4  44.10  1.89
5  42.00  1.65
6  42.10  1.80
};
\addplot[name path=LRoDeL, draw=none, forget plot] table[x=x, y expr=\thisrow{m}-\thisrow{sd}] {
x  m  sd
0  20.80  1.79
1  24.10  1.86
2  25.30  1.83
3  32.10  1.75
4  44.10  1.89
5  42.00  1.65
6  42.10  1.80
};
\addplot[lineO, fill opacity=0.10, draw=none, forget plot] fill between[of=URoDeL and LRoDeL];
\addplot[name path=USim, draw=none, forget plot] table[x=x, y expr=\thisrow{m}+\thisrow{sd}] {
x  m  sd
0  20.80  1.79
2  35.90  1.76
4  41.30  1.76
6  44.10  1.74
};
\addplot[name path=LSim, draw=none, forget plot] table[x=x, y expr=\thisrow{m}-\thisrow{sd}] {
x  m  sd
0  20.80  1.79
2  35.90  1.76
4  41.30  1.76
6  44.10  1.74
};
\addplot[simR, fill opacity=0.10, draw=none, forget plot] fill between[of=USim and LSim];
\addplot[ablT, line width=0.8pt, mark=triangle*, mark size=2.0pt, mark options={fill=ablT, draw=white, line width=0.3pt}] table[x=x, y=m] {
x  m
0  20.80
1  17.60
2  33.70
3  28.80
4  30.30
5  40.20
6  39.30
};
\addplot[stuV, line width=0.8pt, mark=square*, mark size=1.5pt, mark options={fill=stuV, draw=white, line width=0.3pt}] table[x=x, y=m] {
x  m
0  20.80
1  24.10
2  32.70
3  32.50
4  30.90
5  32.20
6  32.80
};
\addplot[simR, line width=0.8pt, mark=diamond*, mark size=1.9pt, mark options={fill=simR, draw=white, line width=0.3pt}] table[x=x, y=m] {
x  m
0  20.80
2  35.90
4  41.30
6  44.10
};
\addplot[lineO, line width=0.8pt, mark=*, mark size=1.6pt, mark options={fill=lineO, draw=white, line width=0.3pt}] table[x=x, y=m] {
x  m
0  20.80
1  24.10
2  25.30
3  32.10
4  44.10
5  42.00
6  42.10
};
\nextgroupplot[title={MultiFieldQA}, ymin=5, ymax=40]
\addplot[name path=UTONLY, draw=none, forget plot] table[x=x, y expr=\thisrow{m}+\thisrow{sd}] {
x  m  sd
0  13.50  2.58
1  16.60  2.90
2  33.00  3.40
3  19.90  2.70
4  28.00  3.20
5  29.80  3.80
6  32.00  2.90
};
\addplot[name path=LTONLY, draw=none, forget plot] table[x=x, y expr=\thisrow{m}-\thisrow{sd}] {
x  m  sd
0  13.50  2.58
1  16.60  2.90
2  33.00  3.40
3  19.90  2.70
4  28.00  3.20
5  29.80  3.80
6  32.00  2.90
};
\addplot[ablT, fill opacity=0.10, draw=none, forget plot] fill between[of=UTONLY and LTONLY];
\addplot[name path=USONLY, draw=none, forget plot] table[x=x, y expr=\thisrow{m}+\thisrow{sd}] {
x  m  sd
0  13.50  2.58
1  18.00  3.08
2  24.20  3.00
3  22.90  2.90
4  25.40  3.60
5  21.00  3.00
6  22.90  2.80
};
\addplot[name path=LSONLY, draw=none, forget plot] table[x=x, y expr=\thisrow{m}-\thisrow{sd}] {
x  m  sd
0  13.50  2.58
1  18.00  3.08
2  24.20  3.00
3  22.90  2.90
4  25.40  3.60
5  21.00  3.00
6  22.90  2.80
};
\addplot[stuV, fill opacity=0.10, draw=none, forget plot] fill between[of=USONLY and LSONLY];
\addplot[name path=URoDeL, draw=none, forget plot] table[x=x, y expr=\thisrow{m}+\thisrow{sd}] {
x  m  sd
0  13.50  2.58
1  18.00  3.08
2  15.60  2.87
3  22.60  2.75
4  30.50  3.27
5  29.30  3.26
6  28.50  3.21
};
\addplot[name path=LRoDeL, draw=none, forget plot] table[x=x, y expr=\thisrow{m}-\thisrow{sd}] {
x  m  sd
0  13.50  2.58
1  18.00  3.08
2  15.60  2.87
3  22.60  2.75
4  30.50  3.27
5  29.30  3.26
6  28.50  3.21
};
\addplot[lineO, fill opacity=0.10, draw=none, forget plot] fill between[of=URoDeL and LRoDeL];
\addplot[name path=USim, draw=none, forget plot] table[x=x, y expr=\thisrow{m}+\thisrow{sd}] {
x  m  sd
0  13.50  2.58
2  21.30  2.96
4  30.20  2.90
6  27.30  2.90
};
\addplot[name path=LSim, draw=none, forget plot] table[x=x, y expr=\thisrow{m}-\thisrow{sd}] {
x  m  sd
0  13.50  2.58
2  21.30  2.96
4  30.20  2.90
6  27.30  2.90
};
\addplot[simR, fill opacity=0.10, draw=none, forget plot] fill between[of=USim and LSim];
\addplot[ablT, line width=0.8pt, mark=triangle*, mark size=2.0pt, mark options={fill=ablT, draw=white, line width=0.3pt}] table[x=x, y=m] {
x  m
0  13.50
1  16.60
2  33.00
3  19.90
4  28.00
5  29.80
6  32.00
};
\addplot[stuV, line width=0.8pt, mark=square*, mark size=1.5pt, mark options={fill=stuV, draw=white, line width=0.3pt}] table[x=x, y=m] {
x  m
0  13.50
1  18.00
2  24.20
3  22.90
4  25.40
5  21.00
6  22.90
};
\addplot[simR, line width=0.8pt, mark=diamond*, mark size=1.9pt, mark options={fill=simR, draw=white, line width=0.3pt}] table[x=x, y=m] {
x  m
0  13.50
2  21.30
4  30.20
6  27.30
};
\addplot[lineO, line width=0.8pt, mark=*, mark size=1.6pt, mark options={fill=lineO, draw=white, line width=0.3pt}] table[x=x, y=m] {
x  m
0  13.50
1  18.00
2  15.60
3  22.60
4  30.50
5  29.30
6  28.50
};
\nextgroupplot[title={NarrativeQA}, ymin=5, ymax=30]
\addplot[name path=UTONLY, draw=none, forget plot] table[x=x, y expr=\thisrow{m}+\thisrow{sd}] {
x  m  sd
0  9.10  0.95
1  8.30  0.90
2  21.50  1.00
3  14.70  1.30
4  18.20  1.40
5  20.30  1.40
6  24.20  1.00
};
\addplot[name path=LTONLY, draw=none, forget plot] table[x=x, y expr=\thisrow{m}-\thisrow{sd}] {
x  m  sd
0  9.10  0.95
1  8.30  0.90
2  21.50  1.00
3  14.70  1.30
4  18.20  1.40
5  20.30  1.40
6  24.20  1.00
};
\addplot[ablT, fill opacity=0.10, draw=none, forget plot] fill between[of=UTONLY and LTONLY];
\addplot[name path=USONLY, draw=none, forget plot] table[x=x, y expr=\thisrow{m}+\thisrow{sd}] {
x  m  sd
0  9.10  0.95
1  11.00  0.95
2  18.50  1.10
3  20.40  1.10
4  19.60  1.10
5  19.40  1.00
6  22.20  1.10
};
\addplot[name path=LSONLY, draw=none, forget plot] table[x=x, y expr=\thisrow{m}-\thisrow{sd}] {
x  m  sd
0  9.10  0.95
1  11.00  0.95
2  18.50  1.10
3  20.40  1.10
4  19.60  1.10
5  19.40  1.00
6  22.20  1.10
};
\addplot[stuV, fill opacity=0.10, draw=none, forget plot] fill between[of=USONLY and LSONLY];
\addplot[name path=URoDeL, draw=none, forget plot] table[x=x, y expr=\thisrow{m}+\thisrow{sd}] {
x  m  sd
0  9.10  0.95
1  11.00  0.95
2  11.80  1.17
3  18.90  1.36
4  24.00  1.25
5  23.30  1.24
6  24.60  1.32
};
\addplot[name path=LRoDeL, draw=none, forget plot] table[x=x, y expr=\thisrow{m}-\thisrow{sd}] {
x  m  sd
0  9.10  0.95
1  11.00  0.95
2  11.80  1.17
3  18.90  1.36
4  24.00  1.25
5  23.30  1.24
6  24.60  1.32
};
\addplot[lineO, fill opacity=0.10, draw=none, forget plot] fill between[of=URoDeL and LRoDeL];
\addplot[name path=USim, draw=none, forget plot] table[x=x, y expr=\thisrow{m}+\thisrow{sd}] {
x  m  sd
0  9.10  0.95
2  17.90  1.37
4  21.90  1.33
6  23.70  1.22
};
\addplot[name path=LSim, draw=none, forget plot] table[x=x, y expr=\thisrow{m}-\thisrow{sd}] {
x  m  sd
0  9.10  0.95
2  17.90  1.37
4  21.90  1.33
6  23.70  1.22
};
\addplot[simR, fill opacity=0.10, draw=none, forget plot] fill between[of=USim and LSim];
\addplot[ablT, line width=0.8pt, mark=triangle*, mark size=2.0pt, mark options={fill=ablT, draw=white, line width=0.3pt}] table[x=x, y=m] {
x  m
0  9.10
1  8.30
2  21.50
3  14.70
4  18.20
5  20.30
6  24.20
};
\addplot[stuV, line width=0.8pt, mark=square*, mark size=1.5pt, mark options={fill=stuV, draw=white, line width=0.3pt}] table[x=x, y=m] {
x  m
0  9.10
1  11.00
2  18.50
3  20.40
4  19.60
5  19.40
6  22.20
};
\addplot[simR, line width=0.8pt, mark=diamond*, mark size=1.9pt, mark options={fill=simR, draw=white, line width=0.3pt}] table[x=x, y=m] {
x  m
0  9.10
2  17.90
4  21.90
6  23.70
};
\addplot[lineO, line width=0.8pt, mark=*, mark size=1.6pt, mark options={fill=lineO, draw=white, line width=0.3pt}] table[x=x, y=m] {
x  m
0  9.10
1  11.00
2  11.80
3  18.90
4  24.00
5  23.30
6  24.60
};
\nextgroupplot[title={LoCoMo}, ymin=5, ymax=40]
\addplot[name path=UTONLY, draw=none, forget plot] table[x=x, y expr=\thisrow{m}+\thisrow{sd}] {
x  m  sd
0  8.50  0.64
1  7.80  0.60
2  26.70  0.80
3  20.20  0.80
4  25.80  0.90
5  26.60  1.00
6  31.00  0.90
};
\addplot[name path=LTONLY, draw=none, forget plot] table[x=x, y expr=\thisrow{m}-\thisrow{sd}] {
x  m  sd
0  8.50  0.64
1  7.80  0.60
2  26.70  0.80
3  20.20  0.80
4  25.80  0.90
5  26.60  1.00
6  31.00  0.90
};
\addplot[ablT, fill opacity=0.10, draw=none, forget plot] fill between[of=UTONLY and LTONLY];
\addplot[name path=USONLY, draw=none, forget plot] table[x=x, y expr=\thisrow{m}+\thisrow{sd}] {
x  m  sd
0  8.50  0.64
1  10.50  0.68
2  21.20  0.80
3  23.70  0.80
4  23.70  0.80
5  24.50  0.90
6  25.50  0.90
};
\addplot[name path=LSONLY, draw=none, forget plot] table[x=x, y expr=\thisrow{m}-\thisrow{sd}] {
x  m  sd
0  8.50  0.64
1  10.50  0.68
2  21.20  0.80
3  23.70  0.80
4  23.70  0.80
5  24.50  0.90
6  25.50  0.90
};
\addplot[stuV, fill opacity=0.10, draw=none, forget plot] fill between[of=USONLY and LSONLY];
\addplot[name path=URoDeL, draw=none, forget plot] table[x=x, y expr=\thisrow{m}+\thisrow{sd}] {
x  m  sd
0  8.50  0.64
1  10.50  0.68
2  11.80  0.72
3  19.40  0.80
4  34.10  0.95
5  33.40  0.97
6  36.20  0.97
};
\addplot[name path=LRoDeL, draw=none, forget plot] table[x=x, y expr=\thisrow{m}-\thisrow{sd}] {
x  m  sd
0  8.50  0.64
1  10.50  0.68
2  11.80  0.72
3  19.40  0.80
4  34.10  0.95
5  33.40  0.97
6  36.20  0.97
};
\addplot[lineO, fill opacity=0.10, draw=none, forget plot] fill between[of=URoDeL and LRoDeL];
\addplot[name path=USim, draw=none, forget plot] table[x=x, y expr=\thisrow{m}+\thisrow{sd}] {
x  m  sd
0  8.50  0.64
2  19.50  0.83
4  34.70  0.96
6  36.90  0.94
};
\addplot[name path=LSim, draw=none, forget plot] table[x=x, y expr=\thisrow{m}-\thisrow{sd}] {
x  m  sd
0  8.50  0.64
2  19.50  0.83
4  34.70  0.96
6  36.90  0.94
};
\addplot[simR, fill opacity=0.10, draw=none, forget plot] fill between[of=USim and LSim];
\addplot[ablT, line width=0.8pt, mark=triangle*, mark size=2.0pt, mark options={fill=ablT, draw=white, line width=0.3pt}] table[x=x, y=m] {
x  m
0  8.50
1  7.80
2  26.70
3  20.20
4  25.80
5  26.60
6  31.00
};
\addplot[stuV, line width=0.8pt, mark=square*, mark size=1.5pt, mark options={fill=stuV, draw=white, line width=0.3pt}] table[x=x, y=m] {
x  m
0  8.50
1  10.50
2  21.20
3  23.70
4  23.70
5  24.50
6  25.50
};
\addplot[simR, line width=0.8pt, mark=diamond*, mark size=1.9pt, mark options={fill=simR, draw=white, line width=0.3pt}] table[x=x, y=m] {
x  m
0  8.50
2  19.50
4  34.70
6  36.90
};
\addplot[lineO, line width=0.8pt, mark=*, mark size=1.6pt, mark options={fill=lineO, draw=white, line width=0.3pt}] table[x=x, y=m] {
x  m
0  8.50
1  10.50
2  11.80
3  19.40
4  34.10
5  33.40
6  36.20
};
\end{groupplot}
\node[anchor=south, yshift=0.42cm] at (group c3r1.north) {\pgfplotslegendfromname{abllegend}};
\end{tikzpicture}
\caption{F1 on each dataset after each epoch with Qwen3.5-4B, when both roles are trained (RoDeL-Seq and RoDeL-Sim), only the student, or only the teacher. RoDeL-Sim trains both adapters in each update $u_k$; its F1 after $u_k$ is plotted at epoch $2k$, where each adapter has had $k$ updates in both variants. HotpotQA and MuSiQue (ID) are in-domain and the others out-of-domain. Fr.: frozen roles (PRAG-style). Shaded bands show $\pm 1$ standard deviation of a test-set bootstrap.}
\label{fig:abl_perset}
\end{figure}

At the end of training, RoDeL-Seq leads the student-only variant by 2.3 points in domain and 3.1 out of domain, and it leads the teacher-only variant by 12.7 and 5.2. A paired bootstrap test over test documents (\Cref{app:results}) gives $p<0.05$ for every lead except the 2.3-point one. RoDeL-Sim ends at 44.2 in domain and 47.2 out of domain, above both single-role variants. The student-only variant gains most in its first two updates and changes little afterwards, ending at 34.6 in domain and 41.8 out of domain. The teacher-only variant fluctuates from update to update. Its out-of-domain average peaks at 45.7 after the fifth update, 1.9 below the best of RoDeL-Seq, and falls to 39.7 after the sixth.

\wrapneed{6.3cm} %
\begin{wrapfigure}{r}{\colw}
\centering
\begin{tikzpicture}
\begin{groupplot}[
  group style={group size=2 by 1, horizontal sep=0.22cm},
  scale only axis, width=3.05cm, height=3.1cm,
  xmin=-0.5, xmax=6.5, ymin=15, ymax=50,
  xtick={0,...,6}, xticklabels={Fr.,1,2,3,4,5,6},
  ytick={20,30,40,50},
  axis lines=left, axis line style={-, black!60}, tick style={black!60}, major tick length=2pt,
  tick label style={font=\scriptsize, inner sep=1.5pt}, label style={font=\footnotesize},
  title style={font=\footnotesize, yshift=-3pt},
  ymajorgrids, grid style={black!22, dashed},
  extra x ticks={0.5}, extra x tick labels={},
  extra x tick style={grid=major, major grid style={black!25, solid}, tick style={draw=none}},
  clip=false,
]
\nextgroupplot[title={In-domain}, ylabel={Average F1}, ylabel style={yshift=-2pt}, legend style={font=\scriptsize, draw=none, at={(1.036,1.20)}, anchor=south, legend columns=2, legend cell align=left, /tikz/every even column/.append style={column sep=12pt}}, reverse legend]
\addplot[name path=UTONLY, draw=none, forget plot] table[x=x, y expr=\thisrow{m}+\thisrow{sd}] {
x  m  sd
0  23.40  1.82
1  20.00  1.77
2  20.00  1.60
3  27.10  1.90
4  32.40  1.97
5  29.10  1.96
6  24.20  1.82
};
\addplot[name path=LTONLY, draw=none, forget plot] table[x=x, y expr=\thisrow{m}-\thisrow{sd}] {
x  m  sd
0  23.40  1.82
1  20.00  1.77
2  20.00  1.60
3  27.10  1.90
4  32.40  1.97
5  29.10  1.96
6  24.20  1.82
};
\addplot[ablT, fill opacity=0.12, draw=none, forget plot] fill between[of=UTONLY and LTONLY];
\addplot[name path=USONLY, draw=none, forget plot] table[x=x, y expr=\thisrow{m}+\thisrow{sd}] {
x  m  sd
0  23.40  1.82
1  24.50  1.92
2  31.80  2.04
3  33.90  2.07
4  33.90  2.07
5  34.10  2.07
6  34.60  2.02
};
\addplot[name path=LSONLY, draw=none, forget plot] table[x=x, y expr=\thisrow{m}-\thisrow{sd}] {
x  m  sd
0  23.40  1.82
1  24.50  1.92
2  31.80  2.04
3  33.90  2.07
4  33.90  2.07
5  34.10  2.07
6  34.60  2.02
};
\addplot[stuV, fill opacity=0.12, draw=none, forget plot] fill between[of=USONLY and LSONLY];
\addplot[name path=URoDeL, draw=none, forget plot] table[x=x, y expr=\thisrow{m}+\thisrow{sd}] {
x  m  sd
0  23.40  1.82
1  24.50  1.92
2  26.50  1.99
3  32.90  2.06
4  37.30  2.06
5  38.00  2.11
6  36.90  2.05
};
\addplot[name path=LRoDeL, draw=none, forget plot] table[x=x, y expr=\thisrow{m}-\thisrow{sd}] {
x  m  sd
0  23.40  1.82
1  24.50  1.92
2  26.50  1.99
3  32.90  2.06
4  37.30  2.06
5  38.00  2.11
6  36.90  2.05
};
\addplot[lineO, fill opacity=0.12, draw=none, forget plot] fill between[of=URoDeL and LRoDeL];
\addplot[name path=USim, draw=none, forget plot] table[x=x, y expr=\thisrow{m}+\thisrow{sd}] {
x  m  sd
0  23.40  1.82
2  25.20  1.88
4  40.10  2.19
6  44.20  2.11
};
\addplot[name path=LSim, draw=none, forget plot] table[x=x, y expr=\thisrow{m}-\thisrow{sd}] {
x  m  sd
0  23.40  1.82
2  25.20  1.88
4  40.10  2.19
6  44.20  2.11
};
\addplot[simR, fill opacity=0.12, draw=none, forget plot] fill between[of=USim and LSim];
\addplot[ablT, line width=1.0pt, mark=triangle*, mark size=2.2pt, mark options={fill=ablT, draw=white, line width=0.45pt}] table[x=x, y=m] {
x  m
0  23.40
1  20.00
2  20.00
3  27.10
4  32.40
5  29.10
6  24.20
};
\addlegendentry{Teacher only}
\addplot[stuV, line width=1.0pt, mark=square*, mark size=1.7pt, mark options={fill=stuV, draw=white, line width=0.45pt}] table[x=x, y=m] {
x  m
0  23.40
1  24.50
2  31.80
3  33.90
4  33.90
5  34.10
6  34.60
};
\addlegendentry{Student only}
\addplot[simR, line width=1.0pt, mark=diamond*, mark size=2.2pt, mark options={fill=simR, draw=white, line width=0.45pt}] table[x=x, y=m] {
x  m
0  23.40
2  25.20
4  40.10
6  44.20
};
\addlegendentry{RoDeL-Sim}
\addplot[lineO, line width=1.0pt, mark=*, mark size=1.8pt, mark options={fill=lineO, draw=white, line width=0.45pt}] table[x=x, y=m] {
x  m
0  23.40
1  24.50
2  26.50
3  32.90
4  37.30
5  38.00
6  36.90
};
\addlegendentry{RoDeL-Seq}
\nextgroupplot[title={Out-of-domain}, yticklabels={}]
\addplot[name path=UTONLY, draw=none, forget plot] table[x=x, y expr=\thisrow{m}+\thisrow{sd}] {
x  m  sd
0  30.10  0.81
1  28.20  0.83
2  36.00  0.76
3  37.90  0.80
4  40.50  0.80
5  45.70  0.85
6  39.70  0.78
};
\addplot[name path=LTONLY, draw=none, forget plot] table[x=x, y expr=\thisrow{m}-\thisrow{sd}] {
x  m  sd
0  30.10  0.81
1  28.20  0.83
2  36.00  0.76
3  37.90  0.80
4  40.50  0.80
5  45.70  0.85
6  39.70  0.78
};
\addplot[ablT, fill opacity=0.12, draw=none, forget plot] fill between[of=UTONLY and LTONLY];
\addplot[name path=USONLY, draw=none, forget plot] table[x=x, y expr=\thisrow{m}+\thisrow{sd}] {
x  m  sd
0  30.10  0.81
1  32.00  0.84
2  39.40  0.82
3  40.70  0.81
4  40.10  0.87
5  40.20  0.82
6  41.80  0.82
};
\addplot[name path=LSONLY, draw=none, forget plot] table[x=x, y expr=\thisrow{m}-\thisrow{sd}] {
x  m  sd
0  30.10  0.81
1  32.00  0.84
2  39.40  0.82
3  40.70  0.81
4  40.10  0.87
5  40.20  0.82
6  41.80  0.82
};
\addplot[stuV, fill opacity=0.12, draw=none, forget plot] fill between[of=USONLY and LSONLY];
\addplot[name path=URoDeL, draw=none, forget plot] table[x=x, y expr=\thisrow{m}+\thisrow{sd}] {
x  m  sd
0  30.10  0.81
1  32.00  0.84
2  33.20  0.82
3  40.60  0.80
4  47.60  0.82
5  46.40  0.81
6  44.90  0.81
};
\addplot[name path=LRoDeL, draw=none, forget plot] table[x=x, y expr=\thisrow{m}-\thisrow{sd}] {
x  m  sd
0  30.10  0.81
1  32.00  0.84
2  33.20  0.82
3  40.60  0.80
4  47.60  0.82
5  46.40  0.81
6  44.90  0.81
};
\addplot[lineO, fill opacity=0.12, draw=none, forget plot] fill between[of=URoDeL and LRoDeL];
\addplot[name path=USim, draw=none, forget plot] table[x=x, y expr=\thisrow{m}+\thisrow{sd}] {
x  m  sd
0  30.10  0.81
2  36.70  0.81
4  47.80  0.78
6  47.20  0.77
};
\addplot[name path=LSim, draw=none, forget plot] table[x=x, y expr=\thisrow{m}-\thisrow{sd}] {
x  m  sd
0  30.10  0.81
2  36.70  0.81
4  47.80  0.78
6  47.20  0.77
};
\addplot[simR, fill opacity=0.12, draw=none, forget plot] fill between[of=USim and LSim];
\addplot[ablT, line width=1.0pt, mark=triangle*, mark size=2.2pt, mark options={fill=ablT, draw=white, line width=0.45pt}] table[x=x, y=m] {
x  m
0  30.10
1  28.20
2  36.00
3  37.90
4  40.50
5  45.70
6  39.70
};
\addplot[stuV, line width=1.0pt, mark=square*, mark size=1.7pt, mark options={fill=stuV, draw=white, line width=0.45pt}] table[x=x, y=m] {
x  m
0  30.10
1  32.00
2  39.40
3  40.70
4  40.10
5  40.20
6  41.80
};
\addplot[simR, line width=1.0pt, mark=diamond*, mark size=2.2pt, mark options={fill=simR, draw=white, line width=0.45pt}] table[x=x, y=m] {
x  m
0  30.10
2  36.70
4  47.80
6  47.20
};
\addplot[lineO, line width=1.0pt, mark=*, mark size=1.8pt, mark options={fill=lineO, draw=white, line width=0.45pt}] table[x=x, y=m] {
x  m
0  30.10
1  32.00
2  33.20
3  40.60
4  47.60
5  46.40
6  44.90
};
\end{groupplot}
\end{tikzpicture}
\caption{Average F1 after each epoch with Qwen3.5-4B. Lines, Fr., and shaded bands are as in \Cref{fig:abl_perset}.}
\label{fig:abl_avg}
\end{wrapfigure}

Training the teacher together with the student also yields a better teacher, in both variants.
Compared after the same number of teacher updates, with the base model reading the memory adapters (\Cref{fig:abl_roles}), the teacher of RoDeL-Seq scores higher than the teacher-only variant by 3.8, 9.0, and 7.1 points out of domain after one, two, and three updates, and by 6.2, 12.8, and 8.9 points in domain, all with $p<0.05$. \Cref{fig:simteacher} repeats the comparison for the teacher of RoDeL-Sim. After one, two, and three updates, this teacher scores 28.2, 34.9, and 40.4 in domain and 36.7, 45.1, and 48.0 out of domain, above the teacher of the teacher-only variant after every update, all with $p<0.05$.

\wrapend

\begin{figure}[!htb]
\begin{minipage}[t]{\colw}
\centering
\begin{tikzpicture}
\begin{groupplot}[
  group style={group size=2 by 1, horizontal sep=0.22cm},
  scale only axis, width=2.92cm, height=2.5cm,
  xmin=-0.4, xmax=3.4, ymin=13, ymax=52,
  xtick={0,...,3}, xticklabels={Fr.,1,2,3},
  ytick={20,30,40,50},
  axis lines=left, axis line style={-, black!60}, tick style={black!60}, major tick length=2pt,
  tick label style={font=\scriptsize, inner sep=1.5pt}, label style={font=\footnotesize},
  title style={font=\footnotesize, yshift=-3pt},
  ymajorgrids, grid style={black!22, dashed},
  clip=false,
]
\nextgroupplot[title={In-domain}, ylabel={Average F1}, ylabel style={yshift=-2pt}, xlabel={Teacher updates}, xlabel style={font=\scriptsize, xshift=1.59cm, yshift=2pt}, legend to name=ablroles0, legend columns=1, legend cell align=left, legend style={font=\scriptsize, draw=none, row sep=-1pt}]
\addplot[name path=UA, draw=none, forget plot] table[x=x, y expr=\thisrow{m}+\thisrow{sd}] {
x  m  sd
0  23.40  1.82
1  20.00  1.77
2  20.00  1.60
3  27.10  1.90
};
\addplot[name path=LA, draw=none, forget plot] table[x=x, y expr=\thisrow{m}-\thisrow{sd}] {
x  m  sd
0  23.40  1.82
1  20.00  1.77
2  20.00  1.60
3  27.10  1.90
};
\addplot[ablT, fill opacity=0.12, draw=none, forget plot] fill between[of=UA and LA];
\addplot[name path=UR, draw=none, forget plot] table[x=x, y expr=\thisrow{m}+\thisrow{sd}] {
x  m  sd
0  23.40  1.82
1  26.20  1.96
2  32.80  2.04
3  36.00  2.10
};
\addplot[name path=LR, draw=none, forget plot] table[x=x, y expr=\thisrow{m}-\thisrow{sd}] {
x  m  sd
0  23.40  1.82
1  26.20  1.96
2  32.80  2.04
3  36.00  2.10
};
\addplot[lineO, fill opacity=0.12, draw=none, forget plot] fill between[of=UR and LR];
\addplot[ablT, line width=1.0pt, mark=triangle*, mark size=2.2pt, mark options={fill=ablT, draw=white, line width=0.45pt}, forget plot,
  nodes near coords, point meta=explicit symbolic, every node near coord/.append style={font=\tiny, text=black!70, anchor=north, yshift=-1pt}]
  table[x=x, y=m, meta=l] {
x  m  l
0  23.4  {}
1  20.0  20.0
2  20.0  20.0
3  27.1  27.1
};
\addplot[lineO, line width=1.0pt, mark=*, mark size=1.8pt, mark options={fill=lineO, draw=white, line width=0.45pt}, forget plot,
  nodes near coords, point meta=explicit symbolic, every node near coord/.append style={font=\tiny, text=black!70, anchor=south, yshift=1pt}]
  table[x=x, y=m, meta=l] {
x  m  l
0  23.4  23.4
1  26.2  26.2
2  32.8  32.8
3  36.0  36.0
};
\addlegendimage{lineO, line width=1.0pt, mark=*, mark size=1.8pt, mark options={fill=lineO, draw=white, line width=0.45pt}}
\addlegendentry{RoDeL-Seq teacher + base student}
\addlegendimage{ablT, line width=1.0pt, mark=triangle*, mark size=2.2pt, mark options={fill=ablT, draw=white, line width=0.45pt}}
\addlegendentry{Teacher-only teacher + base student}
\nextgroupplot[title={Out-of-domain}, yticklabels={}]
\addplot[name path=UA, draw=none, forget plot] table[x=x, y expr=\thisrow{m}+\thisrow{sd}] {
x  m  sd
0  30.10  0.81
1  28.20  0.83
2  36.00  0.76
3  37.90  0.80
};
\addplot[name path=LA, draw=none, forget plot] table[x=x, y expr=\thisrow{m}-\thisrow{sd}] {
x  m  sd
0  30.10  0.81
1  28.20  0.83
2  36.00  0.76
3  37.90  0.80
};
\addplot[ablT, fill opacity=0.12, draw=none, forget plot] fill between[of=UA and LA];
\addplot[name path=UR, draw=none, forget plot] table[x=x, y expr=\thisrow{m}+\thisrow{sd}] {
x  m  sd
0  30.10  0.81
1  32.00  0.82
2  45.00  0.84
3  45.00  0.81
};
\addplot[name path=LR, draw=none, forget plot] table[x=x, y expr=\thisrow{m}-\thisrow{sd}] {
x  m  sd
0  30.10  0.81
1  32.00  0.82
2  45.00  0.84
3  45.00  0.81
};
\addplot[lineO, fill opacity=0.12, draw=none, forget plot] fill between[of=UR and LR];
\addplot[ablT, line width=1.0pt, mark=triangle*, mark size=2.2pt, mark options={fill=ablT, draw=white, line width=0.45pt}, forget plot,
  nodes near coords, point meta=explicit symbolic, every node near coord/.append style={font=\tiny, text=black!70, anchor=north, yshift=-1pt}]
  table[x=x, y=m, meta=l] {
x  m  l
0  30.1  {}
1  28.2  28.2
2  36.0  36.0
3  37.9  37.9
};
\addplot[lineO, line width=1.0pt, mark=*, mark size=1.8pt, mark options={fill=lineO, draw=white, line width=0.45pt}, forget plot,
  nodes near coords, point meta=explicit symbolic, every node near coord/.append style={font=\tiny, text=black!70, anchor=south, yshift=1pt}]
  table[x=x, y=m, meta=l] {
x  m  l
0  30.1  30.1
1  32.0  32.0
2  45.0  45.0
3  45.0  45.0
};
\end{groupplot}
\node[anchor=south, yshift=0.40cm] at ($(group c1r1.north)!0.5!(group c2r1.north)$) {\pgfplotslegendfromname{ablroles0}};
\end{tikzpicture}
\caption{Average F1 after the same number of teacher updates with Qwen3.5-4B. Each line names the teacher that writes the memory adapters and the student that reads them; ``base'' is the base model without a role adapter. Fr.: frozen roles. Shaded bands show $\pm 1$ standard deviation of a test-set bootstrap.}
\label{fig:abl_roles}
\end{minipage}\hfill
\begin{minipage}[t]{\colw}
\centering
\begin{tikzpicture}
\begin{groupplot}[
  group style={group size=2 by 1, horizontal sep=0.22cm},
  scale only axis, width=2.92cm, height=2.5cm,
  xmin=-0.4, xmax=3.4, ymin=13, ymax=52,
  xtick={0,...,3}, xticklabels={Fr.,1,2,3},
  ytick={20,30,40,50},
  axis lines=left, axis line style={-, black!60}, tick style={black!60}, major tick length=2pt,
  tick label style={font=\scriptsize, inner sep=1.5pt}, label style={font=\footnotesize},
  title style={font=\footnotesize, yshift=-3pt},
  ymajorgrids, grid style={black!22, dashed},
  clip=false,
]
\nextgroupplot[title={In-domain}, ylabel={Average F1}, ylabel style={yshift=-2pt}, xlabel={Teacher updates}, xlabel style={font=\scriptsize, xshift=1.59cm, yshift=2pt}, legend to name=simteacher0, legend columns=1, legend cell align=left, legend style={font=\scriptsize, draw=none, row sep=-1pt}]
\addplot[name path=UA, draw=none, forget plot] table[x=x, y expr=\thisrow{m}+\thisrow{sd}] {
x  m  sd
0  23.40  1.82
1  20.00  1.77
2  20.00  1.60
3  27.10  1.90
};
\addplot[name path=LA, draw=none, forget plot] table[x=x, y expr=\thisrow{m}-\thisrow{sd}] {
x  m  sd
0  23.40  1.82
1  20.00  1.77
2  20.00  1.60
3  27.10  1.90
};
\addplot[ablT, fill opacity=0.12, draw=none, forget plot] fill between[of=UA and LA];
\addplot[name path=UR, draw=none, forget plot] table[x=x, y expr=\thisrow{m}+\thisrow{sd}] {
x  m  sd
0  23.40  1.82
1  28.20  2.02
2  34.90  2.07
3  40.40  2.12
};
\addplot[name path=LR, draw=none, forget plot] table[x=x, y expr=\thisrow{m}-\thisrow{sd}] {
x  m  sd
0  23.40  1.82
1  28.20  2.02
2  34.90  2.07
3  40.40  2.12
};
\addplot[lineO, fill opacity=0.12, draw=none, forget plot] fill between[of=UR and LR];
\addplot[ablT, line width=1.0pt, mark=triangle*, mark size=2.2pt, mark options={fill=ablT, draw=white, line width=0.45pt}, forget plot,
  nodes near coords, point meta=explicit symbolic, every node near coord/.append style={font=\tiny, text=black!70, anchor=north, yshift=-1pt}]
  table[x=x, y=m, meta=l] {
x  m  l
0  23.4  {}
1  20.0  20.0
2  20.0  20.0
3  27.1  27.1
};
\addplot[lineO, line width=1.0pt, mark=*, mark size=1.8pt, mark options={fill=lineO, draw=white, line width=0.45pt}, forget plot,
  nodes near coords, point meta=explicit symbolic, every node near coord/.append style={font=\tiny, text=black!70, anchor=south, yshift=1pt}]
  table[x=x, y=m, meta=l] {
x  m  l
0  23.4  23.4
1  28.2  28.2
2  34.9  34.9
3  40.4  40.4
};
\addlegendimage{lineO, line width=1.0pt, mark=*, mark size=1.8pt, mark options={fill=lineO, draw=white, line width=0.45pt}}
\addlegendentry{RoDeL-Sim teacher + base student}
\addlegendimage{ablT, line width=1.0pt, mark=triangle*, mark size=2.2pt, mark options={fill=ablT, draw=white, line width=0.45pt}}
\addlegendentry{Teacher-only teacher + base student}
\nextgroupplot[title={Out-of-domain}, yticklabels={}]
\addplot[name path=UA, draw=none, forget plot] table[x=x, y expr=\thisrow{m}+\thisrow{sd}] {
x  m  sd
0  30.10  0.81
1  28.20  0.83
2  36.00  0.76
3  37.90  0.80
};
\addplot[name path=LA, draw=none, forget plot] table[x=x, y expr=\thisrow{m}-\thisrow{sd}] {
x  m  sd
0  30.10  0.81
1  28.20  0.83
2  36.00  0.76
3  37.90  0.80
};
\addplot[ablT, fill opacity=0.12, draw=none, forget plot] fill between[of=UA and LA];
\addplot[name path=UR, draw=none, forget plot] table[x=x, y expr=\thisrow{m}+\thisrow{sd}] {
x  m  sd
0  30.10  0.81
1  36.70  0.81
2  45.10  0.81
3  48.00  0.82
};
\addplot[name path=LR, draw=none, forget plot] table[x=x, y expr=\thisrow{m}-\thisrow{sd}] {
x  m  sd
0  30.10  0.81
1  36.70  0.81
2  45.10  0.81
3  48.00  0.82
};
\addplot[lineO, fill opacity=0.12, draw=none, forget plot] fill between[of=UR and LR];
\addplot[ablT, line width=1.0pt, mark=triangle*, mark size=2.2pt, mark options={fill=ablT, draw=white, line width=0.45pt}, forget plot,
  nodes near coords, point meta=explicit symbolic, every node near coord/.append style={font=\tiny, text=black!70, anchor=north, yshift=-1pt}]
  table[x=x, y=m, meta=l] {
x  m  l
0  30.1  {}
1  28.2  28.2
2  36.0  36.0
3  37.9  37.9
};
\addplot[lineO, line width=1.0pt, mark=*, mark size=1.8pt, mark options={fill=lineO, draw=white, line width=0.45pt}, forget plot,
  nodes near coords, point meta=explicit symbolic, every node near coord/.append style={font=\tiny, text=black!70, anchor=south, yshift=1pt}]
  table[x=x, y=m, meta=l] {
x  m  l
0  30.1  30.1
1  36.7  36.7
2  45.1  45.1
3  48.0  48.0
};
\end{groupplot}
\node[anchor=south, yshift=0.40cm] at ($(group c1r1.north)!0.5!(group c2r1.north)$) {\pgfplotslegendfromname{simteacher0}};
\end{tikzpicture}
\caption{Average F1 after the same number of teacher updates with Qwen3.5-4B, with the teacher of RoDeL-Sim; lines and shaded bands are as in \Cref{fig:abl_roles}.}
\label{fig:simteacher}
\end{minipage}
\end{figure}

\subsection{Learning Dynamics}
\label{subsec:training}

\begin{figure}[!htb]
\begin{minipage}[t]{\colw}
\centering
\begin{tikzpicture}
\begin{axis}[scale only axis, width=4.98cm, height=4.0cm, xmin=0, xmax=208, tick label style={font=\scriptsize}, label style={font=\footnotesize}, clip=false,
  ymin=-2.6, ymax=1.0, ytick={-2,-1,0,1}, xtick={0,50,100,150,200},
  axis x line*=bottom, axis y line*=left, axis line style={-, black!60}, tick style={black!60}, major tick length=2pt,
  xlabel={Training step}, ylabel={Training reward},
  legend style={font=\scriptsize, draw=none, fill=none, at={(0.5,1.13)}, anchor=south, legend columns=2, /tikz/every even column/.append style={column sep=8pt}},
  legend cell align=left,
]
\draw[black!25] (axis cs:34.5,-2.6) -- (axis cs:34.5,1.0);
\draw[black!25] (axis cs:69.5,-2.6) -- (axis cs:69.5,1.0);
\draw[black!25] (axis cs:103.5,-2.6) -- (axis cs:103.5,1.0);
\draw[black!25] (axis cs:138.5,-2.6) -- (axis cs:138.5,1.0);
\draw[black!25] (axis cs:172.5,-2.6) -- (axis cs:172.5,1.0);
\node[font=\scriptsize, text=stuV, anchor=south] at (axis cs:17.5,1.0) {$s_1$};
\node[font=\scriptsize, text=tchV, anchor=south] at (axis cs:52.0,1.0) {$t_1$};
\node[font=\scriptsize, text=stuV, anchor=south] at (axis cs:86.5,1.0) {$s_2$};
\node[font=\scriptsize, text=tchV, anchor=south] at (axis cs:121.0,1.0) {$t_2$};
\node[font=\scriptsize, text=stuV, anchor=south] at (axis cs:155.5,1.0) {$s_3$};
\node[font=\scriptsize, text=tchV, anchor=south] at (axis cs:190.0,1.0) {$t_3$};
\addplot[stuV, line width=0.4pt, opacity=0.35, forget plot] coordinates {(1,-0.292) (2,-1.312) (3,-0.656) (4,-1.562) (5,-0.953) (6,-1.078) (7,-0.938) (8,-0.775) (9,-0.625) (10,-0.339) (11,-1.188) (12,-1.323) (13,-0.958) (14,-0.067) (15,-0.897) (16,-1.224) (17,-0.812) (18,-1.094) (19,-0.896) (20,-0.958) (21,-0.521) (22,-0.906) (23,-0.363) (24,-0.580) (25,-0.704) (26,-0.961) (27,-0.125) (28,-0.487) (29,-0.225) (30,-0.312) (31,-0.974) (32,-0.227) (33,-0.359) (34,-0.230)};
\addplot[stuV, line width=1.2pt] coordinates {(1,-0.956) (2,-0.955) (3,-0.975) (4,-0.970) (5,-1.039) (6,-0.941) (7,-0.896) (8,-0.842) (9,-0.895) (10,-0.878) (11,-0.754) (12,-0.771) (13,-0.857) (14,-0.924) (15,-0.911) (16,-0.850) (17,-0.850) (18,-0.915) (19,-0.916) (20,-0.793) (21,-0.760) (22,-0.704) (23,-0.713) (24,-0.594) (25,-0.589) (26,-0.492) (27,-0.485) (28,-0.541) (29,-0.473) (30,-0.387) (31,-0.402) (32,-0.388) (33,-0.420) (34,-0.448)};
\addlegendentry{Reward ($s_k$)}
\addplot[tchV, line width=0.4pt, opacity=0.35, forget plot] coordinates {(35,-2.144) (36,-0.356) (37,-0.825) (38,-2.438) (39,-0.620) (40,-0.890) (41,-1.273) (42,-0.624) (43,-2.123) (44,-1.753) (45,-1.073) (46,-0.477) (47,-2.433) (48,-1.401) (49,-0.887) (50,-1.866) (51,-0.754) (52,-1.174) (53,-0.999) (54,-1.201) (55,-1.153) (56,-0.370) (57,-1.125) (58,-1.508) (59,-0.534) (60,-1.688) (61,-1.680) (62,-0.666) (63,-1.085) (64,-1.357) (65,-1.004) (66,-1.556) (67,-2.200) (68,-1.934) (69,-0.533)};
\addplot[tchV, line width=1.2pt] coordinates {(35,-1.441) (36,-1.277) (37,-1.212) (38,-1.221) (39,-1.004) (40,-1.256) (41,-1.389) (42,-1.194) (43,-1.173) (44,-1.394) (45,-1.412) (46,-1.450) (47,-1.413) (48,-1.270) (49,-1.285) (50,-1.359) (51,-1.183) (52,-1.148) (53,-1.074) (54,-0.968) (55,-1.076) (56,-0.984) (57,-1.083) (58,-1.151) (59,-1.082) (60,-1.184) (61,-1.217) (62,-1.145) (63,-1.291) (64,-1.364) (65,-1.400) (66,-1.381) (67,-1.431) (68,-1.445) (69,-1.556)};
\addlegendentry{Reward ($t_k$)}
\addplot[stuV, line width=0.4pt, opacity=0.35, forget plot] coordinates {(70,-0.156) (71,-0.750) (72,-0.756) (73,-1.141) (74,-0.682) (75,-0.831) (76,-1.104) (77,-0.744) (78,0.159) (79,-0.016) (80,-0.028) (81,-0.531) (82,0.047) (83,-0.453) (84,-0.416) (85,-1.138) (86,-0.396) (87,0.018) (88,-0.010) (89,0.013) (90,-0.088) (91,0.105) (92,0.025) (93,0.339) (94,0.453) (95,0.521) (96,0.316) (97,0.342) (98,0.431) (99,0.125) (100,0.106) (101,0.609) (102,0.188) (103,0.477)};
\addplot[stuV, line width=1.2pt, forget plot] coordinates {(70,-0.701) (71,-0.697) (72,-0.719) (73,-0.774) (74,-0.858) (75,-0.728) (76,-0.623) (77,-0.464) (78,-0.442) (79,-0.317) (80,-0.224) (81,-0.177) (82,-0.362) (83,-0.416) (84,-0.410) (85,-0.335) (86,-0.340) (87,-0.288) (88,-0.214) (89,-0.048) (90,0.057) (91,0.120) (92,0.195) (93,0.239) (94,0.300) (95,0.347) (96,0.361) (97,0.328) (98,0.350) (99,0.302) (100,0.325) (101,0.323) (102,0.301) (103,0.345)};
\addplot[tchV, line width=0.4pt, opacity=0.35, forget plot] coordinates {(104,-1.037) (105,-0.124) (106,-0.272) (107,-0.720) (108,0.171) (109,-0.038) (110,0.102) (111,0.060) (112,-0.279) (113,0.068) (114,-0.060) (115,0.084) (116,-0.252) (117,0.189) (118,0.299) (119,-0.409) (120,0.383) (121,0.143) (122,0.579) (123,0.014) (124,0.137) (125,0.292) (126,0.111) (127,-0.222) (128,0.317) (129,0.014) (130,0.437) (131,0.471) (132,0.262) (133,0.269) (134,0.340) (135,-0.049) (136,-0.210) (137,0.159) (138,0.494)};
\addplot[tchV, line width=1.2pt, forget plot] coordinates {(104,-0.538) (105,-0.396) (106,-0.337) (107,-0.274) (108,-0.117) (109,-0.139) (110,-0.091) (111,0.003) (112,-0.009) (113,-0.040) (114,-0.027) (115,0.007) (116,-0.012) (117,0.033) (118,0.062) (119,0.133) (120,0.171) (121,0.164) (122,0.163) (123,0.237) (124,0.151) (125,0.175) (126,0.095) (127,0.155) (128,0.203) (129,0.199) (130,0.221) (131,0.301) (132,0.249) (133,0.217) (134,0.177) (135,0.181) (136,0.167) (137,0.147) (138,0.099)};
\addplot[stuV, line width=0.4pt, opacity=0.35, forget plot] coordinates {(139,0.231) (140,0.254) (141,0.321) (142,0.250) (143,0.512) (144,0.240) (145,0.300) (146,0.312) (147,0.519) (148,0.345) (149,0.401) (150,0.318) (151,0.138) (152,0.410) (153,0.030) (154,0.426) (155,0.172) (156,0.302) (157,0.365) (158,0.452) (159,0.485) (160,0.443) (161,0.502) (162,0.255) (163,0.406) (164,0.536) (165,0.577) (166,0.518) (167,0.438) (168,0.292) (169,0.053) (170,0.642) (171,0.271) (172,0.495)};
\addplot[stuV, line width=1.2pt, forget plot] coordinates {(139,0.264) (140,0.314) (141,0.301) (142,0.301) (143,0.313) (144,0.351) (145,0.354) (146,0.376) (147,0.348) (148,0.333) (149,0.349) (150,0.309) (151,0.295) (152,0.271) (153,0.257) (154,0.263) (155,0.308) (156,0.319) (157,0.378) (158,0.389) (159,0.401) (160,0.415) (161,0.440) (162,0.458) (163,0.462) (164,0.462) (165,0.432) (166,0.403) (167,0.437) (168,0.399) (169,0.387) (170,0.365) (171,0.351) (172,0.365)};
\addplot[tchV, line width=0.4pt, opacity=0.35, forget plot] coordinates {(173,-0.188) (174,0.512) (175,0.331) (176,-0.072) (177,0.423) (178,0.497) (179,0.286) (180,0.309) (181,0.033) (182,0.086) (183,0.325) (184,0.516) (185,-0.088) (186,0.536) (187,0.496) (188,0.000) (189,0.591) (190,0.360) (191,0.551) (192,0.410) (193,0.532) (194,0.543) (195,0.408) (196,0.070) (197,0.557) (198,0.214) (199,0.386) (200,0.411) (201,0.548) (202,0.330) (203,0.268) (204,0.230) (205,0.035) (206,0.117) (207,0.446)};
\addplot[tchV, line width=1.2pt, forget plot] coordinates {(173,0.146) (174,0.201) (175,0.251) (176,0.256) (177,0.327) (178,0.258) (179,0.223) (180,0.280) (181,0.293) (182,0.210) (183,0.245) (184,0.272) (185,0.267) (186,0.339) (187,0.344) (188,0.349) (189,0.421) (190,0.420) (191,0.427) (192,0.485) (193,0.411) (194,0.439) (195,0.391) (196,0.387) (197,0.370) (198,0.371) (199,0.359) (200,0.388) (201,0.341) (202,0.315) (203,0.277) (204,0.282) (205,0.238) (206,0.219) (207,0.207)};
\addlegendimage{lineO, line width=1.0pt, draw opacity=0.45, mark=*, mark size=1.7pt, mark options={fill=lineO, draw=white, line width=0.4pt, fill opacity=0.45, draw opacity=0.45}}
\addlegendentry{Test F1 (ID)}
\addlegendimage{lineO, line width=1.0pt, draw opacity=0.45, mark=square*, mark size=1.5pt, mark options={fill=lineO, draw=white, line width=0.4pt, fill opacity=0.45, draw opacity=0.45}}
\addlegendentry{Test F1 (OOD)}
\end{axis}
\begin{axis}[scale only axis, width=4.98cm, height=4.0cm, xmin=0, xmax=208, tick label style={font=\scriptsize}, label style={font=\footnotesize}, clip=false,
  ymin=0, ymax=60, ytick={10,20,30,40,50,60},
  axis x line=none, axis y line*=right, axis line style={-, lineO!80}, tick style={lineO!80}, major tick length=2pt,
  yticklabel style={text=lineO}, ylabel={Test F1}, ylabel style={text=lineO},
]
\addplot[name path=UID, draw=none, forget plot] table[x=x, y expr=\thisrow{m}+\thisrow{sd}] {
x  m  sd  l
0  23.4  1.82  23.4
34  24.5  1.92  24.5
69  26.5  1.99  26.5
103  32.9  2.06  32.9
138  37.3  2.06  37.3
172  38.0  2.11  38.0
207  36.9  2.05  36.9
};
\addplot[name path=LID, draw=none, forget plot] table[x=x, y expr=\thisrow{m}-\thisrow{sd}] {
x  m  sd  l
0  23.4  1.82  23.4
34  24.5  1.92  24.5
69  26.5  1.99  26.5
103  32.9  2.06  32.9
138  37.3  2.06  37.3
172  38.0  2.11  38.0
207  36.9  2.05  36.9
};
\addplot[lineO, fill opacity=0.08, draw=none, forget plot] fill between[of=UID and LID];
\addplot[lineO, line width=1.0pt, draw opacity=0.45, mark=*, mark size=1.7pt, mark options={fill=lineO, draw=white, line width=0.4pt, fill opacity=0.45, draw opacity=0.45},
  nodes near coords, point meta=explicit symbolic, every node near coord/.append style={font=\tiny, text=black!75, anchor=north, yshift=-2pt, fill=white, fill opacity=0.8, text opacity=1, inner sep=0.6pt}]
  table[x=x, y=m, meta=l] {
x  m  sd  l
0  23.4  1.82  23.4
34  24.5  1.92  24.5
69  26.5  1.99  26.5
103  32.9  2.06  32.9
138  37.3  2.06  37.3
172  38.0  2.11  38.0
207  36.9  2.05  36.9
};
\addplot[name path=UOOD, draw=none, forget plot] table[x=x, y expr=\thisrow{m}+\thisrow{sd}] {
x  m  sd  l
0  30.1  0.81  30.1
34  32.0  0.84  32.0
69  33.2  0.82  33.2
103  40.6  0.80  40.6
138  47.6  0.82  47.6
172  46.4  0.81  46.4
207  44.9  0.81  44.9
};
\addplot[name path=LOOD, draw=none, forget plot] table[x=x, y expr=\thisrow{m}-\thisrow{sd}] {
x  m  sd  l
0  30.1  0.81  30.1
34  32.0  0.84  32.0
69  33.2  0.82  33.2
103  40.6  0.80  40.6
138  47.6  0.82  47.6
172  46.4  0.81  46.4
207  44.9  0.81  44.9
};
\addplot[lineO, fill opacity=0.08, draw=none, forget plot] fill between[of=UOOD and LOOD];
\addplot[lineO, line width=1.0pt, draw opacity=0.45, mark=square*, mark size=1.5pt, mark options={fill=lineO, draw=white, line width=0.4pt, fill opacity=0.45, draw opacity=0.45},
  nodes near coords, point meta=explicit symbolic, every node near coord/.append style={font=\tiny, text=black!75, anchor=south, yshift=1.5pt, fill=white, fill opacity=0.8, text opacity=1, inner sep=0.6pt}]
  table[x=x, y=m, meta=l] {
x  m  sd  l
0  30.1  0.81  30.1
34  32.0  0.84  32.0
69  33.2  0.82  33.2
103  40.6  0.80  40.6
138  47.6  0.82  47.6
172  46.4  0.81  46.4
207  44.9  0.81  44.9
};
\end{axis}
\end{tikzpicture}
\caption{Training reward of RoDeL-Seq with Qwen3.5-4B at each step of the six updates (left axis) and average test F1 after each update (right axis). Thin lines show the reward of each step and thick lines a moving average over seven steps within an update. Student and teacher updates compute the reward differently (\Cref{subsec:student,subsec:teacher}), so rewards are not comparable across updates of different roles. Shaded bands show $\pm 1$ standard deviation of a test-set bootstrap.}
\label{fig:train_reward}
\end{minipage}\hfill
\begin{minipage}[t]{\colw}
\centering
\begin{tikzpicture}
\begin{axis}[scale only axis, width=4.98cm, height=4.0cm, xmin=0, xmax=106, tick label style={font=\scriptsize}, label style={font=\footnotesize}, clip=false,
  ymin=-2.8, ymax=1.0, ytick={-2,-1,0,1}, xtick={0,35,70,105},
  axis x line*=bottom, axis y line*=left, axis line style={-, black!60}, tick style={black!60}, major tick length=2pt,
  xlabel={Training step}, ylabel={Training reward},
  legend style={font=\scriptsize, draw=none, fill=none, at={(0.5,1.13)}, anchor=south, legend columns=2, /tikz/every even column/.append style={column sep=8pt}},
  legend cell align=left,
]
\draw[black!25] (axis cs:35.5,-2.8) -- (axis cs:35.5,1.0);
\draw[black!25] (axis cs:70.5,-2.8) -- (axis cs:70.5,1.0);
\node[font=\scriptsize, text=black!75, anchor=south] at (axis cs:18.0,1.0) {$u_1$};
\node[font=\scriptsize, text=black!75, anchor=south] at (axis cs:53.0,1.0) {$u_2$};
\node[font=\scriptsize, text=black!75, anchor=south] at (axis cs:88.0,1.0) {$u_3$};
\addplot[tchV, line width=0.4pt, opacity=0.35, forget plot] coordinates {(1,-2.523) (2,-0.575) (3,-1.379) (4,-2.576) (5,-0.781) (6,-1.270) (7,-1.511) (8,-1.270) (9,-1.894) (10,-2.287) (11,-1.059) (12,-0.735) (13,-2.675) (14,-0.853) (15,-1.046) (16,-1.722) (17,-0.908) (18,-1.320) (19,-1.450) (20,-1.368) (21,-0.715) (22,-0.758) (23,-1.012) (24,-1.605) (25,-0.393) (26,-1.306) (27,-0.808) (28,-1.187) (29,-0.607) (30,-1.400) (31,-0.735) (32,-1.630) (33,-1.442) (34,-1.181) (35,-0.720)};
\addplot[stuV, line width=0.4pt, opacity=0.35, forget plot] coordinates {(1,-1.648) (2,0.030) (3,-0.527) (4,-1.773) (5,-0.389) (6,-0.325) (7,-0.769) (8,-0.355) (9,-1.494) (10,-1.543) (11,-0.718) (12,-0.073) (13,-1.708) (14,-0.131) (15,-0.129) (16,-1.260) (17,-0.333) (18,-0.622) (19,-0.604) (20,-0.537) (21,-0.148) (22,-0.061) (23,-0.364) (24,-1.047) (25,0.134) (26,-0.881) (27,0.005) (28,-0.269) (29,0.075) (30,-0.772) (31,-0.359) (32,-0.822) (33,-0.932) (34,-0.770) (35,-0.461)};
\addplot[tchV, line width=0.4pt, opacity=0.35, forget plot] coordinates {(36,-0.833) (37,-0.273) (38,-0.147) (39,-1.282) (40,0.180) (41,-0.329) (42,-0.264) (43,0.037) (44,-1.200) (45,-0.945) (46,-0.669) (47,-0.227) (48,-0.896) (49,-0.304) (50,-0.253) (51,-0.733) (52,-0.159) (53,-0.640) (54,0.151) (55,-0.057) (56,0.039) (57,0.216) (58,-0.117) (59,-0.280) (60,0.407) (61,-0.011) (62,0.346) (63,0.211) (64,0.198) (65,0.073) (66,0.288) (67,-0.048) (68,-0.139) (69,0.064) (70,0.399)};
\addplot[stuV, line width=0.4pt, opacity=0.35, forget plot] coordinates {(36,-0.510) (37,0.540) (38,0.262) (39,-0.882) (40,0.199) (41,0.336) (42,-0.059) (43,0.309) (44,-0.672) (45,-0.608) (46,-0.035) (47,0.374) (48,-0.825) (49,0.274) (50,0.361) (51,-0.414) (52,0.223) (53,-0.049) (54,0.524) (55,0.202) (56,0.524) (57,0.484) (58,0.081) (59,-0.216) (60,0.629) (61,0.089) (62,0.348) (63,0.448) (64,0.568) (65,0.290) (66,0.453) (67,0.165) (68,-0.022) (69,0.097) (70,0.511)};
\addplot[tchV, line width=0.4pt, opacity=0.35, forget plot] coordinates {(71,-0.258) (72,0.766) (73,0.451) (74,-0.058) (75,0.558) (76,0.528) (77,0.375) (78,0.459) (79,-0.020) (80,0.030) (81,0.369) (82,0.558) (83,-0.051) (84,0.697) (85,0.366) (86,0.250) (87,0.352) (88,0.607) (89,0.728) (90,0.454) (91,0.802) (92,0.691) (93,0.270) (94,0.086) (95,0.685) (96,0.095) (97,0.445) (98,0.614) (99,0.784) (100,0.434) (101,0.449) (102,0.193) (103,0.070) (104,0.319) (105,0.487)};
\addplot[stuV, line width=0.4pt, opacity=0.35, forget plot] coordinates {(71,-0.070) (72,0.879) (73,0.527) (74,0.020) (75,0.550) (76,0.750) (77,0.416) (78,0.548) (79,0.222) (80,0.072) (81,0.479) (82,0.611) (83,0.069) (84,0.692) (85,0.553) (86,0.237) (87,0.429) (88,0.524) (89,0.627) (90,0.464) (91,0.668) (92,0.678) (93,0.376) (94,0.077) (95,0.663) (96,0.173) (97,0.515) (98,0.639) (99,0.664) (100,0.355) (101,0.323) (102,0.227) (103,0.189) (104,0.244) (105,0.449)};
\addplot[stuV, line width=1.2pt] coordinates {(1,-0.979) (2,-0.861) (3,-0.772) (4,-0.772) (5,-0.587) (6,-0.805) (7,-0.950) (8,-0.799) (9,-0.754) (10,-0.951) (11,-0.860) (12,-0.828) (13,-0.795) (14,-0.622) (15,-0.608) (16,-0.684) (17,-0.517) (18,-0.519) (19,-0.509) (20,-0.381) (21,-0.483) (22,-0.375) (23,-0.415) (24,-0.337) (25,-0.355) (26,-0.335) (27,-0.394) (28,-0.295) (29,-0.432) (30,-0.439) (31,-0.550) (32,-0.577) (33,-0.686) (34,-0.669) (35,-0.746)};
\addlegendentry{Student reward}
\addplot[tchV, line width=1.2pt] coordinates {(1,-1.763) (2,-1.567) (3,-1.517) (4,-1.516) (5,-1.337) (6,-1.526) (7,-1.656) (8,-1.439) (9,-1.432) (10,-1.633) (11,-1.539) (12,-1.507) (13,-1.482) (14,-1.285) (15,-1.323) (16,-1.425) (17,-1.238) (18,-1.218) (19,-1.177) (20,-1.076) (21,-1.175) (22,-1.043) (23,-1.022) (24,-0.942) (25,-1.010) (26,-0.988) (27,-1.044) (28,-0.919) (29,-1.096) (30,-1.116) (31,-1.169) (32,-1.102) (33,-1.185) (34,-1.142) (35,-1.243)};
\addlegendentry{Teacher reward}
\addplot[stuV, line width=1.2pt, forget plot] coordinates {(36,-0.147) (37,-0.078) (38,-0.009) (39,-0.016) (40,0.101) (41,-0.072) (42,-0.197) (43,-0.076) (44,-0.051) (45,-0.217) (46,-0.169) (47,-0.162) (48,-0.125) (49,-0.006) (50,-0.008) (51,0.013) (52,0.160) (53,0.196) (54,0.213) (55,0.284) (56,0.221) (57,0.318) (58,0.256) (59,0.277) (60,0.266) (61,0.278) (62,0.308) (63,0.404) (64,0.337) (65,0.321) (66,0.286) (67,0.295) (68,0.249) (69,0.241) (70,0.188)};
\addplot[tchV, line width=1.2pt, forget plot] coordinates {(36,-0.634) (37,-0.471) (38,-0.447) (39,-0.421) (40,-0.297) (41,-0.429) (42,-0.543) (43,-0.456) (44,-0.514) (45,-0.595) (46,-0.601) (47,-0.642) (48,-0.575) (49,-0.463) (50,-0.459) (51,-0.405) (52,-0.285) (53,-0.236) (54,-0.169) (55,-0.081) (56,-0.098) (57,0.051) (58,0.028) (59,0.086) (60,0.110) (61,0.108) (62,0.135) (63,0.216) (64,0.151) (65,0.133) (66,0.092) (67,0.119) (68,0.106) (69,0.113) (70,0.069)};
\addplot[stuV, line width=1.2pt, forget plot] coordinates {(71,0.339) (72,0.381) (73,0.443) (74,0.439) (75,0.527) (76,0.433) (77,0.368) (78,0.434) (79,0.443) (80,0.345) (81,0.385) (82,0.385) (83,0.388) (84,0.439) (85,0.445) (86,0.447) (87,0.504) (88,0.500) (89,0.518) (90,0.538) (91,0.488) (92,0.508) (93,0.443) (94,0.450) (95,0.446) (96,0.444) (97,0.441) (98,0.476) (99,0.414) (100,0.416) (101,0.377) (102,0.350) (103,0.298) (104,0.286) (105,0.277)};
\addplot[tchV, line width=1.2pt, forget plot] coordinates {(71,0.225) (72,0.292) (73,0.331) (74,0.337) (75,0.440) (76,0.328) (77,0.267) (78,0.328) (79,0.328) (80,0.246) (81,0.292) (82,0.278) (83,0.317) (84,0.363) (85,0.397) (86,0.421) (87,0.493) (88,0.508) (89,0.555) (90,0.558) (91,0.520) (92,0.531) (93,0.440) (94,0.439) (95,0.412) (96,0.426) (97,0.449) (98,0.501) (99,0.431) (100,0.427) (101,0.409) (102,0.391) (103,0.325) (104,0.304) (105,0.267)};
\addlegendimage{lineO, line width=1.0pt, draw opacity=0.45, mark=*, mark size=1.7pt, mark options={fill=lineO, draw=white, line width=0.4pt, fill opacity=0.45, draw opacity=0.45}}
\addlegendentry{Test F1 (ID)}
\addlegendimage{lineO, line width=1.0pt, draw opacity=0.45, mark=square*, mark size=1.5pt, mark options={fill=lineO, draw=white, line width=0.4pt, fill opacity=0.45, draw opacity=0.45}}
\addlegendentry{Test F1 (OOD)}
\end{axis}
\begin{axis}[scale only axis, width=4.98cm, height=4.0cm, xmin=0, xmax=106, tick label style={font=\scriptsize}, label style={font=\footnotesize}, clip=false,
  ymin=0, ymax=60, ytick={10,20,30,40,50,60},
  axis x line=none, axis y line*=right, axis line style={-, lineO!80}, tick style={lineO!80}, major tick length=2pt,
  yticklabel style={text=lineO}, ylabel={Test F1}, ylabel style={text=lineO},
]
\addplot[name path=SUID, draw=none, forget plot] table[x=x, y expr=\thisrow{m}+\thisrow{sd}] {
x  m  sd  l
0  23.4  1.82  23.4
35  25.2  1.88  25.2
70  40.1  2.19  40.1
105  44.2  2.11  44.2
};
\addplot[name path=SLID, draw=none, forget plot] table[x=x, y expr=\thisrow{m}-\thisrow{sd}] {
x  m  sd  l
0  23.4  1.82  23.4
35  25.2  1.88  25.2
70  40.1  2.19  40.1
105  44.2  2.11  44.2
};
\addplot[lineO, fill opacity=0.08, draw=none, forget plot] fill between[of=SUID and SLID];
\addplot[lineO, line width=1.0pt, draw opacity=0.45, mark=*, mark size=1.7pt, mark options={fill=lineO, draw=white, line width=0.4pt, fill opacity=0.45, draw opacity=0.45},
  nodes near coords, point meta=explicit symbolic, every node near coord/.append style={font=\tiny, text=black!75, anchor=north, yshift=-2pt, fill=white, fill opacity=0.8, text opacity=1, inner sep=0.6pt}]
  table[x=x, y=m, meta=l] {
x  m  sd  l
0  23.4  1.82  23.4
35  25.2  1.88  25.2
70  40.1  2.19  40.1
105  44.2  2.11  44.2
};
\addplot[name path=SUOOD, draw=none, forget plot] table[x=x, y expr=\thisrow{m}+\thisrow{sd}] {
x  m  sd  l
0  30.1  0.81  30.1
35  36.7  0.81  36.7
70  47.8  0.78  47.8
105  47.2  0.77  47.2
};
\addplot[name path=SLOOD, draw=none, forget plot] table[x=x, y expr=\thisrow{m}-\thisrow{sd}] {
x  m  sd  l
0  30.1  0.81  30.1
35  36.7  0.81  36.7
70  47.8  0.78  47.8
105  47.2  0.77  47.2
};
\addplot[lineO, fill opacity=0.08, draw=none, forget plot] fill between[of=SUOOD and SLOOD];
\addplot[lineO, line width=1.0pt, draw opacity=0.45, mark=square*, mark size=1.5pt, mark options={fill=lineO, draw=white, line width=0.4pt, fill opacity=0.45, draw opacity=0.45}]
  table[x=x, y=m] {
x  m  sd  l
0  30.1  0.81  30.1
35  36.7  0.81  36.7
70  47.8  0.78  47.8
105  47.2  0.77  47.2
};
\begin{scope}[every node/.append style={font=\tiny, text=black!75, fill=white, fill opacity=0.8, text opacity=1, inner sep=0.6pt}]
\node[anchor=south, yshift=1.5pt] at (axis cs:0,30.1) {30.1};
\node[anchor=south, yshift=1.5pt] at (axis cs:35,36.7) {36.7};
\node[anchor=north west, xshift=1.5pt, yshift=-1.5pt] at (axis cs:70,47.8) {47.8};
\node[anchor=south, yshift=1.5pt] at (axis cs:105,47.2) {47.2};
\end{scope}
\end{axis}
\end{tikzpicture}
\caption{Training reward of RoDeL-Sim with Qwen3.5-4B at each step of the three simultaneous updates (left axis) and average test F1 after each update (right axis). Each step gives a student reward and a teacher reward, which are computed differently and are thus not comparable with each other. Lines and shaded bands are as in \Cref{fig:train_reward}.}
\label{fig:sim_reward}
\end{minipage}
\end{figure}

\Cref{fig:train_reward,fig:sim_reward} show the training reward at each step of the two variants together with the average test F1 after each update. A student step is rewarded by the mean reward $r$ of its sampled responses (\Cref{eq:reward}), and a teacher step by the mean reward $R_i$ of its answer sets (\Cref{eq:set_reward}). In RoDeL-Sim, every step gives both rewards.

In RoDeL-Seq, the reward rises most in the second round, from $-0.55$ to $+0.36$ in $s_2$ and from $-0.19$ to $+0.23$ in $t_2$. Each update trains one role against the latest version of the other, so the joint rise in the second round indicates that, after adapting to each other in the first round, the student and the teacher co-evolve. The test F1 closely follows this trend. The second round raises its average by 10.8 points in domain and 14.4 out of domain.

In RoDeL-Sim, both rewards rise during $u_1$ and $u_2$ and stabilize in $u_3$. The student reward goes from $-0.86$ to $-0.46$ in $u_1$ and from $-0.10$ to $+0.32$ in $u_2$, and it stays at $+0.40$ in $u_3$. The teacher reward goes from $-1.56$ to $-1.04$ in $u_1$, from $-0.52$ to $+0.16$ in $u_2$, and from $+0.29$ to $+0.42$ in $u_3$. The rewards $R_i$ come from the responses of the student that is trained in the same steps, so a rise of the teacher reward reflects the progress of both roles. The test F1 moves with the rewards: its average rises by 16.7 points in domain and 17.7 out of domain over $u_1$ and $u_2$, and changes by $+4.1$ and $-0.6$ in $u_3$.

In both variants, the trend of the F1 on the test documents thus agrees closely with that of the reward on the training documents. The F1 gains most where the reward rises most, in the second round of RoDeL-Seq and in the first two updates of RoDeL-Sim, and gains little afterwards.

\subsection{Efficiency}
\label{subsec:efficiency}

Training takes 35.8 hours of wall-clock time for RoDeL-Seq and 37.2 hours for RoDeL-Sim on the eight-GPU node described in \Cref{app:impl}. In RoDeL-Seq, the three teacher updates take 12.5, 9.5, and 11.4 hours, and the three student updates take 0.9, 0.8, and 0.7 hours. The three simultaneous updates of RoDeL-Sim take 15.3, 12.9, and 9.0 hours. On the same node, SEAL-style takes 66.4 hours for its six epochs, and PaST-style takes 2.5 hours for its six epochs.

Writing a memory adapter takes two steps (\Cref{subsec:context_distill}): the teacher synthesizes answers to the synthetic questions of the document, and the memory adapter is distilled on them. \Cref{tab:write_perset} gives the time of both steps per test document for the two variants. Averaged over the 1,542 test documents, writing takes 36.9 seconds per document for RoDeL-Seq and 34.3 for RoDeL-Sim, most of it for distillation, which sums to 15.8 and 14.7 hours over all test documents. The time grows with the number of synthetic questions per document, from 8.3 seconds for RoDeL-Seq and 6.1 for RoDeL-Sim on SQuAD, with 18.6 questions per document, to 634.4 and 819.9 seconds on LoCoMo, with 1,136.

\begin{table}[!htb]
\centering
\footnotesize
\setlength{\tabcolsep}{2.8pt}
\caption{Time of writing one memory adapter with Qwen3.5-4B, in seconds per test document, for RoDeL-Seq (teacher after $t_2$) and RoDeL-Sim (teacher after $u_3$). Questions: number of synthetic questions per document. All: average over the 1,542 test documents. Datasets are ordered by the length of their documents, as in \Cref{tab:efficiency}.}
\label{tab:write_perset}
\begin{tabular}{@{}ll rrrrrrrrrr r@{}}
\toprule
Step & Method & SQuAD & 2Wiki & HotpotQA & MuSiQue & Qasper & TriviaQA & NQ & MFQA & NarQA & LoCoMo & All \\
\midrule
Questions & & 18.6 & 37.7 & 37.5 & 48.4 & 53.3 & 83.4 & 84.8 & 39.7 & 347.1 & 1136.0 & 68.3 \\
\addlinespace
Synthesis (s) & RoDeL-Seq & 0.4 & 0.9 & 1.3 & 1.2 & 2.2 & 2.4 & 3.4 & 2.1 & 33.9 & 150.2 & 3.8 \\
 & RoDeL-Sim & 0.3 & 0.7 & 1.2 & 0.9 & 1.3 & 1.5 & 2.1 & 1.4 & 16.5 & 85.7 & 2.2 \\
\addlinespace
Distillation (s) & RoDeL-Seq & 7.9 & 16.2 & 16.9 & 22.7 & 33.1 & 39.3 & 42.5 & 22.7 & 166.8 & 484.2 & 33.1 \\
 & RoDeL-Sim & 5.8 & 16.3 & 15.4 & 19.9 & 26.2 & 35.3 & 36.6 & 19.6 & 180.7 & 734.2 & 32.0 \\
\addlinespace
Total (s) & RoDeL-Seq & 8.3 & 17.1 & 18.2 & 23.8 & 35.3 & 41.7 & 45.9 & 24.8 & 200.8 & 634.4 & 36.9 \\
 & RoDeL-Sim & 6.1 & 17.0 & 16.6 & 20.7 & 27.5 & 36.9 & 38.7 & 21.0 & 197.2 & 819.9 & 34.3 \\
\bottomrule
\end{tabular}
\end{table}

In RoDeL, reading the memory adapter takes the place of having the document in the context. \Cref{tab:efficiency} compares the cost of answering with RoDeL against that of the base model $\theta$ with the whole document in its context, the token-memory counterpart (In-context). The comparison covers all ten datasets, listed by increasing prompt length of In-context. The two variants answer in the same way (\Cref{subsec:inference}), and the table lists the cost of each.

The prompt of RoDeL holds the instruction and the question, whereas the prompt of In-context grows with the document. The peak memory of In-context grows accordingly, reaching 5.8 times that of either variant on LoCoMo, while the memory of both variants stays nearly constant across datasets.

\begin{table}[!htb]
\centering
\footnotesize
\setlength{\tabcolsep}{3.4pt}
\caption{Cost of answering test questions with Qwen3.5-4B for In-context, the base model $\theta$ with the whole document in its context, prompted with the teacher prompt $x_T(D,q)$, and for the two variants of RoDeL. Datasets are ordered by the prompt length of In-context. Prompt and Time are averaged per question; GPU Peak is the peak GPU memory allocated.}
\label{tab:efficiency}
\begin{tabular}{@{}llrrrrrrrrrr@{}}
\toprule
Metric & Method & SQuAD & 2Wiki & HotpotQA & MuSiQue & Qasper & TriviaQA & NQ & MFQA & NarQA & LoCoMo \\
\midrule
Prompt (tokens) & In-context & 0.3k & 1.1k & 1.5k & 2.7k & 5.0k & 6.6k & 7.1k & 7.1k & 16.9k & 23.5k \\
 & RoDeL-Seq & 0.1k & 0.1k & 0.1k & 0.1k & 0.1k & 0.1k & 0.1k & 0.1k & 0.1k & 0.1k \\
 & RoDeL-Sim & 0.1k & 0.1k & 0.1k & 0.1k & 0.1k & 0.1k & 0.1k & 0.1k & 0.1k & 0.1k \\
\addlinespace
Time (s) & In-context & 5.2 & 24.2 & 21.2 & 28.9 & 12.7 & 12.0 & 25.4 & 22.7 & 5.2 & 10.9 \\
 & RoDeL-Seq & 3.6 & 18.9 & 10.8 & 25.5 & 10.6 & 7.6 & 15.0 & 15.9 & 4.6 & 4.5 \\
 & RoDeL-Sim & 3.0 & 19.7 & 17.9 & 25.9 & 9.2 & 6.7 & 12.3 & 14.6 & 4.2 & 4.5 \\
\addlinespace
GPU Peak (GiB) & In-context & 9.8 & 8.4 & 8.3 & 8.6 & 17.3 & 14.2 & 14.3 & 13.3 & 26.8 & 59.8 \\
 & RoDeL-Seq & 10.3 & 8.7 & 8.7 & 8.7 & 9.2 & 9.1 & 8.9 & 8.9 & 10.2 & 10.3 \\
 & RoDeL-Sim & 10.3 & 8.7 & 8.7 & 8.7 & 9.2 & 9.1 & 8.9 & 8.9 & 10.3 & 10.3 \\
\bottomrule
\end{tabular}
\end{table}

\subsection{Generality}
\label{subsec:generality}

Both variants are also trained with three other base models, on the training documents and synthetic questions of the main experiments and with the same three rounds of updates. With Qwen3.5-2B, a smaller model of the same family (\Cref{app:sizes}), training both roles raises the average F1 from 8.2 to 31.5 in domain and from 15.4 to 40.7 out of domain with RoDeL-Seq, and to 30.0 and 40.2 with RoDeL-Sim. With Qwen3.5-0.8B, an even smaller model of that family, it raises the average F1 from 3.7 to 13.6 in domain and from 4.7 to 23.6 out of domain with RoDeL-Seq, and to 13.8 and 23.7 with RoDeL-Sim. With Gemma-4-E2B~\citep{gemmateam2026gemma4technicalreport}, a base model from another model family (\Cref{app:gemma}), it raises the average F1 from 16.1 to 27.0 in domain and from 31.4 to 39.6 out of domain with RoDeL-Seq, and to 24.1 and 40.5 with RoDeL-Sim. All of these gains over the frozen roles (PRAG-style) have $p<0.05$. Compared with the baselines trained with Gemma-4-E2B, each method at its best point, both variants have a higher out-of-domain average than every baseline, and RoDeL-Seq also has the highest in-domain average (\Cref{tab:main_gemma}). These results show that, with either variant, training both roles improves over the frozen roles across three model sizes and two model families.

\section{Conclusion}
\label{sec:conclusion}

We presented RoDeL, a role-decoupled learning method for context distillation that trains the teacher and the student as two separate adapters on a frozen base model. The student learns with reinforcement learning to answer questions by reading memory adapters written by the teacher, and the teacher learns with SR-GRPO from rewards relayed by the student, so that each role adapts to the current version of the other. The two variants of RoDeL differ in when the adapters are updated: RoDeL-Seq alternates between student updates and teacher updates, with the other role frozen during each, and RoDeL-Sim updates both adapters in every training step. On each of two in-domain and eight out-of-domain datasets, RoDeL-Seq or RoDeL-Sim outperforms representative baselines, and both variants outperform them on average. The same holds in long-context retrieval scenarios.
Further, RoDeL shows that, in token-to-parameter memory, how information is written into parameters and how it is read from them can be learned together, each adapting to the other, whether the two roles are updated sequentially or simultaneously. This perspective may offer insights for context distillation and other ways of writing new information into parameters, such as test-time training, parametric RAG, knowledge editing, and continual learning through memory consolidation~\citep{behrouz2026sleep}.

Role decoupling also allows the teacher and the student to use different base models. The two roles are connected by the answer sets that the teacher synthesizes, which enter distillation as text, and, during training, by the rewards of the student's responses, so the teacher adapter can be attached to a different base model from that of the memory and student adapters, e.g., a larger model that writes for a smaller one. In SEAL, by contrast, a single model plays both roles~\citep{zweiger2025self}. Experiments in this paper are conducted within context self-distillation, where the teacher and the student share one base model, and this work lays the foundation for training the two roles on different base models.

\FloatBarrier
\bibliography{custom}
\bibliographystyle{unsrtnat}

\clearpage
\beginappendix
\section{Datasets}
\label{app:data}

\Cref{tab:datasets} summarizes all datasets. A document's text is the concatenation of its passages, each with its title. Gold answers include their aliases.

\begin{table}[!htb]
\centering
\footnotesize
\setlength{\tabcolsep}{5.0pt}
\caption{Datasets. Doc length: median number of tokens per document, counted with the Qwen3.5 tokenizer. \#Q: number of training questions for the training data and of test questions otherwise. \#Syn.\ Q: average number of synthetic questions per document.}
\label{tab:datasets}
\begin{tabular}{@{}lllrrrr@{}}
\toprule
 & Dataset & Documents & \#Docs & Doc length & \#Q & \#Syn.\ Q \\
\midrule
\multirow{2}{*}{Training} & HotpotQA & 10 passages of one training-split question & 180 & 1,418 & 180 & 37.7 \\
 & MuSiQue & 20 passages of one training-split question & 90 & 2,331 & 90 & 48.2 \\
\midrule
\multirow{2}{*}{In-domain} & HotpotQA & 10 passages of one question (distractor setting) & 200 & 1,285 & 200 & 37.5 \\
 & MuSiQue & 20 passages of one question & 200 & 2,357 & 200 & 48.4 \\
\midrule
\multirow{8}{*}{Out-of-domain} & SQuAD & one paragraph & 200 & 161 & 1,118 & 18.6 \\
 & 2Wiki & 10 passages of one question & 200 & 862 & 200 & 37.7 \\
 & NQ & full Wikipedia page & 200 & 5,696 & 218 & 84.8 \\
 & TriviaQA & full Wikipedia page & 200 & 5,502 & 643 & 83.4 \\
 & Qasper & NLP paper & 200 & 4,574 & 549 & 53.3 \\
 & MultiFieldQA & LongBench MultiFieldQA-en document & 80 & 7,539 & 105 & 39.7 \\
 & NarrativeQA & 30 books and 22 movie scripts & 52 & 17,367 & 1,568 & 347.1 \\
 & LoCoMo & multi-session conversation & 10 & 24,440 & 1,540 & 1,136.0 \\
\bottomrule
\end{tabular}
\end{table}

\paragraph{Training documents.} Training documents have disjoint passages. All HotpotQA training questions are bridge questions; MuSiQue training questions take 2--4 hops. Another 20 HotpotQA and 10 MuSiQue documents, built in the same way, are held out to monitor training.

\paragraph{Synthetic questions.} Qwen3-4B in its thinking mode generates the synthetic questions of each document from the document alone. They stay fixed throughout training and evaluation. Their average number per document ranges from 18.6 on SQuAD to 1,136.0 on LoCoMo (\Cref{tab:datasets}).

\paragraph{Test questions.} Only answerable questions whose shortest gold answer has fewer than 20 tokens are kept. The in-domain HotpotQA questions comprise 160 bridge and 40 comparison questions, and the MuSiQue questions comprise 103, 63, and 34 questions of 2, 3, and 4 hops. LoCoMo excludes adversarial questions, and Qasper excludes questions that all annotators mark as unanswerable. Test documents of HotpotQA (distractor setting), MuSiQue, SQuAD v1.1, 2Wiki, and NQ come from the publicly released evaluation splits of these datasets, and those of Qasper from its test split. TriviaQA documents come from its training split and are used for evaluation.

\section{Baselines}
\label{app:baselines}

All baselines use the base model, synthetic questions, memory-adapter configuration, distillation, decoding, and rewards of RoDeL (\Cref{app:impl}). They keep the structure of each original method, namely which roles are trained and whether they share parameters, and differ from the originals as follows.

\paragraph{PRAG-style.} Parametric RAG~\citep{su2025parametric} trains each document's LoRA on rewrites of the document and generated question--answer pairs, and combines it with retrieval. In this work, for a fair comparison, the untrained base model writes each memory adapter through the synthesis and distillation of \Cref{subsec:context_distill} and reads it.

\paragraph{SEAL-style.} SEAL~\citep{zweiger2025self} trains a model to generate its own fine-tuning data (self-edits) with rejection-sampling-based self-training. The same model is then fine-tuned on these data, so one set of parameters both writes and reads. In our SEAL-style variant, the teacher and the student therefore share one LoRA and have no separate $\phi_s$. This LoRA is trained for six epochs by RoDeL-Seq's teacher update with the same hyperparameters, with two changes: each memory adapter is distilled onto the base model with the shared LoRA attached instead of the base model alone, and the student reads it with both the shared LoRA and the memory adapter attached. Updates of the shared LoRA thus also change the student. At test time, memory adapters are written and read in the same way.

\paragraph{PaST-style.} PaST~\citep{tang-etal-2026-knowledge} obtains a skill vector as the parameter difference of a model trained with reinforcement learning on a source domain, and adds it to a model fine-tuned on new data. It trains on about ten synthetic questions plus the original question per document, with rewards judged by GPT-4.1. In our PaST-style variant, the skill is the student adapter $\phi_s$, trained by RoDeL-Seq's student update with the same hyperparameters on the single training question of each document with the F1 reward; knowledge is injected by the memory adapters written by the base model. At test time, $\phi_s$ is attached with scale 1.

\section{Implementation Details}
\label{app:impl}

\paragraph{Hardware.} Unless stated otherwise, all training, evaluation, data construction, and baseline runs used a single node with eight GPUs, each providing 141\,GB of HBM3e memory.

\paragraph{Hyperparameters and prompts.} \Cref{tab:hparams} lists the hyperparameters. The teacher and the student use the chat template with thinking mode enabled, and their prompts are shown below. The teacher prompt begins with a sentence that states the document type, \promptvar{intro}, e.g., ``Below is a set of documents.''

\begin{table}[!htb]
\centering
\small
\caption{Hyperparameters of RoDeL.}
\label{tab:hparams}
\begin{tabular}{@{}lp{11.2cm}@{}}
\toprule
Item & Setting \\
\midrule
Base model $\theta$ & Qwen3.5-4B in the main experiments (text model only), thinking mode, frozen \\
Question generator & Qwen3-4B, thinking mode, sees the document only \\
Answer $\mathrm{ans}(y)$ & First non-empty line after the reasoning trace \\
Generation limit & 1024 tokens for all generation (synthesis, student responses, evaluation) \\
LoRA targets & All linear layers: attention q/k/v/o, MLP gate/up/down, and the five projections of linear-attention layers \\
$\phi_t$, $\phi_s$ & Rank 64, $\alpha=128$, no dropout \\
$\phi_m$ & Rank 16, $\alpha=32$, dropout 0.05 \\
Distill & Cross-entropy on valid responses; AdamW, learning rate $2\times10^{-4}$, weight decay 0.01, batch size 4, 5 epochs, gradient clipping 1.0 \\
RL optimization & AdamW, constant learning rate $10^{-5}$, no weight decay, gradient clipping 1.0; one gradient step per batch; optimizer re-initialized at each update; each update passes once over $\mathcal D$ and keeps its last step \\
RoDeL & $K=3$, $M=4$, $G=4$, $c=2$, $\lambda=2$, $\epsilon=10^{-4}$; each teacher-update batch is drawn from a single source dataset \\
Reward F1 & Token-level F1 after answer normalization, maximum over gold-answer aliases \\
Sampling for RL & Temperature 1.0, no top-$p$/top-$k$ truncation (teacher answer sets in teacher updates, student responses in student updates) \\
Other decoding & Greedy for the frozen student in teacher updates; temperature 0.6, top-$p=0.95$, top-$k=20$ for all other synthesis and for test-time answering; one response per synthetic question \\
\bottomrule
\end{tabular}
\end{table}

\begin{promptbox}{Student Prompt $x(q)$}
Answer the following question. Give a short factual answer (a few words). For yes/no questions, answer exactly yes or no. If the question asks for a date or a number, answer with the date or the number.

\smallskip
Question: \promptvar{question}
\end{promptbox}

\begin{promptbox}{Teacher Prompt $x_T(D,q)$}
\promptvar{intro}

\smallskip
\promptvar{document}

\smallskip
Answer the following question. Give a short factual answer (a few words). For yes/no questions, answer exactly yes or no. If the question asks for a date or a number, answer with the date or the number.

\smallskip
Question: \promptvar{question}
\end{promptbox}

\section{RoDeL across Training Updates}
\label{app:results}

\paragraph{Significance tests.} Each comparison between two systems on a dataset is tested with a paired bootstrap over test documents, using 10,000 resamples and two-sided $p$-values. A difference on a dataset is significant when it stays below 0.05 after Holm correction across the datasets of its group (two in-domain or eight out-of-domain). Tables mark $p<0.05$ with $^*$ and differences that remain significant after Holm correction with $^\dagger$. Differences in average F1 are tested with a paired bootstrap with 2,000 resamples and are significant when $p<0.05$. The bands in \Cref{fig:dynamics,fig:avgsim} show one standard deviation of average F1 under the same per-dataset resampling, with 10,000 resamples.

\paragraph{Per-dataset results.} \Cref{fig:perset4b} plots the F1 of RoDeL-Seq with Qwen3.5-4B on each dataset after each update, and \Cref{tab:steps4b} lists the change made by each update. \Cref{fig:persetsim} plots the F1 of RoDeL-Sim after each simultaneous update.

\begin{table}[!htb]
\centering
\small
\setlength{\tabcolsep}{7.5pt}
\caption{Change in F1 made by each update of RoDeL-Seq with Qwen3.5-4B, the total change from the frozen roles to the end of training ($t_3$ vs.\ Frozen), and the change to the best point of each row (Best vs.\ Frozen; the best point is chosen on the test data and shown for reference). $^*$:~$p<0.05$; $^\dagger$:~significant after Holm correction across the datasets of the group. Averages are tested once per comparison and carry only $^*$.}
\label{tab:steps4b}
\begin{tabular}{@{}lrrrrrr@{\hspace{18pt}}r@{\hspace{18pt}}cr@{\hspace{8pt}}}
\toprule
Dataset & \textcolor{stuV}{$s_1$} & \textcolor{tchV}{$t_1$} & \textcolor{stuV}{$s_2$} & \textcolor{tchV}{$t_2$} & \textcolor{stuV}{$s_3$} & \textcolor{tchV}{$t_3$} & $t_3$ vs.\ Frozen & Best & Best vs.\ Frozen \\
\midrule
HotpotQA & $-$1.6 & +0.7 & +4.1 & +9.8\rlap{$^{*\dagger}$} & $-$4.1 & +0.6 & +9.4\rlap{$^{*\dagger}$} & \textcolor{tchV}{$t_2$} & +12.9\rlap{$^{*\dagger}$} \\
MuSiQue & +3.9\rlap{$^{*}$} & +3.3 & +8.7\rlap{$^{*\dagger}$} & $-$0.9 & +5.5\rlap{$^{*}$} & $-$2.8 & +17.6\rlap{$^{*\dagger}$} & \textcolor{stuV}{$s_3$} & +20.5\rlap{$^{*\dagger}$} \\
\textit{Average} & +1.1 & +2.0 & +6.4\rlap{$^{*}$} & +4.5\rlap{$^{*}$} & +0.7 & $-$1.1 & +13.5\rlap{$^{*}$} & \textcolor{stuV}{$s_3$} & +14.6\rlap{$^{*}$} \\
\midrule
SQuAD & +2.4\rlap{$^{*}$} & +1.5 & +4.4\rlap{$^{*\dagger}$} & +9.6\rlap{$^{*\dagger}$} & $-$0.2 & $-$3.8\rlap{$^{*\dagger}$} & +14.0\rlap{$^{*\dagger}$} & \textcolor{tchV}{$t_2$} & +17.9\rlap{$^{*\dagger}$} \\
2Wiki & $-$4.2 & +6.5 & +6.6\rlap{$^{*\dagger}$} & +0.8 & +0.6 & $-$8.7\rlap{$^{*}$} & +1.7 & \textcolor{stuV}{$s_3$} & +10.4\rlap{$^{*\dagger}$} \\
NQ & +2.3 & $-$1.3 & +14.2\rlap{$^{*\dagger}$} & +1.8 & $-$2.5 & $-$1.4 & +13.2\rlap{$^{*\dagger}$} & \textcolor{tchV}{$t_2$} & +17.0\rlap{$^{*\dagger}$} \\
TriviaQA & +2.6 & +1.7 & +5.5\rlap{$^{*\dagger}$} & +4.2\rlap{$^{*}$} & $-$2.4 & $-$2.0 & +9.6\rlap{$^{*\dagger}$} & \textcolor{tchV}{$t_2$} & +14.0\rlap{$^{*\dagger}$} \\
Qasper & +3.3\rlap{$^{*\dagger}$} & +1.2 & +6.8\rlap{$^{*\dagger}$} & +12.1\rlap{$^{*\dagger}$} & $-$2.2 & +0.1 & +21.2\rlap{$^{*\dagger}$} & \textcolor{tchV}{$t_2$} & +23.3\rlap{$^{*\dagger}$} \\
MultiFieldQA & +4.5 & $-$2.4 & +7.1\rlap{$^{*\dagger}$} & +7.8\rlap{$^{*\dagger}$} & $-$1.2 & $-$0.7 & +15.0\rlap{$^{*\dagger}$} & \textcolor{tchV}{$t_2$} & +17.0\rlap{$^{*\dagger}$} \\
NarrativeQA & +2.0\rlap{$^{*\dagger}$} & +0.8 & +7.1\rlap{$^{*\dagger}$} & +5.0\rlap{$^{*\dagger}$} & $-$0.6 & +1.3 & +15.6\rlap{$^{*\dagger}$} & \textcolor{tchV}{$t_3$} & +15.6\rlap{$^{*\dagger}$} \\
LoCoMo & +2.0\rlap{$^{*\dagger}$} & +1.2 & +7.6\rlap{$^{*\dagger}$} & +14.8\rlap{$^{*\dagger}$} & $-$0.8 & +2.8\rlap{$^{*}$} & +27.6\rlap{$^{*\dagger}$} & \textcolor{tchV}{$t_3$} & +27.6\rlap{$^{*\dagger}$} \\
\textit{Average} & +1.9\rlap{$^{*}$} & +1.2 & +7.4\rlap{$^{*}$} & +7.0\rlap{$^{*}$} & $-$1.2 & $-$1.5\rlap{$^{*}$} & +14.8\rlap{$^{*}$} & \textcolor{tchV}{$t_2$} & +17.4\rlap{$^{*}$} \\
\bottomrule
\end{tabular}
\end{table}

\section{Other Model Sizes}
\label{app:sizes}

Both variants are also trained with Qwen3.5-2B and Qwen3.5-0.8B as the base model, using the recipe, number of steps, hyperparameters, and synthetic questions of the main experiments.

\begin{figure}[!htb]
\begin{minipage}[t]{\colw}
\centering
\pgfplotstableread{
x  id    idsd  ood   oodsd
0  8.2  1.26  15.4  0.66
1  15.6  1.63  19.9  0.71
2  25.6  1.91  33.0  0.79
3  26.0  1.91  36.5  0.82
4  26.8  1.93  39.9  0.83
5  27.2  1.90  40.3  0.83
6  31.5  1.97  40.7  0.85
}\dyntab
\begin{tikzpicture}
\begin{axis}[
  width=\colw, height=5.2cm,
  xmin=-0.5, xmax=6.5, ymin=-2, ymax=56,
  xtick={0,...,6},
  xticklabels={Frozen,\textcolor{stuV}{$s_1$},\textcolor{tchV}{$t_1$},\textcolor{stuV}{$s_2$},\textcolor{tchV}{$t_2$},\textcolor{stuV}{$s_3$},\textcolor{tchV}{$t_3$}},
  ytick={0,10,20,30,40,50},
  ylabel={Average F1},
  axis lines=left, axis line style={-, black!60}, tick style={black!60},
  tick label style={font=\small}, label style={font=\small},
  ymajorgrids, grid style={black!22, dashed},
  set layers, clip=false,
  legend style={font=\small, at={(0.5,1.03)}, anchor=south, draw=none, fill=none, legend columns=2, /tikz/every even column/.append style={column sep=10pt}},
  legend cell align=left,
]
\begin{pgfonlayer}{axis background}
\draw[black!25] (axis cs:0.5,-2) -- (axis cs:0.5,56);
\draw[black!25] (axis cs:2.5,-2) -- (axis cs:2.5,56);
\draw[black!25] (axis cs:4.5,-2) -- (axis cs:4.5,56);
\end{pgfonlayer}
\node[font=\scriptsize, text=black!55, anchor=north] at (axis cs:1.5,56) {Round 1};
\node[font=\scriptsize, text=black!55, anchor=north] at (axis cs:3.5,56) {Round 2};
\node[font=\scriptsize, text=black!55, anchor=north] at (axis cs:5.5,56) {Round 3};
\addplot[name path=idU, draw=none, forget plot] table[x=x, y expr=\thisrow{id}+\thisrow{idsd}] {\dyntab};
\addplot[name path=idL, draw=none, forget plot] table[x=x, y expr=\thisrow{id}-\thisrow{idsd}] {\dyntab};
\addplot[lineO, fill opacity=0.15, draw=none, forget plot] fill between[of=idU and idL];
\addplot[name path=oodU, draw=none, forget plot] table[x=x, y expr=\thisrow{ood}+\thisrow{oodsd}] {\dyntab};
\addplot[name path=oodL, draw=none, forget plot] table[x=x, y expr=\thisrow{ood}-\thisrow{oodsd}] {\dyntab};
\addplot[lineO, fill opacity=0.15, draw=none, forget plot] fill between[of=oodU and oodL];
\addplot[lineO, line width=1.2pt, mark=*, mark size=2.3pt, mark options={fill=lineO, draw=white, line width=0.6pt},
  nodes near coords, point meta=explicit symbolic, every node near coord/.append style={font=\scriptsize, text=black!70, anchor=north, yshift=-4pt}]
  table[x=x, y=id, meta=id] {\dyntab};
\addlegendentry{In-domain}
\addplot[lineO, line width=1.2pt, mark=square*, mark size=2.1pt, mark options={fill=lineO, draw=white, line width=0.6pt},
  nodes near coords, point meta=explicit symbolic, every node near coord/.append style={font=\scriptsize, text=black!70, anchor=south, yshift=2pt}]
  table[x=x, y=ood, meta=ood] {\dyntab};
\addlegendentry{Out-of-domain}
\end{axis}
\end{tikzpicture}
\caption{Average F1 of RoDeL-Seq with Qwen3.5-2B after each update. Shaded bands show $\pm 1$ standard deviation of a test-set bootstrap.}
\label{fig:avg2b}
\end{minipage}\hfill
\begin{minipage}[t]{\colw}
\centering
\pgfplotstableread{
x  id    idsd  ood   oodsd
0  8.2  1.26  15.4  0.66
1  26.3  1.94  34.8  0.81
2  27.8  1.91  39.0  0.84
3  30.0  1.94  40.2  0.80
}\simtab
\begin{tikzpicture}
\begin{axis}[
  width=\colw, height=5.2cm,
  xmin=-0.5, xmax=3.5, ymin=-2, ymax=56,
  xtick={0,...,3},
  xticklabels={Frozen,$u_1$,$u_2$,$u_3$},
  ytick={0,10,20,30,40,50},
  ylabel={Average F1},
  axis lines=left, axis line style={-, black!60}, tick style={black!60},
  tick label style={font=\small}, label style={font=\small},
  ymajorgrids, grid style={black!22, dashed},
  clip=false,
  legend style={font=\small, at={(0.5,1.03)}, anchor=south, draw=none, fill=none, legend columns=2, /tikz/every even column/.append style={column sep=10pt}},
  legend cell align=left,
]
\addplot[name path=idU, draw=none, forget plot] table[x=x, y expr=\thisrow{id}+\thisrow{idsd}] {\simtab};
\addplot[name path=idL, draw=none, forget plot] table[x=x, y expr=\thisrow{id}-\thisrow{idsd}] {\simtab};
\addplot[lineO, fill opacity=0.15, draw=none, forget plot] fill between[of=idU and idL];
\addplot[name path=oodU, draw=none, forget plot] table[x=x, y expr=\thisrow{ood}+\thisrow{oodsd}] {\simtab};
\addplot[name path=oodL, draw=none, forget plot] table[x=x, y expr=\thisrow{ood}-\thisrow{oodsd}] {\simtab};
\addplot[lineO, fill opacity=0.15, draw=none, forget plot] fill between[of=oodU and oodL];
\addplot[lineO, line width=1.2pt, mark=*, mark size=2.3pt, mark options={fill=lineO, draw=white, line width=0.6pt},
  nodes near coords, point meta=explicit symbolic, every node near coord/.append style={font=\scriptsize, text=black!70, anchor=north, yshift=-4pt}]
  table[x=x, y=id, meta=id] {\simtab};
\addlegendentry{In-domain}
\addplot[lineO, line width=1.2pt, mark=square*, mark size=2.1pt, mark options={fill=lineO, draw=white, line width=0.6pt},
  nodes near coords, point meta=explicit symbolic, every node near coord/.append style={font=\scriptsize, text=black!70, anchor=south, yshift=2pt}]
  table[x=x, y=ood, meta=ood] {\simtab};
\addlegendentry{Out-of-domain}
\end{axis}
\end{tikzpicture}
\caption{Average F1 of RoDeL-Sim with Qwen3.5-2B after each simultaneous update. Shaded bands are as in \Cref{fig:avg2b}.}
\label{fig:avg2bsim}
\end{minipage}
\end{figure}

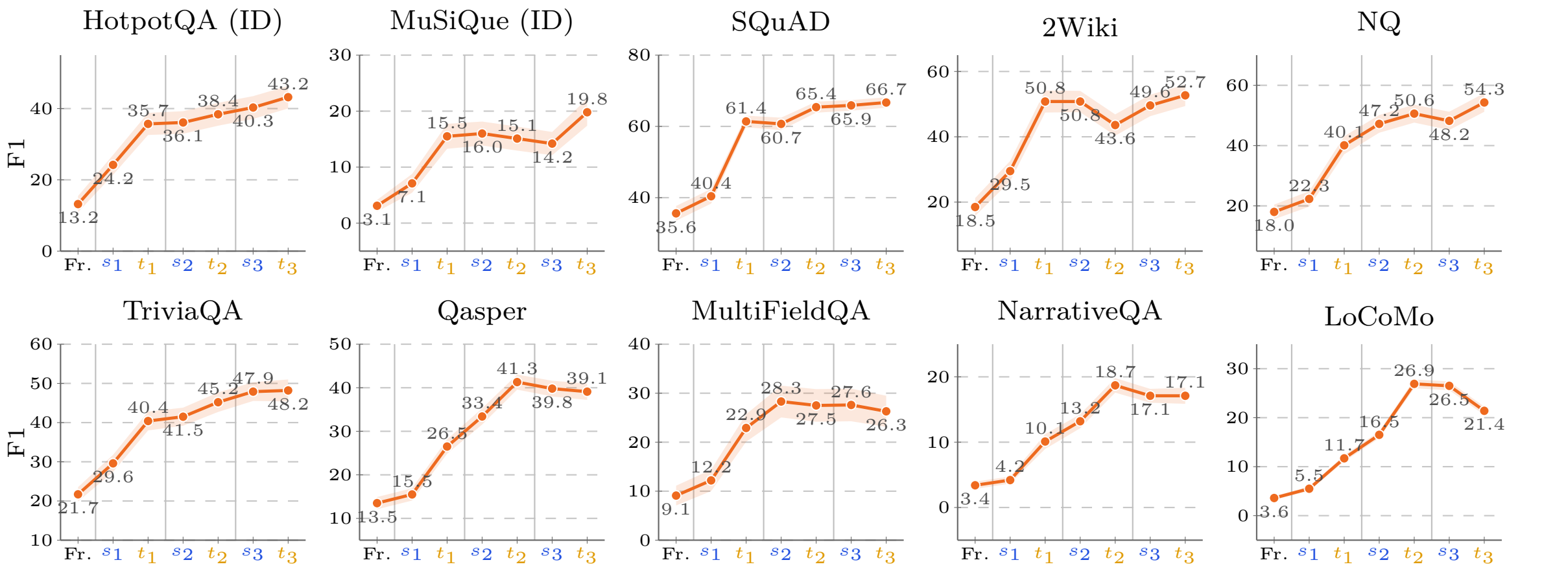
\begin{figure}[!htb]
\centering
\begin{tikzpicture}
\begin{groupplot}[
  group style={group size=5 by 2, horizontal sep=0.55cm, vertical sep=0.95cm},
  scale only axis, width=2.5cm, height=2.0cm,
  xmin=-0.5, xmax=6.5,
  xtick={0,...,6},
  xticklabels={Fr.,\textcolor{stuV}{$s_1$},\textcolor{tchV}{$t_1$},\textcolor{stuV}{$s_2$},\textcolor{tchV}{$t_2$},\textcolor{stuV}{$s_3$},\textcolor{tchV}{$t_3$}},
  axis lines=left, axis line style={-, black!60}, tick style={black!60}, major tick length=2pt,
  tick label style={font=\tiny, inner sep=1pt},
  title style={font=\footnotesize, yshift=-4pt},
  ymajorgrids, grid style={black!22, dashed},
  extra x ticks={0.5,2.5,4.5}, extra x tick labels={},
  extra x tick style={grid=major, major grid style={black!25, solid}, tick style={draw=none}},
  clip=false,
]
\nextgroupplot[title={HotpotQA (ID)}, ymin=0, ymax=55, ytick={0,20,40}, ylabel={F1}, ylabel style={font=\scriptsize, yshift=-3pt}]
\addplot[name path=U, draw=none, forget plot] table[x=x, y expr=\thisrow{m}+\thisrow{sd}] {
x  m  sd
0  13.2  2.28
1  24.2  2.79
2  35.7  3.11
3  36.1  3.17
4  38.4  3.21
5  40.3  3.20
6  43.2  3.16
};
\addplot[name path=L, draw=none, forget plot] table[x=x, y expr=\thisrow{m}-\thisrow{sd}] {
x  m  sd
0  13.2  2.28
1  24.2  2.79
2  35.7  3.11
3  36.1  3.17
4  38.4  3.21
5  40.3  3.20
6  43.2  3.16
};
\addplot[lineO, fill opacity=0.15, draw=none, forget plot] fill between[of=U and L];
\addplot[lineO, line width=0.9pt, mark=*, mark size=1.4pt, mark options={fill=lineO, draw=white, line width=0.35pt}] table[x=x, y=m] {
x  m  sd
0  13.2  2.28
1  24.2  2.79
2  35.7  3.11
3  36.1  3.17
4  38.4  3.21
5  40.3  3.20
6  43.2  3.16
};
\node[font=\tiny, text=black!70, anchor=north, yshift=-1.6pt, inner sep=0.5pt] at (axis cs:0,13.2) {13.2};
\node[font=\tiny, text=black!70, anchor=north, yshift=-1.6pt, inner sep=0.5pt] at (axis cs:1,24.2) {24.2};
\node[font=\tiny, text=black!70, anchor=south, yshift=1.6pt, inner sep=0.5pt] at (axis cs:2,35.7) {35.7};
\node[font=\tiny, text=black!70, anchor=north, yshift=-1.6pt, inner sep=0.5pt] at (axis cs:3,36.1) {36.1};
\node[font=\tiny, text=black!70, anchor=south, yshift=1.6pt, inner sep=0.5pt] at (axis cs:4,38.4) {38.4};
\node[font=\tiny, text=black!70, anchor=north, yshift=-1.6pt, inner sep=0.5pt] at (axis cs:5,40.3) {40.3};
\node[font=\tiny, text=black!70, anchor=south, yshift=1.6pt, inner sep=0.5pt] at (axis cs:6,43.2) {43.2};
\nextgroupplot[title={MuSiQue (ID)}, ymin=-5, ymax=30, ytick={0,10,20,30}]
\addplot[name path=U, draw=none, forget plot] table[x=x, y expr=\thisrow{m}+\thisrow{sd}] {
x  m  sd
0  3.1  1.05
1  7.1  1.68
2  15.5  2.23
3  16.0  2.14
4  15.1  2.15
5  14.2  2.08
6  19.8  2.41
};
\addplot[name path=L, draw=none, forget plot] table[x=x, y expr=\thisrow{m}-\thisrow{sd}] {
x  m  sd
0  3.1  1.05
1  7.1  1.68
2  15.5  2.23
3  16.0  2.14
4  15.1  2.15
5  14.2  2.08
6  19.8  2.41
};
\addplot[lineO, fill opacity=0.15, draw=none, forget plot] fill between[of=U and L];
\addplot[lineO, line width=0.9pt, mark=*, mark size=1.4pt, mark options={fill=lineO, draw=white, line width=0.35pt}] table[x=x, y=m] {
x  m  sd
0  3.1  1.05
1  7.1  1.68
2  15.5  2.23
3  16.0  2.14
4  15.1  2.15
5  14.2  2.08
6  19.8  2.41
};
\node[font=\tiny, text=black!70, anchor=north, yshift=-1.6pt, inner sep=0.5pt] at (axis cs:0,3.1) {3.1};
\node[font=\tiny, text=black!70, anchor=north, yshift=-1.6pt, inner sep=0.5pt] at (axis cs:1,7.1) {7.1};
\node[font=\tiny, text=black!70, anchor=south, yshift=1.6pt, inner sep=0.5pt] at (axis cs:2,15.5) {15.5};
\node[font=\tiny, text=black!70, anchor=north, yshift=-1.6pt, inner sep=0.5pt] at (axis cs:3,16.0) {16.0};
\node[font=\tiny, text=black!70, anchor=south, yshift=1.6pt, inner sep=0.5pt] at (axis cs:4,15.1) {15.1};
\node[font=\tiny, text=black!70, anchor=north, yshift=-1.6pt, inner sep=0.5pt] at (axis cs:5,14.2) {14.2};
\node[font=\tiny, text=black!70, anchor=south, yshift=1.6pt, inner sep=0.5pt] at (axis cs:6,19.8) {19.8};
\nextgroupplot[title={SQuAD}, ymin=25, ymax=80, ytick={40,60,80}]
\addplot[name path=U, draw=none, forget plot] table[x=x, y expr=\thisrow{m}+\thisrow{sd}] {
x  m  sd
0  35.6  2.07
1  40.4  2.07
2  61.4  1.62
3  60.7  1.75
4  65.4  1.55
5  65.9  1.49
6  66.7  1.46
};
\addplot[name path=L, draw=none, forget plot] table[x=x, y expr=\thisrow{m}-\thisrow{sd}] {
x  m  sd
0  35.6  2.07
1  40.4  2.07
2  61.4  1.62
3  60.7  1.75
4  65.4  1.55
5  65.9  1.49
6  66.7  1.46
};
\addplot[lineO, fill opacity=0.15, draw=none, forget plot] fill between[of=U and L];
\addplot[lineO, line width=0.9pt, mark=*, mark size=1.4pt, mark options={fill=lineO, draw=white, line width=0.35pt}] table[x=x, y=m] {
x  m  sd
0  35.6  2.07
1  40.4  2.07
2  61.4  1.62
3  60.7  1.75
4  65.4  1.55
5  65.9  1.49
6  66.7  1.46
};
\node[font=\tiny, text=black!70, anchor=north, yshift=-1.6pt, inner sep=0.5pt] at (axis cs:0,35.6) {35.6};
\node[font=\tiny, text=black!70, anchor=south, yshift=1.6pt, inner sep=0.5pt] at (axis cs:1,40.4) {40.4};
\node[font=\tiny, text=black!70, anchor=south, yshift=1.6pt, inner sep=0.5pt] at (axis cs:2,61.4) {61.4};
\node[font=\tiny, text=black!70, anchor=north, yshift=-1.6pt, inner sep=0.5pt] at (axis cs:3,60.7) {60.7};
\node[font=\tiny, text=black!70, anchor=south, yshift=1.6pt, inner sep=0.5pt] at (axis cs:4,65.4) {65.4};
\node[font=\tiny, text=black!70, anchor=north, yshift=-1.6pt, inner sep=0.5pt] at (axis cs:5,65.9) {65.9};
\node[font=\tiny, text=black!70, anchor=south, yshift=1.6pt, inner sep=0.5pt] at (axis cs:6,66.7) {66.7};
\nextgroupplot[title={2Wiki}, ymin=5, ymax=65, ytick={20,40,60}]
\addplot[name path=U, draw=none, forget plot] table[x=x, y expr=\thisrow{m}+\thisrow{sd}] {
x  m  sd
0  18.5  2.74
1  29.5  3.10
2  50.8  3.37
3  50.8  3.26
4  43.6  3.26
5  49.6  3.29
6  52.7  3.29
};
\addplot[name path=L, draw=none, forget plot] table[x=x, y expr=\thisrow{m}-\thisrow{sd}] {
x  m  sd
0  18.5  2.74
1  29.5  3.10
2  50.8  3.37
3  50.8  3.26
4  43.6  3.26
5  49.6  3.29
6  52.7  3.29
};
\addplot[lineO, fill opacity=0.15, draw=none, forget plot] fill between[of=U and L];
\addplot[lineO, line width=0.9pt, mark=*, mark size=1.4pt, mark options={fill=lineO, draw=white, line width=0.35pt}] table[x=x, y=m] {
x  m  sd
0  18.5  2.74
1  29.5  3.10
2  50.8  3.37
3  50.8  3.26
4  43.6  3.26
5  49.6  3.29
6  52.7  3.29
};
\node[font=\tiny, text=black!70, anchor=north, yshift=-1.6pt, inner sep=0.5pt] at (axis cs:0,18.5) {18.5};
\node[font=\tiny, text=black!70, anchor=north, yshift=-1.6pt, inner sep=0.5pt] at (axis cs:1,29.5) {29.5};
\node[font=\tiny, text=black!70, anchor=south, yshift=1.6pt, inner sep=0.5pt] at (axis cs:2,50.8) {50.8};
\node[font=\tiny, text=black!70, anchor=north, yshift=-1.6pt, inner sep=0.5pt] at (axis cs:3,50.8) {50.8};
\node[font=\tiny, text=black!70, anchor=north, yshift=-1.6pt, inner sep=0.5pt] at (axis cs:4,43.6) {43.6};
\node[font=\tiny, text=black!70, anchor=south, yshift=1.6pt, inner sep=0.5pt] at (axis cs:5,49.6) {49.6};
\node[font=\tiny, text=black!70, anchor=south, yshift=1.6pt, inner sep=0.5pt] at (axis cs:6,52.7) {52.7};
\nextgroupplot[title={NQ}, ymin=5, ymax=70, ytick={20,40,60}]
\addplot[name path=U, draw=none, forget plot] table[x=x, y expr=\thisrow{m}+\thisrow{sd}] {
x  m  sd
0  18.0  2.43
1  22.3  2.57
2  40.1  2.98
3  47.2  2.94
4  50.6  2.98
5  48.2  3.03
6  54.3  3.09
};
\addplot[name path=L, draw=none, forget plot] table[x=x, y expr=\thisrow{m}-\thisrow{sd}] {
x  m  sd
0  18.0  2.43
1  22.3  2.57
2  40.1  2.98
3  47.2  2.94
4  50.6  2.98
5  48.2  3.03
6  54.3  3.09
};
\addplot[lineO, fill opacity=0.15, draw=none, forget plot] fill between[of=U and L];
\addplot[lineO, line width=0.9pt, mark=*, mark size=1.4pt, mark options={fill=lineO, draw=white, line width=0.35pt}] table[x=x, y=m] {
x  m  sd
0  18.0  2.43
1  22.3  2.57
2  40.1  2.98
3  47.2  2.94
4  50.6  2.98
5  48.2  3.03
6  54.3  3.09
};
\node[font=\tiny, text=black!70, anchor=north, yshift=-1.6pt, inner sep=0.5pt] at (axis cs:0,18.0) {18.0};
\node[font=\tiny, text=black!70, anchor=south, yshift=1.6pt, inner sep=0.5pt] at (axis cs:1,22.3) {22.3};
\node[font=\tiny, text=black!70, anchor=south, yshift=1.6pt, inner sep=0.5pt] at (axis cs:2,40.1) {40.1};
\node[font=\tiny, text=black!70, anchor=south, yshift=1.6pt, inner sep=0.5pt] at (axis cs:3,47.2) {47.2};
\node[font=\tiny, text=black!70, anchor=south, yshift=1.6pt, inner sep=0.5pt] at (axis cs:4,50.6) {50.6};
\node[font=\tiny, text=black!70, anchor=north, yshift=-1.6pt, inner sep=0.5pt] at (axis cs:5,48.2) {48.2};
\node[font=\tiny, text=black!70, anchor=south, yshift=1.6pt, inner sep=0.5pt] at (axis cs:6,54.3) {54.3};
\nextgroupplot[title={TriviaQA}, ymin=10, ymax=60, ytick={10,20,30,40,50,60}, ylabel={F1}, ylabel style={font=\scriptsize, yshift=-3pt}]
\addplot[name path=U, draw=none, forget plot] table[x=x, y expr=\thisrow{m}+\thisrow{sd}] {
x  m  sd
0  21.7  1.84
1  29.6  2.18
2  40.4  2.40
3  41.5  2.31
4  45.2  2.49
5  47.9  2.38
6  48.2  2.81
};
\addplot[name path=L, draw=none, forget plot] table[x=x, y expr=\thisrow{m}-\thisrow{sd}] {
x  m  sd
0  21.7  1.84
1  29.6  2.18
2  40.4  2.40
3  41.5  2.31
4  45.2  2.49
5  47.9  2.38
6  48.2  2.81
};
\addplot[lineO, fill opacity=0.15, draw=none, forget plot] fill between[of=U and L];
\addplot[lineO, line width=0.9pt, mark=*, mark size=1.4pt, mark options={fill=lineO, draw=white, line width=0.35pt}] table[x=x, y=m] {
x  m  sd
0  21.7  1.84
1  29.6  2.18
2  40.4  2.40
3  41.5  2.31
4  45.2  2.49
5  47.9  2.38
6  48.2  2.81
};
\node[font=\tiny, text=black!70, anchor=north, yshift=-1.6pt, inner sep=0.5pt] at (axis cs:0,21.7) {21.7};
\node[font=\tiny, text=black!70, anchor=north, yshift=-1.6pt, inner sep=0.5pt] at (axis cs:1,29.6) {29.6};
\node[font=\tiny, text=black!70, anchor=south, yshift=1.6pt, inner sep=0.5pt] at (axis cs:2,40.4) {40.4};
\node[font=\tiny, text=black!70, anchor=north, yshift=-1.6pt, inner sep=0.5pt] at (axis cs:3,41.5) {41.5};
\node[font=\tiny, text=black!70, anchor=south, yshift=1.6pt, inner sep=0.5pt] at (axis cs:4,45.2) {45.2};
\node[font=\tiny, text=black!70, anchor=south, yshift=1.6pt, inner sep=0.5pt] at (axis cs:5,47.9) {47.9};
\node[font=\tiny, text=black!70, anchor=north, yshift=-1.6pt, inner sep=0.5pt] at (axis cs:6,48.2) {48.2};
\nextgroupplot[title={Qasper}, ymin=5, ymax=50, ytick={10,20,30,40,50}]
\addplot[name path=U, draw=none, forget plot] table[x=x, y expr=\thisrow{m}+\thisrow{sd}] {
x  m  sd
0  13.5  1.41
1  15.5  1.41
2  26.5  1.66
3  33.4  1.86
4  41.3  1.73
5  39.8  1.81
6  39.1  1.89
};
\addplot[name path=L, draw=none, forget plot] table[x=x, y expr=\thisrow{m}-\thisrow{sd}] {
x  m  sd
0  13.5  1.41
1  15.5  1.41
2  26.5  1.66
3  33.4  1.86
4  41.3  1.73
5  39.8  1.81
6  39.1  1.89
};
\addplot[lineO, fill opacity=0.15, draw=none, forget plot] fill between[of=U and L];
\addplot[lineO, line width=0.9pt, mark=*, mark size=1.4pt, mark options={fill=lineO, draw=white, line width=0.35pt}] table[x=x, y=m] {
x  m  sd
0  13.5  1.41
1  15.5  1.41
2  26.5  1.66
3  33.4  1.86
4  41.3  1.73
5  39.8  1.81
6  39.1  1.89
};
\node[font=\tiny, text=black!70, anchor=north, yshift=-1.6pt, inner sep=0.5pt] at (axis cs:0,13.5) {13.5};
\node[font=\tiny, text=black!70, anchor=south, yshift=1.6pt, inner sep=0.5pt] at (axis cs:1,15.5) {15.5};
\node[font=\tiny, text=black!70, anchor=south, yshift=1.6pt, inner sep=0.5pt] at (axis cs:2,26.5) {26.5};
\node[font=\tiny, text=black!70, anchor=south, yshift=1.6pt, inner sep=0.5pt] at (axis cs:3,33.4) {33.4};
\node[font=\tiny, text=black!70, anchor=south, yshift=1.6pt, inner sep=0.5pt] at (axis cs:4,41.3) {41.3};
\node[font=\tiny, text=black!70, anchor=north, yshift=-1.6pt, inner sep=0.5pt] at (axis cs:5,39.8) {39.8};
\node[font=\tiny, text=black!70, anchor=south, yshift=1.6pt, inner sep=0.5pt] at (axis cs:6,39.1) {39.1};
\nextgroupplot[title={MultiFieldQA}, ymin=0, ymax=40, ytick={0,10,20,30,40}]
\addplot[name path=U, draw=none, forget plot] table[x=x, y expr=\thisrow{m}+\thisrow{sd}] {
x  m  sd
0  9.1  2.02
1  12.2  2.19
2  22.9  2.81
3  28.3  3.25
4  27.5  3.33
5  27.6  3.33
6  26.3  3.18
};
\addplot[name path=L, draw=none, forget plot] table[x=x, y expr=\thisrow{m}-\thisrow{sd}] {
x  m  sd
0  9.1  2.02
1  12.2  2.19
2  22.9  2.81
3  28.3  3.25
4  27.5  3.33
5  27.6  3.33
6  26.3  3.18
};
\addplot[lineO, fill opacity=0.15, draw=none, forget plot] fill between[of=U and L];
\addplot[lineO, line width=0.9pt, mark=*, mark size=1.4pt, mark options={fill=lineO, draw=white, line width=0.35pt}] table[x=x, y=m] {
x  m  sd
0  9.1  2.02
1  12.2  2.19
2  22.9  2.81
3  28.3  3.25
4  27.5  3.33
5  27.6  3.33
6  26.3  3.18
};
\node[font=\tiny, text=black!70, anchor=north, yshift=-1.6pt, inner sep=0.5pt] at (axis cs:0,9.1) {9.1};
\node[font=\tiny, text=black!70, anchor=south, yshift=1.6pt, inner sep=0.5pt] at (axis cs:1,12.2) {12.2};
\node[font=\tiny, text=black!70, anchor=south, yshift=1.6pt, inner sep=0.5pt] at (axis cs:2,22.9) {22.9};
\node[font=\tiny, text=black!70, anchor=south, yshift=1.6pt, inner sep=0.5pt] at (axis cs:3,28.3) {28.3};
\node[font=\tiny, text=black!70, anchor=north, yshift=-1.6pt, inner sep=0.5pt] at (axis cs:4,27.5) {27.5};
\node[font=\tiny, text=black!70, anchor=south, yshift=1.6pt, inner sep=0.5pt] at (axis cs:5,27.6) {27.6};
\node[font=\tiny, text=black!70, anchor=north, yshift=-1.6pt, inner sep=0.5pt] at (axis cs:6,26.3) {26.3};
\nextgroupplot[title={NarrativeQA}, ymin=-5, ymax=25, ytick={0,10,20}]
\addplot[name path=U, draw=none, forget plot] table[x=x, y expr=\thisrow{m}+\thisrow{sd}] {
x  m  sd
0  3.4  0.56
1  4.2  0.54
2  10.1  1.14
3  13.2  1.07
4  18.7  1.12
5  17.1  1.06
6  17.1  1.23
};
\addplot[name path=L, draw=none, forget plot] table[x=x, y expr=\thisrow{m}-\thisrow{sd}] {
x  m  sd
0  3.4  0.56
1  4.2  0.54
2  10.1  1.14
3  13.2  1.07
4  18.7  1.12
5  17.1  1.06
6  17.1  1.23
};
\addplot[lineO, fill opacity=0.15, draw=none, forget plot] fill between[of=U and L];
\addplot[lineO, line width=0.9pt, mark=*, mark size=1.4pt, mark options={fill=lineO, draw=white, line width=0.35pt}] table[x=x, y=m] {
x  m  sd
0  3.4  0.56
1  4.2  0.54
2  10.1  1.14
3  13.2  1.07
4  18.7  1.12
5  17.1  1.06
6  17.1  1.23
};
\node[font=\tiny, text=black!70, anchor=north, yshift=-1.6pt, inner sep=0.5pt] at (axis cs:0,3.4) {3.4};
\node[font=\tiny, text=black!70, anchor=south, yshift=1.6pt, inner sep=0.5pt] at (axis cs:1,4.2) {4.2};
\node[font=\tiny, text=black!70, anchor=south, yshift=1.6pt, inner sep=0.5pt] at (axis cs:2,10.1) {10.1};
\node[font=\tiny, text=black!70, anchor=south, yshift=1.6pt, inner sep=0.5pt] at (axis cs:3,13.2) {13.2};
\node[font=\tiny, text=black!70, anchor=south, yshift=1.6pt, inner sep=0.5pt] at (axis cs:4,18.7) {18.7};
\node[font=\tiny, text=black!70, anchor=north, yshift=-1.6pt, inner sep=0.5pt] at (axis cs:5,17.1) {17.1};
\node[font=\tiny, text=black!70, anchor=south, yshift=1.6pt, inner sep=0.5pt] at (axis cs:6,17.1) {17.1};
\nextgroupplot[title={LoCoMo}, ymin=-5, ymax=35, ytick={0,10,20,30}]
\addplot[name path=U, draw=none, forget plot] table[x=x, y expr=\thisrow{m}+\thisrow{sd}] {
x  m  sd
0  3.6  0.38
1  5.5  0.46
2  11.7  0.64
3  16.5  0.69
4  26.9  0.90
5  26.5  0.89
6  21.4  0.85
};
\addplot[name path=L, draw=none, forget plot] table[x=x, y expr=\thisrow{m}-\thisrow{sd}] {
x  m  sd
0  3.6  0.38
1  5.5  0.46
2  11.7  0.64
3  16.5  0.69
4  26.9  0.90
5  26.5  0.89
6  21.4  0.85
};
\addplot[lineO, fill opacity=0.15, draw=none, forget plot] fill between[of=U and L];
\addplot[lineO, line width=0.9pt, mark=*, mark size=1.4pt, mark options={fill=lineO, draw=white, line width=0.35pt}] table[x=x, y=m] {
x  m  sd
0  3.6  0.38
1  5.5  0.46
2  11.7  0.64
3  16.5  0.69
4  26.9  0.90
5  26.5  0.89
6  21.4  0.85
};
\node[font=\tiny, text=black!70, anchor=north, yshift=-1.6pt, inner sep=0.5pt] at (axis cs:0,3.6) {3.6};
\node[font=\tiny, text=black!70, anchor=south, yshift=1.6pt, inner sep=0.5pt] at (axis cs:1,5.5) {5.5};
\node[font=\tiny, text=black!70, anchor=south, yshift=1.6pt, inner sep=0.5pt] at (axis cs:2,11.7) {11.7};
\node[font=\tiny, text=black!70, anchor=south, yshift=1.6pt, inner sep=0.5pt] at (axis cs:3,16.5) {16.5};
\node[font=\tiny, text=black!70, anchor=south, yshift=1.6pt, inner sep=0.5pt] at (axis cs:4,26.9) {26.9};
\node[font=\tiny, text=black!70, anchor=north, yshift=-1.6pt, inner sep=0.5pt] at (axis cs:5,26.5) {26.5};
\node[font=\tiny, text=black!70, anchor=north, yshift=-1.6pt, inner sep=0.5pt] at (axis cs:6,21.4) {21.4};
\end{groupplot}
\end{tikzpicture}
\caption{F1 of RoDeL-Seq with Qwen3.5-2B on each dataset after each update. HotpotQA and MuSiQue (ID) are in-domain and the others out-of-domain. Fr.: frozen roles (PRAG-style). Shaded bands show $\pm 1$ standard deviation of a test-set bootstrap.}
\label{fig:perset2b}
\end{figure}

\begin{figure}[!htb]
\centering
\begin{tikzpicture}
\begin{groupplot}[
  group style={group size=5 by 2, horizontal sep=0.55cm, vertical sep=0.95cm},
  scale only axis, width=2.5cm, height=2.0cm,
  xmin=-0.5, xmax=3.5,
  xtick={0,...,3},
  xticklabels={Fr.,$u_1$,$u_2$,$u_3$},
  axis lines=left, axis line style={-, black!60}, tick style={black!60}, major tick length=2pt,
  tick label style={font=\tiny, inner sep=1pt},
  title style={font=\footnotesize, yshift=-4pt},
  ymajorgrids, grid style={black!22, dashed},
  clip=false,
]
\nextgroupplot[title={HotpotQA (ID)}, ymin=0, ymax=55, ytick={0,20,40}, ylabel={F1}, ylabel style={font=\scriptsize, yshift=-3pt}]
\addplot[name path=U, draw=none, forget plot] table[x=x, y expr=\thisrow{m}+\thisrow{sd}] {
x  m  sd
0  13.2  2.28
1  37.5  3.23
2  40.0  3.16
3  43.8  3.23
};
\addplot[name path=L, draw=none, forget plot] table[x=x, y expr=\thisrow{m}-\thisrow{sd}] {
x  m  sd
0  13.2  2.28
1  37.5  3.23
2  40.0  3.16
3  43.8  3.23
};
\addplot[lineO, fill opacity=0.15, draw=none, forget plot] fill between[of=U and L];
\addplot[lineO, line width=0.9pt, mark=*, mark size=1.4pt, mark options={fill=lineO, draw=white, line width=0.35pt}] table[x=x, y=m] {
x  m  sd
0  13.2  2.28
1  37.5  3.23
2  40.0  3.16
3  43.8  3.23
};
\node[font=\tiny, text=black!70, anchor=north, yshift=-1.6pt, inner sep=0.5pt] at (axis cs:0,13.2) {13.2};
\node[font=\tiny, text=black!70, anchor=south, yshift=1.6pt, inner sep=0.5pt] at (axis cs:1,37.5) {37.5};
\node[font=\tiny, text=black!70, anchor=south east, yshift=1.6pt, inner sep=0.5pt] at (axis cs:2,40.0) {40.0};
\node[font=\tiny, text=black!70, anchor=south, yshift=1.6pt, inner sep=0.5pt] at (axis cs:3,43.8) {43.8};
\nextgroupplot[title={MuSiQue (ID)}, ymin=-5, ymax=30, ytick={0,10,20,30}]
\addplot[name path=U, draw=none, forget plot] table[x=x, y expr=\thisrow{m}+\thisrow{sd}] {
x  m  sd
0  3.1  1.05
1  15.1  2.16
2  15.6  2.19
3  16.2  2.14
};
\addplot[name path=L, draw=none, forget plot] table[x=x, y expr=\thisrow{m}-\thisrow{sd}] {
x  m  sd
0  3.1  1.05
1  15.1  2.16
2  15.6  2.19
3  16.2  2.14
};
\addplot[lineO, fill opacity=0.15, draw=none, forget plot] fill between[of=U and L];
\addplot[lineO, line width=0.9pt, mark=*, mark size=1.4pt, mark options={fill=lineO, draw=white, line width=0.35pt}] table[x=x, y=m] {
x  m  sd
0  3.1  1.05
1  15.1  2.16
2  15.6  2.19
3  16.2  2.14
};
\node[font=\tiny, text=black!70, anchor=north, yshift=-1.6pt, inner sep=0.5pt] at (axis cs:0,3.1) {3.1};
\node[font=\tiny, text=black!70, anchor=south, yshift=1.6pt, inner sep=0.5pt] at (axis cs:1,15.1) {15.1};
\node[font=\tiny, text=black!70, anchor=south, yshift=1.6pt, inner sep=0.5pt] at (axis cs:2,15.6) {15.6};
\node[font=\tiny, text=black!70, anchor=south, yshift=1.6pt, inner sep=0.5pt] at (axis cs:3,16.2) {16.2};
\nextgroupplot[title={SQuAD}, ymin=25, ymax=80, ytick={40,60,80}]
\addplot[name path=U, draw=none, forget plot] table[x=x, y expr=\thisrow{m}+\thisrow{sd}] {
x  m  sd
0  35.6  2.07
1  64.2  1.53
2  66.5  1.52
3  67.9  1.48
};
\addplot[name path=L, draw=none, forget plot] table[x=x, y expr=\thisrow{m}-\thisrow{sd}] {
x  m  sd
0  35.6  2.07
1  64.2  1.53
2  66.5  1.52
3  67.9  1.48
};
\addplot[lineO, fill opacity=0.15, draw=none, forget plot] fill between[of=U and L];
\addplot[lineO, line width=0.9pt, mark=*, mark size=1.4pt, mark options={fill=lineO, draw=white, line width=0.35pt}] table[x=x, y=m] {
x  m  sd
0  35.6  2.07
1  64.2  1.53
2  66.5  1.52
3  67.9  1.48
};
\node[font=\tiny, text=black!70, anchor=north, yshift=-1.6pt, inner sep=0.5pt] at (axis cs:0,35.6) {35.6};
\node[font=\tiny, text=black!70, anchor=south, yshift=1.6pt, inner sep=0.5pt] at (axis cs:1,64.2) {64.2};
\node[font=\tiny, text=black!70, anchor=south, yshift=1.6pt, inner sep=0.5pt] at (axis cs:2,66.5) {66.5};
\node[font=\tiny, text=black!70, anchor=south, yshift=1.6pt, inner sep=0.5pt] at (axis cs:3,67.9) {67.9};
\nextgroupplot[title={2Wiki}, ymin=5, ymax=65, ytick={20,40,60}]
\addplot[name path=U, draw=none, forget plot] table[x=x, y expr=\thisrow{m}+\thisrow{sd}] {
x  m  sd
0  18.5  2.74
1  53.9  3.36
2  53.0  3.26
3  53.6  3.21
};
\addplot[name path=L, draw=none, forget plot] table[x=x, y expr=\thisrow{m}-\thisrow{sd}] {
x  m  sd
0  18.5  2.74
1  53.9  3.36
2  53.0  3.26
3  53.6  3.21
};
\addplot[lineO, fill opacity=0.15, draw=none, forget plot] fill between[of=U and L];
\addplot[lineO, line width=0.9pt, mark=*, mark size=1.4pt, mark options={fill=lineO, draw=white, line width=0.35pt}] table[x=x, y=m] {
x  m  sd
0  18.5  2.74
1  53.9  3.36
2  53.0  3.26
3  53.6  3.21
};
\node[font=\tiny, text=black!70, anchor=north, yshift=-1.6pt, inner sep=0.5pt] at (axis cs:0,18.5) {18.5};
\node[font=\tiny, text=black!70, anchor=south, yshift=1.6pt, inner sep=0.5pt] at (axis cs:1,53.9) {53.9};
\node[font=\tiny, text=black!70, anchor=north, yshift=-1.6pt, inner sep=0.5pt] at (axis cs:2,53.0) {53.0};
\node[font=\tiny, text=black!70, anchor=south, yshift=1.6pt, inner sep=0.5pt] at (axis cs:3,53.6) {53.6};
\nextgroupplot[title={NQ}, ymin=5, ymax=70, ytick={20,40,60}]
\addplot[name path=U, draw=none, forget plot] table[x=x, y expr=\thisrow{m}+\thisrow{sd}] {
x  m  sd
0  18.0  2.43
1  44.3  3.07
2  46.5  3.03
3  50.6  2.88
};
\addplot[name path=L, draw=none, forget plot] table[x=x, y expr=\thisrow{m}-\thisrow{sd}] {
x  m  sd
0  18.0  2.43
1  44.3  3.07
2  46.5  3.03
3  50.6  2.88
};
\addplot[lineO, fill opacity=0.15, draw=none, forget plot] fill between[of=U and L];
\addplot[lineO, line width=0.9pt, mark=*, mark size=1.4pt, mark options={fill=lineO, draw=white, line width=0.35pt}] table[x=x, y=m] {
x  m  sd
0  18.0  2.43
1  44.3  3.07
2  46.5  3.03
3  50.6  2.88
};
\node[font=\tiny, text=black!70, anchor=north, yshift=-1.6pt, inner sep=0.5pt] at (axis cs:0,18.0) {18.0};
\node[font=\tiny, text=black!70, anchor=south, yshift=1.6pt, inner sep=0.5pt] at (axis cs:1,44.3) {44.3};
\node[font=\tiny, text=black!70, anchor=south east, yshift=1.6pt, inner sep=0.5pt] at (axis cs:2,46.5) {46.5};
\node[font=\tiny, text=black!70, anchor=south, yshift=1.6pt, inner sep=0.5pt] at (axis cs:3,50.6) {50.6};
\nextgroupplot[title={TriviaQA}, ymin=10, ymax=60, ytick={10,20,30,40,50,60}, ylabel={F1}, ylabel style={font=\scriptsize, yshift=-3pt}]
\addplot[name path=U, draw=none, forget plot] table[x=x, y expr=\thisrow{m}+\thisrow{sd}] {
x  m  sd
0  21.7  1.84
1  39.5  2.43
2  46.0  2.54
3  46.4  2.22
};
\addplot[name path=L, draw=none, forget plot] table[x=x, y expr=\thisrow{m}-\thisrow{sd}] {
x  m  sd
0  21.7  1.84
1  39.5  2.43
2  46.0  2.54
3  46.4  2.22
};
\addplot[lineO, fill opacity=0.15, draw=none, forget plot] fill between[of=U and L];
\addplot[lineO, line width=0.9pt, mark=*, mark size=1.4pt, mark options={fill=lineO, draw=white, line width=0.35pt}] table[x=x, y=m] {
x  m  sd
0  21.7  1.84
1  39.5  2.43
2  46.0  2.54
3  46.4  2.22
};
\node[font=\tiny, text=black!70, anchor=north, yshift=-1.6pt, inner sep=0.5pt] at (axis cs:0,21.7) {21.7};
\node[font=\tiny, text=black!70, anchor=south east, yshift=1.6pt, inner sep=0.5pt] at (axis cs:1,39.5) {39.5};
\node[font=\tiny, text=black!70, anchor=south, yshift=1.6pt, inner sep=0.5pt] at (axis cs:2,46.0) {46.0};
\node[font=\tiny, text=black!70, anchor=south, yshift=1.6pt, inner sep=0.5pt] at (axis cs:3,46.4) {46.4};
\nextgroupplot[title={Qasper}, ymin=5, ymax=50, ytick={10,20,30,40,50}]
\addplot[name path=U, draw=none, forget plot] table[x=x, y expr=\thisrow{m}+\thisrow{sd}] {
x  m  sd
0  13.5  1.41
1  28.7  1.68
2  37.9  1.84
3  36.1  1.81
};
\addplot[name path=L, draw=none, forget plot] table[x=x, y expr=\thisrow{m}-\thisrow{sd}] {
x  m  sd
0  13.5  1.41
1  28.7  1.68
2  37.9  1.84
3  36.1  1.81
};
\addplot[lineO, fill opacity=0.15, draw=none, forget plot] fill between[of=U and L];
\addplot[lineO, line width=0.9pt, mark=*, mark size=1.4pt, mark options={fill=lineO, draw=white, line width=0.35pt}] table[x=x, y=m] {
x  m  sd
0  13.5  1.41
1  28.7  1.68
2  37.9  1.84
3  36.1  1.81
};
\node[font=\tiny, text=black!70, anchor=north, yshift=-1.6pt, inner sep=0.5pt] at (axis cs:0,13.5) {13.5};
\node[font=\tiny, text=black!70, anchor=south east, yshift=1.6pt, inner sep=0.5pt] at (axis cs:1,28.7) {28.7};
\node[font=\tiny, text=black!70, anchor=south, yshift=1.6pt, inner sep=0.5pt] at (axis cs:2,37.9) {37.9};
\node[font=\tiny, text=black!70, anchor=north, yshift=-1.6pt, inner sep=0.5pt] at (axis cs:3,36.1) {36.1};
\nextgroupplot[title={MultiFieldQA}, ymin=0, ymax=40, ytick={0,10,20,30,40}]
\addplot[name path=U, draw=none, forget plot] table[x=x, y expr=\thisrow{m}+\thisrow{sd}] {
x  m  sd
0  9.1  2.02
1  23.2  3.03
2  25.9  3.37
3  26.5  3.26
};
\addplot[name path=L, draw=none, forget plot] table[x=x, y expr=\thisrow{m}-\thisrow{sd}] {
x  m  sd
0  9.1  2.02
1  23.2  3.03
2  25.9  3.37
3  26.5  3.26
};
\addplot[lineO, fill opacity=0.15, draw=none, forget plot] fill between[of=U and L];
\addplot[lineO, line width=0.9pt, mark=*, mark size=1.4pt, mark options={fill=lineO, draw=white, line width=0.35pt}] table[x=x, y=m] {
x  m  sd
0  9.1  2.02
1  23.2  3.03
2  25.9  3.37
3  26.5  3.26
};
\node[font=\tiny, text=black!70, anchor=north, yshift=-1.6pt, inner sep=0.5pt] at (axis cs:0,9.1) {9.1};
\node[font=\tiny, text=black!70, anchor=south east, yshift=1.6pt, inner sep=0.5pt] at (axis cs:1,23.2) {23.2};
\node[font=\tiny, text=black!70, anchor=south, yshift=1.6pt, inner sep=0.5pt] at (axis cs:2,25.9) {25.9};
\node[font=\tiny, text=black!70, anchor=south, yshift=1.6pt, inner sep=0.5pt] at (axis cs:3,26.5) {26.5};
\nextgroupplot[title={NarrativeQA}, ymin=-5, ymax=25, ytick={0,10,20}]
\addplot[name path=U, draw=none, forget plot] table[x=x, y expr=\thisrow{m}+\thisrow{sd}] {
x  m  sd
0  3.4  0.56
1  10.3  1.08
2  15.8  1.35
3  17.3  1.08
};
\addplot[name path=L, draw=none, forget plot] table[x=x, y expr=\thisrow{m}-\thisrow{sd}] {
x  m  sd
0  3.4  0.56
1  10.3  1.08
2  15.8  1.35
3  17.3  1.08
};
\addplot[lineO, fill opacity=0.15, draw=none, forget plot] fill between[of=U and L];
\addplot[lineO, line width=0.9pt, mark=*, mark size=1.4pt, mark options={fill=lineO, draw=white, line width=0.35pt}] table[x=x, y=m] {
x  m  sd
0  3.4  0.56
1  10.3  1.08
2  15.8  1.35
3  17.3  1.08
};
\node[font=\tiny, text=black!70, anchor=north, yshift=-1.6pt, inner sep=0.5pt] at (axis cs:0,3.4) {3.4};
\node[font=\tiny, text=black!70, anchor=south east, yshift=1.6pt, inner sep=0.5pt] at (axis cs:1,10.3) {10.3};
\node[font=\tiny, text=black!70, anchor=south east, yshift=1.6pt, inner sep=0.5pt] at (axis cs:2,15.8) {15.8};
\node[font=\tiny, text=black!70, anchor=south, yshift=1.6pt, inner sep=0.5pt] at (axis cs:3,17.3) {17.3};
\nextgroupplot[title={LoCoMo}, ymin=-5, ymax=35, ytick={0,10,20,30}]
\addplot[name path=U, draw=none, forget plot] table[x=x, y expr=\thisrow{m}+\thisrow{sd}] {
x  m  sd
0  3.6  0.38
1  14.6  0.71
2  20.2  0.81
3  23.0  0.83
};
\addplot[name path=L, draw=none, forget plot] table[x=x, y expr=\thisrow{m}-\thisrow{sd}] {
x  m  sd
0  3.6  0.38
1  14.6  0.71
2  20.2  0.81
3  23.0  0.83
};
\addplot[lineO, fill opacity=0.15, draw=none, forget plot] fill between[of=U and L];
\addplot[lineO, line width=0.9pt, mark=*, mark size=1.4pt, mark options={fill=lineO, draw=white, line width=0.35pt}] table[x=x, y=m] {
x  m  sd
0  3.6  0.38
1  14.6  0.71
2  20.2  0.81
3  23.0  0.83
};
\node[font=\tiny, text=black!70, anchor=north, yshift=-1.6pt, inner sep=0.5pt] at (axis cs:0,3.6) {3.6};
\node[font=\tiny, text=black!70, anchor=south east, yshift=1.6pt, inner sep=0.5pt] at (axis cs:1,14.6) {14.6};
\node[font=\tiny, text=black!70, anchor=south east, yshift=1.6pt, inner sep=0.5pt] at (axis cs:2,20.2) {20.2};
\node[font=\tiny, text=black!70, anchor=south, yshift=1.6pt, inner sep=0.5pt] at (axis cs:3,23.0) {23.0};
\end{groupplot}
\end{tikzpicture}
\caption{F1 of RoDeL-Sim with Qwen3.5-2B on each dataset after each simultaneous update. Datasets, Fr., and shaded bands are as in \Cref{fig:perset2b}.}
\label{fig:perset2bsim}
\end{figure}
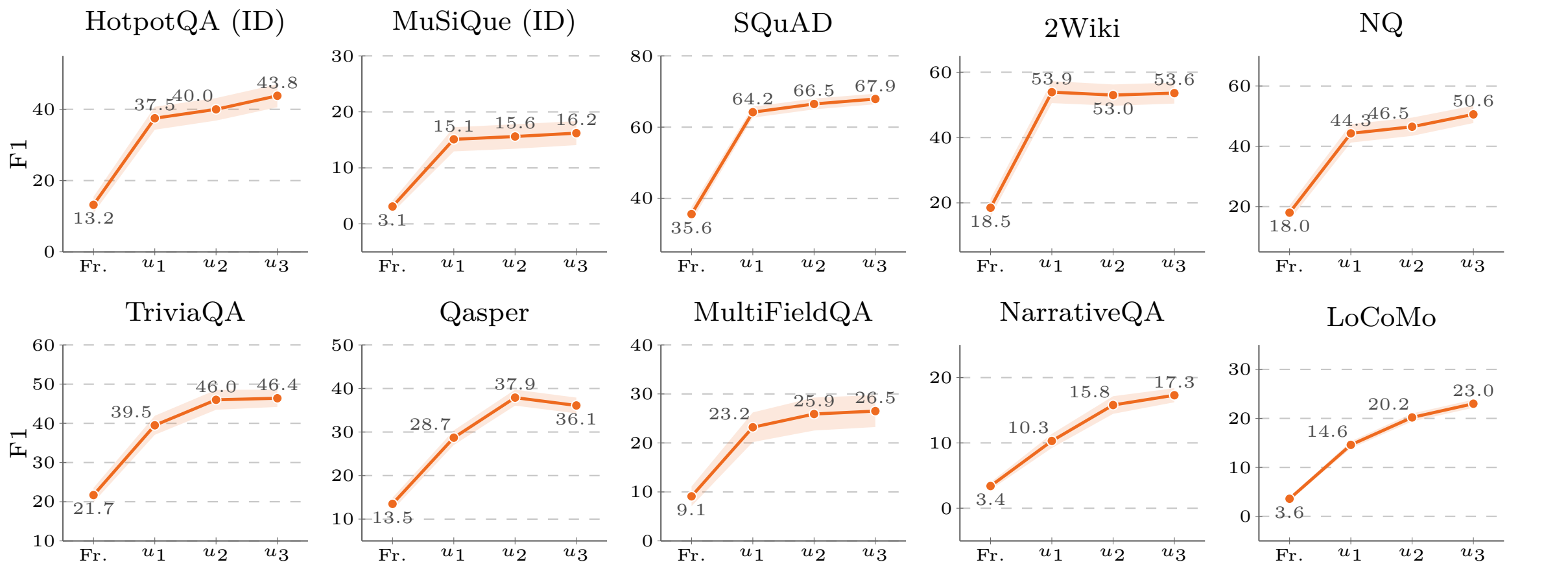

\paragraph{Qwen3.5-2B.} \Cref{fig:avg2b,fig:perset2b} show the F1 of RoDeL-Seq after each update, on average and on each dataset, and \Cref{fig:avg2bsim,fig:perset2bsim} show the same for RoDeL-Sim. At the end of training, both variants score significantly higher than the frozen roles on both averages. The in-domain average rises from 8.2 to 31.5 with RoDeL-Seq and to 30.0 with RoDeL-Sim, and the out-of-domain average rises from 15.4 to 40.7 and to 40.2.

\enlargethispage{6pt} %

\paragraph{Qwen3.5-0.8B.} \Cref{fig:avg08,fig:perset08} show the F1 of RoDeL-Seq after each update, on average and on each dataset, and \Cref{fig:avg08sim,fig:perset08sim} show the same for RoDeL-Sim. At the end of training, both variants score significantly higher than the frozen roles on both averages. The in-domain average rises from 3.7 to 13.6 with RoDeL-Seq and to 13.8 with RoDeL-Sim, and the out-of-domain average rises from 4.7 to 23.6 and to 23.7. With this model, the last update lowers both averages of each variant. The averages of RoDeL-Seq are highest after $s_3$ (15.6 in domain and 24.9 out of domain), and those of RoDeL-Sim are highest after $u_2$ (17.4 and 30.1). The decline is significant for the out-of-domain average of RoDeL-Seq and for both averages of RoDeL-Sim.

\begin{figure}[!htb]
\begin{minipage}[t]{\colw}
\centering
\pgfplotstableread{
x  id    idsd  ood   oodsd
0  3.7  0.84  4.7  0.41
1  4.1  0.88  5.7  0.42
2  1.8  0.65  6.6  0.48
3  4.0  0.72  9.3  0.47
4  12.9  1.35  23.0  0.71
5  15.6  1.56  24.9  0.72
6  13.6  1.46  23.6  0.69
}\dyntab
\begin{tikzpicture}
\begin{axis}[
  width=\colw, height=5.2cm,
  xmin=-0.5, xmax=6.5, ymin=-4, ymax=37,
  xtick={0,...,6},
  xticklabels={Frozen,\textcolor{stuV}{$s_1$},\textcolor{tchV}{$t_1$},\textcolor{stuV}{$s_2$},\textcolor{tchV}{$t_2$},\textcolor{stuV}{$s_3$},\textcolor{tchV}{$t_3$}},
  ytick={0,10,20,30},
  ylabel={Average F1},
  axis lines=left, axis line style={-, black!60}, tick style={black!60},
  tick label style={font=\small}, label style={font=\small},
  ymajorgrids, grid style={black!22, dashed},
  set layers, clip=false,
  legend style={font=\small, at={(0.5,1.03)}, anchor=south, draw=none, fill=none, legend columns=2, /tikz/every even column/.append style={column sep=10pt}},
  legend cell align=left,
]
\begin{pgfonlayer}{axis background}
\draw[black!25] (axis cs:0.5,-4) -- (axis cs:0.5,37);
\draw[black!25] (axis cs:2.5,-4) -- (axis cs:2.5,37);
\draw[black!25] (axis cs:4.5,-4) -- (axis cs:4.5,37);
\end{pgfonlayer}
\node[font=\scriptsize, text=black!55, anchor=north] at (axis cs:1.5,37) {Round 1};
\node[font=\scriptsize, text=black!55, anchor=north] at (axis cs:3.5,37) {Round 2};
\node[font=\scriptsize, text=black!55, anchor=north] at (axis cs:5.5,37) {Round 3};
\addplot[name path=idU, draw=none, forget plot] table[x=x, y expr=\thisrow{id}+\thisrow{idsd}] {\dyntab};
\addplot[name path=idL, draw=none, forget plot] table[x=x, y expr=\thisrow{id}-\thisrow{idsd}] {\dyntab};
\addplot[lineO, fill opacity=0.15, draw=none, forget plot] fill between[of=idU and idL];
\addplot[name path=oodU, draw=none, forget plot] table[x=x, y expr=\thisrow{ood}+\thisrow{oodsd}] {\dyntab};
\addplot[name path=oodL, draw=none, forget plot] table[x=x, y expr=\thisrow{ood}-\thisrow{oodsd}] {\dyntab};
\addplot[lineO, fill opacity=0.15, draw=none, forget plot] fill between[of=oodU and oodL];
\addplot[lineO, line width=1.2pt, mark=*, mark size=2.3pt, mark options={fill=lineO, draw=white, line width=0.6pt},
  nodes near coords, point meta=explicit symbolic, every node near coord/.append style={font=\scriptsize, text=black!70, anchor=north, yshift=-4pt}]
  table[x=x, y=id, meta=id] {\dyntab};
\addlegendentry{In-domain}
\addplot[lineO, line width=1.2pt, mark=square*, mark size=2.1pt, mark options={fill=lineO, draw=white, line width=0.6pt},
  nodes near coords, point meta=explicit symbolic, every node near coord/.append style={font=\scriptsize, text=black!70, anchor=south, yshift=2pt}]
  table[x=x, y=ood, meta=ood] {\dyntab};
\addlegendentry{Out-of-domain}
\end{axis}
\end{tikzpicture}
\caption{Average F1 of RoDeL-Seq with Qwen3.5-0.8B after each update. Shaded bands show $\pm 1$ standard deviation of a test-set bootstrap.}
\label{fig:avg08}
\end{minipage}\hfill
\begin{minipage}[t]{\colw}
\centering
\pgfplotstableread{
x  id    idsd  ood   oodsd
0  3.7  0.84  4.7  0.41
1  2.2  0.27  8.2  0.28
2  17.4  1.58  30.1  0.76
3  13.8  1.35  23.7  0.63
}\simtab
\begin{tikzpicture}
\begin{axis}[
  width=\colw, height=5.2cm,
  xmin=-0.5, xmax=3.5, ymin=-4, ymax=37,
  xtick={0,...,3},
  xticklabels={Frozen,$u_1$,$u_2$,$u_3$},
  ytick={0,10,20,30},
  ylabel={Average F1},
  axis lines=left, axis line style={-, black!60}, tick style={black!60},
  tick label style={font=\small}, label style={font=\small},
  ymajorgrids, grid style={black!22, dashed},
  clip=false,
  legend style={font=\small, at={(0.5,1.03)}, anchor=south, draw=none, fill=none, legend columns=2, /tikz/every even column/.append style={column sep=10pt}},
  legend cell align=left,
]
\addplot[name path=idU, draw=none, forget plot] table[x=x, y expr=\thisrow{id}+\thisrow{idsd}] {\simtab};
\addplot[name path=idL, draw=none, forget plot] table[x=x, y expr=\thisrow{id}-\thisrow{idsd}] {\simtab};
\addplot[lineO, fill opacity=0.15, draw=none, forget plot] fill between[of=idU and idL];
\addplot[name path=oodU, draw=none, forget plot] table[x=x, y expr=\thisrow{ood}+\thisrow{oodsd}] {\simtab};
\addplot[name path=oodL, draw=none, forget plot] table[x=x, y expr=\thisrow{ood}-\thisrow{oodsd}] {\simtab};
\addplot[lineO, fill opacity=0.15, draw=none, forget plot] fill between[of=oodU and oodL];
\addplot[lineO, line width=1.2pt, mark=*, mark size=2.3pt, mark options={fill=lineO, draw=white, line width=0.6pt},
  nodes near coords, point meta=explicit symbolic, every node near coord/.append style={font=\scriptsize, text=black!70, anchor=north, yshift=-4pt}]
  table[x=x, y=id, meta=id] {\simtab};
\addlegendentry{In-domain}
\addplot[lineO, line width=1.2pt, mark=square*, mark size=2.1pt, mark options={fill=lineO, draw=white, line width=0.6pt},
  nodes near coords, point meta=explicit symbolic, every node near coord/.append style={font=\scriptsize, text=black!70, anchor=south, yshift=2pt}]
  table[x=x, y=ood, meta=ood] {\simtab};
\addlegendentry{Out-of-domain}
\end{axis}
\end{tikzpicture}
\caption{Average F1 of RoDeL-Sim with Qwen3.5-0.8B after each simultaneous update. Shaded bands are as in \Cref{fig:avg08}.}
\label{fig:avg08sim}
\end{minipage}
\end{figure}

\begin{figure}[!htb]
\centering
\begin{tikzpicture}
\begin{groupplot}[
  group style={group size=5 by 2, horizontal sep=0.55cm, vertical sep=0.95cm},
  scale only axis, width=2.5cm, height=2.0cm,
  xmin=-0.5, xmax=6.5,
  xtick={0,...,6},
  xticklabels={Fr.,\textcolor{stuV}{$s_1$},\textcolor{tchV}{$t_1$},\textcolor{stuV}{$s_2$},\textcolor{tchV}{$t_2$},\textcolor{stuV}{$s_3$},\textcolor{tchV}{$t_3$}},
  axis lines=left, axis line style={-, black!60}, tick style={black!60}, major tick length=2pt,
  tick label style={font=\tiny, inner sep=1pt},
  title style={font=\footnotesize, yshift=-4pt},
  ymajorgrids, grid style={black!22, dashed},
  extra x ticks={0.5,2.5,4.5}, extra x tick labels={},
  extra x tick style={grid=major, major grid style={black!25, solid}, tick style={draw=none}},
  clip=false,
]
\nextgroupplot[title={HotpotQA (ID)}, ymin=-5, ymax=35, ytick={0,10,20,30}, ylabel={F1}, ylabel style={font=\scriptsize, yshift=-3pt}]
\addplot[name path=U, draw=none, forget plot] table[x=x, y expr=\thisrow{m}+\thisrow{sd}] {
x  m  sd
0  6.1  1.59
1  7.0  1.70
2  3.6  1.30
3  6.1  1.30
4  17.3  2.23
5  24.1  2.75
6  19.2  2.52
};
\addplot[name path=L, draw=none, forget plot] table[x=x, y expr=\thisrow{m}-\thisrow{sd}] {
x  m  sd
0  6.1  1.59
1  7.0  1.70
2  3.6  1.30
3  6.1  1.30
4  17.3  2.23
5  24.1  2.75
6  19.2  2.52
};
\addplot[lineO, fill opacity=0.15, draw=none, forget plot] fill between[of=U and L];
\addplot[lineO, line width=0.9pt, mark=*, mark size=1.4pt, mark options={fill=lineO, draw=white, line width=0.35pt}] table[x=x, y=m] {
x  m  sd
0  6.1  1.59
1  7.0  1.70
2  3.6  1.30
3  6.1  1.30
4  17.3  2.23
5  24.1  2.75
6  19.2  2.52
};
\node[font=\tiny, text=black!70, anchor=north, yshift=-1.6pt, inner sep=0.5pt] at (axis cs:0,6.1) {6.1};
\node[font=\tiny, text=black!70, anchor=south, yshift=1.6pt, inner sep=0.5pt] at (axis cs:1,7.0) {7.0};
\node[font=\tiny, text=black!70, anchor=north, yshift=-1.6pt, inner sep=0.5pt] at (axis cs:2,3.6) {3.6};
\node[font=\tiny, text=black!70, anchor=south, yshift=1.6pt, inner sep=0.5pt] at (axis cs:3,6.1) {6.1};
\node[font=\tiny, text=black!70, anchor=south, yshift=1.6pt, inner sep=0.5pt] at (axis cs:4,17.3) {17.3};
\node[font=\tiny, text=black!70, anchor=south, yshift=1.6pt, inner sep=0.5pt] at (axis cs:5,24.1) {24.1};
\node[font=\tiny, text=black!70, anchor=north, yshift=-1.6pt, inner sep=0.5pt] at (axis cs:6,19.2) {19.2};
\nextgroupplot[title={MuSiQue (ID)}, ymin=-5, ymax=15, ytick={0,5,10,15}]
\addplot[name path=U, draw=none, forget plot] table[x=x, y expr=\thisrow{m}+\thisrow{sd}] {
x  m  sd
0  1.3  0.55
1  1.2  0.46
2  0.0  0.00
3  1.8  0.61
4  8.5  1.52
5  7.1  1.45
6  8.0  1.41
};
\addplot[name path=L, draw=none, forget plot] table[x=x, y expr=\thisrow{m}-\thisrow{sd}] {
x  m  sd
0  1.3  0.55
1  1.2  0.46
2  0.0  0.00
3  1.8  0.61
4  8.5  1.52
5  7.1  1.45
6  8.0  1.41
};
\addplot[lineO, fill opacity=0.15, draw=none, forget plot] fill between[of=U and L];
\addplot[lineO, line width=0.9pt, mark=*, mark size=1.4pt, mark options={fill=lineO, draw=white, line width=0.35pt}] table[x=x, y=m] {
x  m  sd
0  1.3  0.55
1  1.2  0.46
2  0.0  0.00
3  1.8  0.61
4  8.5  1.52
5  7.1  1.45
6  8.0  1.41
};
\node[font=\tiny, text=black!70, anchor=south, yshift=1.6pt, inner sep=0.5pt] at (axis cs:0,1.3) {1.3};
\node[font=\tiny, text=black!70, anchor=north, yshift=-1.6pt, inner sep=0.5pt] at (axis cs:1,1.2) {1.2};
\node[font=\tiny, text=black!70, anchor=north, yshift=-1.6pt, inner sep=0.5pt] at (axis cs:2,0.0) {0.0};
\node[font=\tiny, text=black!70, anchor=north, yshift=-1.6pt, inner sep=0.5pt] at (axis cs:3,1.8) {1.8};
\node[font=\tiny, text=black!70, anchor=south, yshift=1.6pt, inner sep=0.5pt] at (axis cs:4,8.5) {8.5};
\node[font=\tiny, text=black!70, anchor=north, yshift=-1.6pt, inner sep=0.5pt] at (axis cs:5,7.1) {7.1};
\node[font=\tiny, text=black!70, anchor=south, yshift=1.6pt, inner sep=0.5pt] at (axis cs:6,8.0) {8.0};
\nextgroupplot[title={SQuAD}, ymin=-5, ymax=45, ytick={0,10,20,30,40}]
\addplot[name path=U, draw=none, forget plot] table[x=x, y expr=\thisrow{m}+\thisrow{sd}] {
x  m  sd
0  8.7  1.25
1  4.9  0.68
2  3.6  0.62
3  7.5  0.57
4  30.8  1.43
5  35.8  1.62
6  30.5  1.48
};
\addplot[name path=L, draw=none, forget plot] table[x=x, y expr=\thisrow{m}-\thisrow{sd}] {
x  m  sd
0  8.7  1.25
1  4.9  0.68
2  3.6  0.62
3  7.5  0.57
4  30.8  1.43
5  35.8  1.62
6  30.5  1.48
};
\addplot[lineO, fill opacity=0.15, draw=none, forget plot] fill between[of=U and L];
\addplot[lineO, line width=0.9pt, mark=*, mark size=1.4pt, mark options={fill=lineO, draw=white, line width=0.35pt}] table[x=x, y=m] {
x  m  sd
0  8.7  1.25
1  4.9  0.68
2  3.6  0.62
3  7.5  0.57
4  30.8  1.43
5  35.8  1.62
6  30.5  1.48
};
\node[font=\tiny, text=black!70, anchor=south, yshift=1.6pt, inner sep=0.5pt] at (axis cs:0,8.7) {8.7};
\node[font=\tiny, text=black!70, anchor=north, yshift=-1.6pt, inner sep=0.5pt] at (axis cs:1,4.9) {4.9};
\node[font=\tiny, text=black!70, anchor=south, yshift=1.6pt, inner sep=0.5pt] at (axis cs:2,3.6) {3.6};
\node[font=\tiny, text=black!70, anchor=north, yshift=-1.6pt, inner sep=0.5pt] at (axis cs:3,7.5) {7.5};
\node[font=\tiny, text=black!70, anchor=south, yshift=1.6pt, inner sep=0.5pt] at (axis cs:4,30.8) {30.8};
\node[font=\tiny, text=black!70, anchor=south, yshift=1.6pt, inner sep=0.5pt] at (axis cs:5,35.8) {35.8};
\node[font=\tiny, text=black!70, anchor=north, yshift=-1.6pt, inner sep=0.5pt] at (axis cs:6,30.5) {30.5};
\nextgroupplot[title={2Wiki}, ymin=-5, ymax=40, ytick={0,10,20,30,40}]
\addplot[name path=U, draw=none, forget plot] table[x=x, y expr=\thisrow{m}+\thisrow{sd}] {
x  m  sd
0  12.1  2.28
1  9.8  2.03
2  6.1  1.70
3  9.5  1.59
4  20.6  2.52
5  26.7  2.86
6  25.2  2.87
};
\addplot[name path=L, draw=none, forget plot] table[x=x, y expr=\thisrow{m}-\thisrow{sd}] {
x  m  sd
0  12.1  2.28
1  9.8  2.03
2  6.1  1.70
3  9.5  1.59
4  20.6  2.52
5  26.7  2.86
6  25.2  2.87
};
\addplot[lineO, fill opacity=0.15, draw=none, forget plot] fill between[of=U and L];
\addplot[lineO, line width=0.9pt, mark=*, mark size=1.4pt, mark options={fill=lineO, draw=white, line width=0.35pt}] table[x=x, y=m] {
x  m  sd
0  12.1  2.28
1  9.8  2.03
2  6.1  1.70
3  9.5  1.59
4  20.6  2.52
5  26.7  2.86
6  25.2  2.87
};
\node[font=\tiny, text=black!70, anchor=south, yshift=1.6pt, inner sep=0.5pt] at (axis cs:0,12.1) {12.1};
\node[font=\tiny, text=black!70, anchor=north, yshift=-1.6pt, inner sep=0.5pt] at (axis cs:1,9.8) {9.8};
\node[font=\tiny, text=black!70, anchor=north, yshift=-1.6pt, inner sep=0.5pt] at (axis cs:2,6.1) {6.1};
\node[font=\tiny, text=black!70, anchor=south, yshift=1.6pt, inner sep=0.5pt] at (axis cs:3,9.5) {9.5};
\node[font=\tiny, text=black!70, anchor=south, yshift=1.6pt, inner sep=0.5pt] at (axis cs:4,20.6) {20.6};
\node[font=\tiny, text=black!70, anchor=south, yshift=1.6pt, inner sep=0.5pt] at (axis cs:5,26.7) {26.7};
\node[font=\tiny, text=black!70, anchor=north, yshift=-1.6pt, inner sep=0.5pt] at (axis cs:6,25.2) {25.2};
\nextgroupplot[title={NQ}, ymin=-5, ymax=40, ytick={0,10,20,30,40}]
\addplot[name path=U, draw=none, forget plot] table[x=x, y expr=\thisrow{m}+\thisrow{sd}] {
x  m  sd
0  4.0  1.10
1  5.2  1.26
2  11.5  1.95
3  15.1  1.95
4  28.0  2.49
5  28.1  2.53
6  30.7  2.72
};
\addplot[name path=L, draw=none, forget plot] table[x=x, y expr=\thisrow{m}-\thisrow{sd}] {
x  m  sd
0  4.0  1.10
1  5.2  1.26
2  11.5  1.95
3  15.1  1.95
4  28.0  2.49
5  28.1  2.53
6  30.7  2.72
};
\addplot[lineO, fill opacity=0.15, draw=none, forget plot] fill between[of=U and L];
\addplot[lineO, line width=0.9pt, mark=*, mark size=1.4pt, mark options={fill=lineO, draw=white, line width=0.35pt}] table[x=x, y=m] {
x  m  sd
0  4.0  1.10
1  5.2  1.26
2  11.5  1.95
3  15.1  1.95
4  28.0  2.49
5  28.1  2.53
6  30.7  2.72
};
\node[font=\tiny, text=black!70, anchor=north, yshift=-1.6pt, inner sep=0.5pt] at (axis cs:0,4.0) {4.0};
\node[font=\tiny, text=black!70, anchor=south, yshift=1.6pt, inner sep=0.5pt] at (axis cs:1,5.2) {5.2};
\node[font=\tiny, text=black!70, anchor=south, yshift=1.6pt, inner sep=0.5pt] at (axis cs:2,11.5) {11.5};
\node[font=\tiny, text=black!70, anchor=north, yshift=-1.6pt, inner sep=0.5pt] at (axis cs:3,15.1) {15.1};
\node[font=\tiny, text=black!70, anchor=south, yshift=1.6pt, inner sep=0.5pt] at (axis cs:4,28.0) {28.0};
\node[font=\tiny, text=black!70, anchor=north, yshift=-1.6pt, inner sep=0.5pt] at (axis cs:5,28.1) {28.1};
\node[font=\tiny, text=black!70, anchor=south, yshift=1.6pt, inner sep=0.5pt] at (axis cs:6,30.7) {30.7};
\nextgroupplot[title={TriviaQA}, ymin=-5, ymax=45, ytick={0,10,20,30,40}, ylabel={F1}, ylabel style={font=\scriptsize, yshift=-3pt}]
\addplot[name path=U, draw=none, forget plot] table[x=x, y expr=\thisrow{m}+\thisrow{sd}] {
x  m  sd
0  7.9  1.32
1  8.5  1.46
2  2.7  0.76
3  7.3  1.05
4  28.5  2.27
5  27.9  2.20
6  25.2  2.01
};
\addplot[name path=L, draw=none, forget plot] table[x=x, y expr=\thisrow{m}-\thisrow{sd}] {
x  m  sd
0  7.9  1.32
1  8.5  1.46
2  2.7  0.76
3  7.3  1.05
4  28.5  2.27
5  27.9  2.20
6  25.2  2.01
};
\addplot[lineO, fill opacity=0.15, draw=none, forget plot] fill between[of=U and L];
\addplot[lineO, line width=0.9pt, mark=*, mark size=1.4pt, mark options={fill=lineO, draw=white, line width=0.35pt}] table[x=x, y=m] {
x  m  sd
0  7.9  1.32
1  8.5  1.46
2  2.7  0.76
3  7.3  1.05
4  28.5  2.27
5  27.9  2.20
6  25.2  2.01
};
\node[font=\tiny, text=black!70, anchor=north, yshift=-1.6pt, inner sep=0.5pt] at (axis cs:0,7.9) {7.9};
\node[font=\tiny, text=black!70, anchor=south, yshift=1.6pt, inner sep=0.5pt] at (axis cs:1,8.5) {8.5};
\node[font=\tiny, text=black!70, anchor=north, yshift=-1.6pt, inner sep=0.5pt] at (axis cs:2,2.7) {2.7};
\node[font=\tiny, text=black!70, anchor=north, yshift=-1.6pt, inner sep=0.5pt] at (axis cs:3,7.3) {7.3};
\node[font=\tiny, text=black!70, anchor=south, yshift=1.6pt, inner sep=0.5pt] at (axis cs:4,28.5) {28.5};
\node[font=\tiny, text=black!70, anchor=north, yshift=-1.6pt, inner sep=0.5pt] at (axis cs:5,27.9) {27.9};
\node[font=\tiny, text=black!70, anchor=north, yshift=-1.6pt, inner sep=0.5pt] at (axis cs:6,25.2) {25.2};
\nextgroupplot[title={Qasper}, ymin=-5, ymax=40, ytick={0,10,20,30,40}]
\addplot[name path=U, draw=none, forget plot] table[x=x, y expr=\thisrow{m}+\thisrow{sd}] {
x  m  sd
0  1.9  0.54
1  12.5  1.28
2  16.4  1.65
3  12.8  1.10
4  23.2  1.37
5  30.4  1.80
6  28.2  1.75
};
\addplot[name path=L, draw=none, forget plot] table[x=x, y expr=\thisrow{m}-\thisrow{sd}] {
x  m  sd
0  1.9  0.54
1  12.5  1.28
2  16.4  1.65
3  12.8  1.10
4  23.2  1.37
5  30.4  1.80
6  28.2  1.75
};
\addplot[lineO, fill opacity=0.15, draw=none, forget plot] fill between[of=U and L];
\addplot[lineO, line width=0.9pt, mark=*, mark size=1.4pt, mark options={fill=lineO, draw=white, line width=0.35pt}] table[x=x, y=m] {
x  m  sd
0  1.9  0.54
1  12.5  1.28
2  16.4  1.65
3  12.8  1.10
4  23.2  1.37
5  30.4  1.80
6  28.2  1.75
};
\node[font=\tiny, text=black!70, anchor=north, yshift=-1.6pt, inner sep=0.5pt] at (axis cs:0,1.9) {1.9};
\node[font=\tiny, text=black!70, anchor=south, yshift=1.6pt, inner sep=0.5pt] at (axis cs:1,12.5) {12.5};
\node[font=\tiny, text=black!70, anchor=south, yshift=1.6pt, inner sep=0.5pt] at (axis cs:2,16.4) {16.4};
\node[font=\tiny, text=black!70, anchor=north, yshift=-1.6pt, inner sep=0.5pt] at (axis cs:3,12.8) {12.8};
\node[font=\tiny, text=black!70, anchor=south, yshift=1.6pt, inner sep=0.5pt] at (axis cs:4,23.2) {23.2};
\node[font=\tiny, text=black!70, anchor=south, yshift=1.6pt, inner sep=0.5pt] at (axis cs:5,30.4) {30.4};
\node[font=\tiny, text=black!70, anchor=north, yshift=-1.6pt, inner sep=0.5pt] at (axis cs:6,28.2) {28.2};
\nextgroupplot[title={MultiFieldQA}, ymin=-5, ymax=30, ytick={0,10,20,30}]
\addplot[name path=U, draw=none, forget plot] table[x=x, y expr=\thisrow{m}+\thisrow{sd}] {
x  m  sd
0  1.9  0.73
1  2.4  0.97
2  8.1  1.98
3  15.4  2.16
4  20.6  3.04
5  18.7  2.66
6  13.8  2.10
};
\addplot[name path=L, draw=none, forget plot] table[x=x, y expr=\thisrow{m}-\thisrow{sd}] {
x  m  sd
0  1.9  0.73
1  2.4  0.97
2  8.1  1.98
3  15.4  2.16
4  20.6  3.04
5  18.7  2.66
6  13.8  2.10
};
\addplot[lineO, fill opacity=0.15, draw=none, forget plot] fill between[of=U and L];
\addplot[lineO, line width=0.9pt, mark=*, mark size=1.4pt, mark options={fill=lineO, draw=white, line width=0.35pt}] table[x=x, y=m] {
x  m  sd
0  1.9  0.73
1  2.4  0.97
2  8.1  1.98
3  15.4  2.16
4  20.6  3.04
5  18.7  2.66
6  13.8  2.10
};
\node[font=\tiny, text=black!70, anchor=north, yshift=-1.6pt, inner sep=0.5pt] at (axis cs:0,1.9) {1.9};
\node[font=\tiny, text=black!70, anchor=south, yshift=1.6pt, inner sep=0.5pt] at (axis cs:1,2.4) {2.4};
\node[font=\tiny, text=black!70, anchor=south, yshift=1.6pt, inner sep=0.5pt] at (axis cs:2,8.1) {8.1};
\node[font=\tiny, text=black!70, anchor=south, yshift=1.6pt, inner sep=0.5pt] at (axis cs:3,15.4) {15.4};
\node[font=\tiny, text=black!70, anchor=south, yshift=1.6pt, inner sep=0.5pt] at (axis cs:4,20.6) {20.6};
\node[font=\tiny, text=black!70, anchor=north, yshift=-1.6pt, inner sep=0.5pt] at (axis cs:5,18.7) {18.7};
\node[font=\tiny, text=black!70, anchor=north, yshift=-1.6pt, inner sep=0.5pt] at (axis cs:6,13.8) {13.8};
\nextgroupplot[title={NarrativeQA}, ymin=-5, ymax=22, ytick={0,5,10,15,20}]
\addplot[name path=U, draw=none, forget plot] table[x=x, y expr=\thisrow{m}+\thisrow{sd}] {
x  m  sd
0  0.5  0.22
1  1.3  0.29
2  1.2  0.26
3  2.3  0.45
4  13.5  0.92
5  12.8  0.93
6  14.1  0.88
};
\addplot[name path=L, draw=none, forget plot] table[x=x, y expr=\thisrow{m}-\thisrow{sd}] {
x  m  sd
0  0.5  0.22
1  1.3  0.29
2  1.2  0.26
3  2.3  0.45
4  13.5  0.92
5  12.8  0.93
6  14.1  0.88
};
\addplot[lineO, fill opacity=0.15, draw=none, forget plot] fill between[of=U and L];
\addplot[lineO, line width=0.9pt, mark=*, mark size=1.4pt, mark options={fill=lineO, draw=white, line width=0.35pt}] table[x=x, y=m] {
x  m  sd
0  0.5  0.22
1  1.3  0.29
2  1.2  0.26
3  2.3  0.45
4  13.5  0.92
5  12.8  0.93
6  14.1  0.88
};
\node[font=\tiny, text=black!70, anchor=north, yshift=-1.6pt, inner sep=0.5pt] at (axis cs:0,0.5) {0.5};
\node[font=\tiny, text=black!70, anchor=south, yshift=1.6pt, inner sep=0.5pt] at (axis cs:1,1.3) {1.3};
\node[font=\tiny, text=black!70, anchor=north, yshift=-1.6pt, inner sep=0.5pt] at (axis cs:2,1.2) {1.2};
\node[font=\tiny, text=black!70, anchor=south, yshift=1.6pt, inner sep=0.5pt] at (axis cs:3,2.3) {2.3};
\node[font=\tiny, text=black!70, anchor=south, yshift=1.6pt, inner sep=0.5pt] at (axis cs:4,13.5) {13.5};
\node[font=\tiny, text=black!70, anchor=north, yshift=-1.6pt, inner sep=0.5pt] at (axis cs:5,12.8) {12.8};
\node[font=\tiny, text=black!70, anchor=south, yshift=1.6pt, inner sep=0.5pt] at (axis cs:6,14.1) {14.1};
\nextgroupplot[title={LoCoMo}, ymin=-5, ymax=30, ytick={0,10,20,30}]
\addplot[name path=U, draw=none, forget plot] table[x=x, y expr=\thisrow{m}+\thisrow{sd}] {
x  m  sd
0  0.6  0.13
1  1.3  0.24
2  3.6  0.40
3  4.7  0.43
4  19.1  0.82
5  18.6  0.81
6  21.0  0.86
};
\addplot[name path=L, draw=none, forget plot] table[x=x, y expr=\thisrow{m}-\thisrow{sd}] {
x  m  sd
0  0.6  0.13
1  1.3  0.24
2  3.6  0.40
3  4.7  0.43
4  19.1  0.82
5  18.6  0.81
6  21.0  0.86
};
\addplot[lineO, fill opacity=0.15, draw=none, forget plot] fill between[of=U and L];
\addplot[lineO, line width=0.9pt, mark=*, mark size=1.4pt, mark options={fill=lineO, draw=white, line width=0.35pt}] table[x=x, y=m] {
x  m  sd
0  0.6  0.13
1  1.3  0.24
2  3.6  0.40
3  4.7  0.43
4  19.1  0.82
5  18.6  0.81
6  21.0  0.86
};
\node[font=\tiny, text=black!70, anchor=north, yshift=-1.6pt, inner sep=0.5pt] at (axis cs:0,0.6) {0.6};
\node[font=\tiny, text=black!70, anchor=south, yshift=1.6pt, inner sep=0.5pt] at (axis cs:1,1.3) {1.3};
\node[font=\tiny, text=black!70, anchor=south, yshift=1.6pt, inner sep=0.5pt] at (axis cs:2,3.6) {3.6};
\node[font=\tiny, text=black!70, anchor=north, yshift=-1.6pt, inner sep=0.5pt] at (axis cs:3,4.7) {4.7};
\node[font=\tiny, text=black!70, anchor=south, yshift=1.6pt, inner sep=0.5pt] at (axis cs:4,19.1) {19.1};
\node[font=\tiny, text=black!70, anchor=north, yshift=-1.6pt, inner sep=0.5pt] at (axis cs:5,18.6) {18.6};
\node[font=\tiny, text=black!70, anchor=south, yshift=1.6pt, inner sep=0.5pt] at (axis cs:6,21.0) {21.0};
\end{groupplot}
\end{tikzpicture}
\caption{F1 of RoDeL-Seq with Qwen3.5-0.8B on each dataset after each update. HotpotQA and MuSiQue (ID) are in-domain and the others out-of-domain. Fr.: frozen roles (PRAG-style). Shaded bands show $\pm 1$ standard deviation of a test-set bootstrap.}
\label{fig:perset08}
\end{figure}

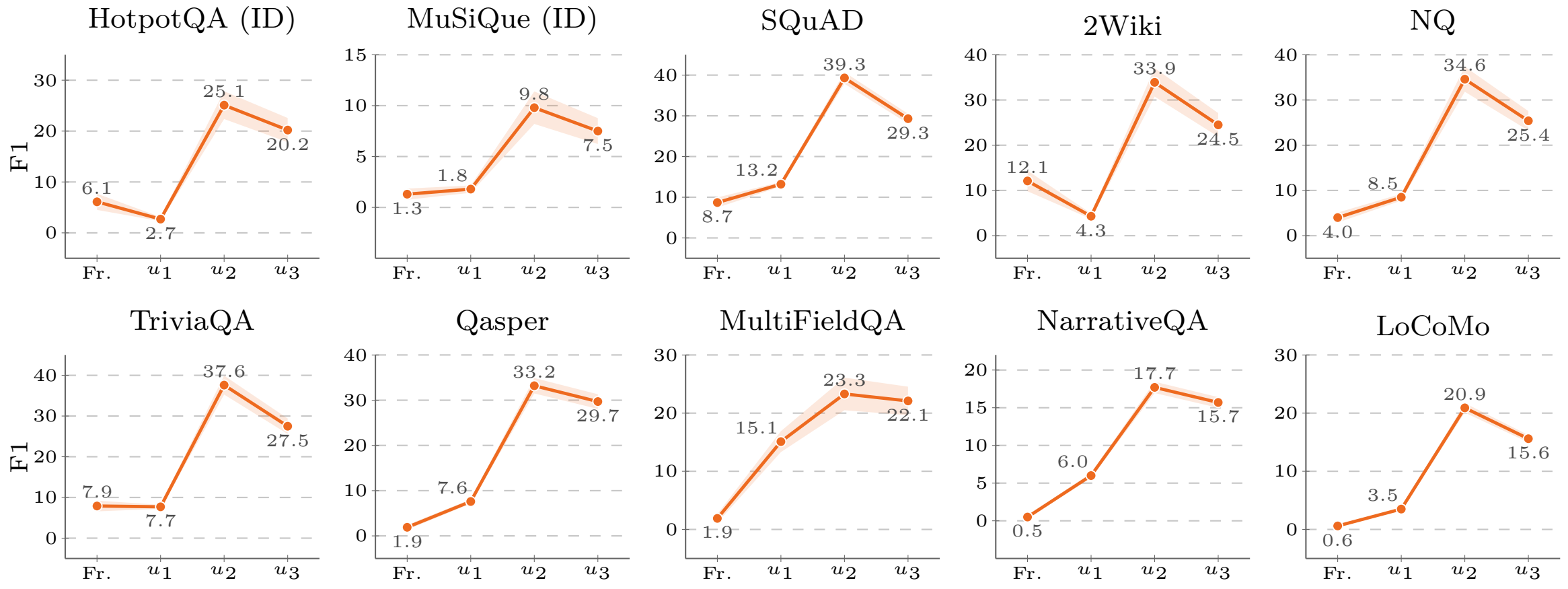
\begin{figure}[!htb]
\centering
\begin{tikzpicture}
\begin{groupplot}[
  group style={group size=5 by 2, horizontal sep=0.55cm, vertical sep=0.95cm},
  scale only axis, width=2.5cm, height=2.0cm,
  xmin=-0.5, xmax=3.5,
  xtick={0,...,3},
  xticklabels={Fr.,$u_1$,$u_2$,$u_3$},
  axis lines=left, axis line style={-, black!60}, tick style={black!60}, major tick length=2pt,
  tick label style={font=\tiny, inner sep=1pt},
  title style={font=\footnotesize, yshift=-4pt},
  ymajorgrids, grid style={black!22, dashed},
  clip=false,
]
\nextgroupplot[title={HotpotQA (ID)}, ymin=-5, ymax=35, ytick={0,10,20,30}, ylabel={F1}, ylabel style={font=\scriptsize, yshift=-3pt}]
\addplot[name path=U, draw=none, forget plot] table[x=x, y expr=\thisrow{m}+\thisrow{sd}] {
x  m  sd
0  6.1  1.59
1  2.7  0.41
2  25.1  2.73
3  20.2  2.38
};
\addplot[name path=L, draw=none, forget plot] table[x=x, y expr=\thisrow{m}-\thisrow{sd}] {
x  m  sd
0  6.1  1.59
1  2.7  0.41
2  25.1  2.73
3  20.2  2.38
};
\addplot[lineO, fill opacity=0.15, draw=none, forget plot] fill between[of=U and L];
\addplot[lineO, line width=0.9pt, mark=*, mark size=1.4pt, mark options={fill=lineO, draw=white, line width=0.35pt}] table[x=x, y=m] {
x  m  sd
0  6.1  1.59
1  2.7  0.41
2  25.1  2.73
3  20.2  2.38
};
\node[font=\tiny, text=black!70, anchor=south, yshift=1.6pt, inner sep=0.5pt] at (axis cs:0,6.1) {6.1};
\node[font=\tiny, text=black!70, anchor=north, yshift=-1.6pt, inner sep=0.5pt] at (axis cs:1,2.7) {2.7};
\node[font=\tiny, text=black!70, anchor=south, yshift=1.6pt, inner sep=0.5pt] at (axis cs:2,25.1) {25.1};
\node[font=\tiny, text=black!70, anchor=north, yshift=-1.6pt, inner sep=0.5pt] at (axis cs:3,20.2) {20.2};
\nextgroupplot[title={MuSiQue (ID)}, ymin=-5, ymax=15, ytick={0,5,10,15}]
\addplot[name path=U, draw=none, forget plot] table[x=x, y expr=\thisrow{m}+\thisrow{sd}] {
x  m  sd
0  1.3  0.55
1  1.8  0.36
2  9.8  1.60
3  7.5  1.27
};
\addplot[name path=L, draw=none, forget plot] table[x=x, y expr=\thisrow{m}-\thisrow{sd}] {
x  m  sd
0  1.3  0.55
1  1.8  0.36
2  9.8  1.60
3  7.5  1.27
};
\addplot[lineO, fill opacity=0.15, draw=none, forget plot] fill between[of=U and L];
\addplot[lineO, line width=0.9pt, mark=*, mark size=1.4pt, mark options={fill=lineO, draw=white, line width=0.35pt}] table[x=x, y=m] {
x  m  sd
0  1.3  0.55
1  1.8  0.36
2  9.8  1.60
3  7.5  1.27
};
\node[font=\tiny, text=black!70, anchor=north, yshift=-1.6pt, inner sep=0.5pt] at (axis cs:0,1.3) {1.3};
\node[font=\tiny, text=black!70, anchor=south east, yshift=1.6pt, inner sep=0.5pt] at (axis cs:1,1.8) {1.8};
\node[font=\tiny, text=black!70, anchor=south, yshift=1.6pt, inner sep=0.5pt] at (axis cs:2,9.8) {9.8};
\node[font=\tiny, text=black!70, anchor=north, yshift=-1.6pt, inner sep=0.5pt] at (axis cs:3,7.5) {7.5};
\nextgroupplot[title={SQuAD}, ymin=-5, ymax=45, ytick={0,10,20,30,40}]
\addplot[name path=U, draw=none, forget plot] table[x=x, y expr=\thisrow{m}+\thisrow{sd}] {
x  m  sd
0  8.7  1.25
1  13.2  0.56
2  39.3  1.43
3  29.3  1.21
};
\addplot[name path=L, draw=none, forget plot] table[x=x, y expr=\thisrow{m}-\thisrow{sd}] {
x  m  sd
0  8.7  1.25
1  13.2  0.56
2  39.3  1.43
3  29.3  1.21
};
\addplot[lineO, fill opacity=0.15, draw=none, forget plot] fill between[of=U and L];
\addplot[lineO, line width=0.9pt, mark=*, mark size=1.4pt, mark options={fill=lineO, draw=white, line width=0.35pt}] table[x=x, y=m] {
x  m  sd
0  8.7  1.25
1  13.2  0.56
2  39.3  1.43
3  29.3  1.21
};
\node[font=\tiny, text=black!70, anchor=north, yshift=-1.6pt, inner sep=0.5pt] at (axis cs:0,8.7) {8.7};
\node[font=\tiny, text=black!70, anchor=south east, yshift=1.6pt, inner sep=0.5pt] at (axis cs:1,13.2) {13.2};
\node[font=\tiny, text=black!70, anchor=south, yshift=1.6pt, inner sep=0.5pt] at (axis cs:2,39.3) {39.3};
\node[font=\tiny, text=black!70, anchor=north, yshift=-1.6pt, inner sep=0.5pt] at (axis cs:3,29.3) {29.3};
\nextgroupplot[title={2Wiki}, ymin=-5, ymax=40, ytick={0,10,20,30,40}]
\addplot[name path=U, draw=none, forget plot] table[x=x, y expr=\thisrow{m}+\thisrow{sd}] {
x  m  sd
0  12.1  2.28
1  4.3  0.61
2  33.9  3.15
3  24.5  2.61
};
\addplot[name path=L, draw=none, forget plot] table[x=x, y expr=\thisrow{m}-\thisrow{sd}] {
x  m  sd
0  12.1  2.28
1  4.3  0.61
2  33.9  3.15
3  24.5  2.61
};
\addplot[lineO, fill opacity=0.15, draw=none, forget plot] fill between[of=U and L];
\addplot[lineO, line width=0.9pt, mark=*, mark size=1.4pt, mark options={fill=lineO, draw=white, line width=0.35pt}] table[x=x, y=m] {
x  m  sd
0  12.1  2.28
1  4.3  0.61
2  33.9  3.15
3  24.5  2.61
};
\node[font=\tiny, text=black!70, anchor=south, yshift=1.6pt, inner sep=0.5pt] at (axis cs:0,12.1) {12.1};
\node[font=\tiny, text=black!70, anchor=north, yshift=-1.6pt, inner sep=0.5pt] at (axis cs:1,4.3) {4.3};
\node[font=\tiny, text=black!70, anchor=south, yshift=1.6pt, inner sep=0.5pt] at (axis cs:2,33.9) {33.9};
\node[font=\tiny, text=black!70, anchor=north, yshift=-1.6pt, inner sep=0.5pt] at (axis cs:3,24.5) {24.5};
\nextgroupplot[title={NQ}, ymin=-5, ymax=40, ytick={0,10,20,30,40}]
\addplot[name path=U, draw=none, forget plot] table[x=x, y expr=\thisrow{m}+\thisrow{sd}] {
x  m  sd
0  4.0  1.10
1  8.5  0.75
2  34.6  2.78
3  25.4  2.15
};
\addplot[name path=L, draw=none, forget plot] table[x=x, y expr=\thisrow{m}-\thisrow{sd}] {
x  m  sd
0  4.0  1.10
1  8.5  0.75
2  34.6  2.78
3  25.4  2.15
};
\addplot[lineO, fill opacity=0.15, draw=none, forget plot] fill between[of=U and L];
\addplot[lineO, line width=0.9pt, mark=*, mark size=1.4pt, mark options={fill=lineO, draw=white, line width=0.35pt}] table[x=x, y=m] {
x  m  sd
0  4.0  1.10
1  8.5  0.75
2  34.6  2.78
3  25.4  2.15
};
\node[font=\tiny, text=black!70, anchor=north, yshift=-1.6pt, inner sep=0.5pt] at (axis cs:0,4.0) {4.0};
\node[font=\tiny, text=black!70, anchor=south east, yshift=1.6pt, inner sep=0.5pt] at (axis cs:1,8.5) {8.5};
\node[font=\tiny, text=black!70, anchor=south, yshift=1.6pt, inner sep=0.5pt] at (axis cs:2,34.6) {34.6};
\node[font=\tiny, text=black!70, anchor=north, yshift=-1.6pt, inner sep=0.5pt] at (axis cs:3,25.4) {25.4};
\nextgroupplot[title={TriviaQA}, ymin=-5, ymax=45, ytick={0,10,20,30,40}, ylabel={F1}, ylabel style={font=\scriptsize, yshift=-3pt}]
\addplot[name path=U, draw=none, forget plot] table[x=x, y expr=\thisrow{m}+\thisrow{sd}] {
x  m  sd
0  7.9  1.32
1  7.7  0.50
2  37.6  2.33
3  27.5  1.90
};
\addplot[name path=L, draw=none, forget plot] table[x=x, y expr=\thisrow{m}-\thisrow{sd}] {
x  m  sd
0  7.9  1.32
1  7.7  0.50
2  37.6  2.33
3  27.5  1.90
};
\addplot[lineO, fill opacity=0.15, draw=none, forget plot] fill between[of=U and L];
\addplot[lineO, line width=0.9pt, mark=*, mark size=1.4pt, mark options={fill=lineO, draw=white, line width=0.35pt}] table[x=x, y=m] {
x  m  sd
0  7.9  1.32
1  7.7  0.50
2  37.6  2.33
3  27.5  1.90
};
\node[font=\tiny, text=black!70, anchor=south, yshift=1.6pt, inner sep=0.5pt] at (axis cs:0,7.9) {7.9};
\node[font=\tiny, text=black!70, anchor=north, yshift=-1.6pt, inner sep=0.5pt] at (axis cs:1,7.7) {7.7};
\node[font=\tiny, text=black!70, anchor=south, yshift=1.6pt, inner sep=0.5pt] at (axis cs:2,37.6) {37.6};
\node[font=\tiny, text=black!70, anchor=north, yshift=-1.6pt, inner sep=0.5pt] at (axis cs:3,27.5) {27.5};
\nextgroupplot[title={Qasper}, ymin=-5, ymax=40, ytick={0,10,20,30,40}]
\addplot[name path=U, draw=none, forget plot] table[x=x, y expr=\thisrow{m}+\thisrow{sd}] {
x  m  sd
0  1.9  0.54
1  7.6  0.47
2  33.2  1.71
3  29.7  1.62
};
\addplot[name path=L, draw=none, forget plot] table[x=x, y expr=\thisrow{m}-\thisrow{sd}] {
x  m  sd
0  1.9  0.54
1  7.6  0.47
2  33.2  1.71
3  29.7  1.62
};
\addplot[lineO, fill opacity=0.15, draw=none, forget plot] fill between[of=U and L];
\addplot[lineO, line width=0.9pt, mark=*, mark size=1.4pt, mark options={fill=lineO, draw=white, line width=0.35pt}] table[x=x, y=m] {
x  m  sd
0  1.9  0.54
1  7.6  0.47
2  33.2  1.71
3  29.7  1.62
};
\node[font=\tiny, text=black!70, anchor=north, yshift=-1.6pt, inner sep=0.5pt] at (axis cs:0,1.9) {1.9};
\node[font=\tiny, text=black!70, anchor=south east, yshift=1.6pt, inner sep=0.5pt] at (axis cs:1,7.6) {7.6};
\node[font=\tiny, text=black!70, anchor=south, yshift=1.6pt, inner sep=0.5pt] at (axis cs:2,33.2) {33.2};
\node[font=\tiny, text=black!70, anchor=north, yshift=-1.6pt, inner sep=0.5pt] at (axis cs:3,29.7) {29.7};
\nextgroupplot[title={MultiFieldQA}, ymin=-5, ymax=30, ytick={0,10,20,30}]
\addplot[name path=U, draw=none, forget plot] table[x=x, y expr=\thisrow{m}+\thisrow{sd}] {
x  m  sd
0  1.9  0.73
1  15.1  1.80
2  23.3  2.81
3  22.1  2.45
};
\addplot[name path=L, draw=none, forget plot] table[x=x, y expr=\thisrow{m}-\thisrow{sd}] {
x  m  sd
0  1.9  0.73
1  15.1  1.80
2  23.3  2.81
3  22.1  2.45
};
\addplot[lineO, fill opacity=0.15, draw=none, forget plot] fill between[of=U and L];
\addplot[lineO, line width=0.9pt, mark=*, mark size=1.4pt, mark options={fill=lineO, draw=white, line width=0.35pt}] table[x=x, y=m] {
x  m  sd
0  1.9  0.73
1  15.1  1.80
2  23.3  2.81
3  22.1  2.45
};
\node[font=\tiny, text=black!70, anchor=north, yshift=-1.6pt, inner sep=0.5pt] at (axis cs:0,1.9) {1.9};
\node[font=\tiny, text=black!70, anchor=south east, yshift=1.6pt, inner sep=0.5pt] at (axis cs:1,15.1) {15.1};
\node[font=\tiny, text=black!70, anchor=south, yshift=1.6pt, inner sep=0.5pt] at (axis cs:2,23.3) {23.3};
\node[font=\tiny, text=black!70, anchor=north, yshift=-1.6pt, inner sep=0.5pt] at (axis cs:3,22.1) {22.1};
\nextgroupplot[title={NarrativeQA}, ymin=-5, ymax=22, ytick={0,5,10,15,20}]
\addplot[name path=U, draw=none, forget plot] table[x=x, y expr=\thisrow{m}+\thisrow{sd}] {
x  m  sd
0  0.5  0.22
1  6.0  0.32
2  17.7  0.75
3  15.7  0.71
};
\addplot[name path=L, draw=none, forget plot] table[x=x, y expr=\thisrow{m}-\thisrow{sd}] {
x  m  sd
0  0.5  0.22
1  6.0  0.32
2  17.7  0.75
3  15.7  0.71
};
\addplot[lineO, fill opacity=0.15, draw=none, forget plot] fill between[of=U and L];
\addplot[lineO, line width=0.9pt, mark=*, mark size=1.4pt, mark options={fill=lineO, draw=white, line width=0.35pt}] table[x=x, y=m] {
x  m  sd
0  0.5  0.22
1  6.0  0.32
2  17.7  0.75
3  15.7  0.71
};
\node[font=\tiny, text=black!70, anchor=north, yshift=-1.6pt, inner sep=0.5pt] at (axis cs:0,0.5) {0.5};
\node[font=\tiny, text=black!70, anchor=south east, yshift=1.6pt, inner sep=0.5pt] at (axis cs:1,6.0) {6.0};
\node[font=\tiny, text=black!70, anchor=south, yshift=1.6pt, inner sep=0.5pt] at (axis cs:2,17.7) {17.7};
\node[font=\tiny, text=black!70, anchor=north, yshift=-1.6pt, inner sep=0.5pt] at (axis cs:3,15.7) {15.7};
\nextgroupplot[title={LoCoMo}, ymin=-5, ymax=30, ytick={0,10,20,30}]
\addplot[name path=U, draw=none, forget plot] table[x=x, y expr=\thisrow{m}+\thisrow{sd}] {
x  m  sd
0  0.6  0.13
1  3.5  0.20
2  20.9  0.82
3  15.6  0.63
};
\addplot[name path=L, draw=none, forget plot] table[x=x, y expr=\thisrow{m}-\thisrow{sd}] {
x  m  sd
0  0.6  0.13
1  3.5  0.20
2  20.9  0.82
3  15.6  0.63
};
\addplot[lineO, fill opacity=0.15, draw=none, forget plot] fill between[of=U and L];
\addplot[lineO, line width=0.9pt, mark=*, mark size=1.4pt, mark options={fill=lineO, draw=white, line width=0.35pt}] table[x=x, y=m] {
x  m  sd
0  0.6  0.13
1  3.5  0.20
2  20.9  0.82
3  15.6  0.63
};
\node[font=\tiny, text=black!70, anchor=north, yshift=-1.6pt, inner sep=0.5pt] at (axis cs:0,0.6) {0.6};
\node[font=\tiny, text=black!70, anchor=south east, yshift=1.6pt, inner sep=0.5pt] at (axis cs:1,3.5) {3.5};
\node[font=\tiny, text=black!70, anchor=south, yshift=1.6pt, inner sep=0.5pt] at (axis cs:2,20.9) {20.9};
\node[font=\tiny, text=black!70, anchor=north, yshift=-1.6pt, inner sep=0.5pt] at (axis cs:3,15.6) {15.6};
\end{groupplot}
\end{tikzpicture}
\caption{F1 of RoDeL-Sim with Qwen3.5-0.8B on each dataset after each simultaneous update. Datasets, Fr., and shaded bands are as in \Cref{fig:perset08}.}
\label{fig:perset08sim}
\end{figure}

\FloatBarrier

\section{Other Model Family}
\label{app:gemma}

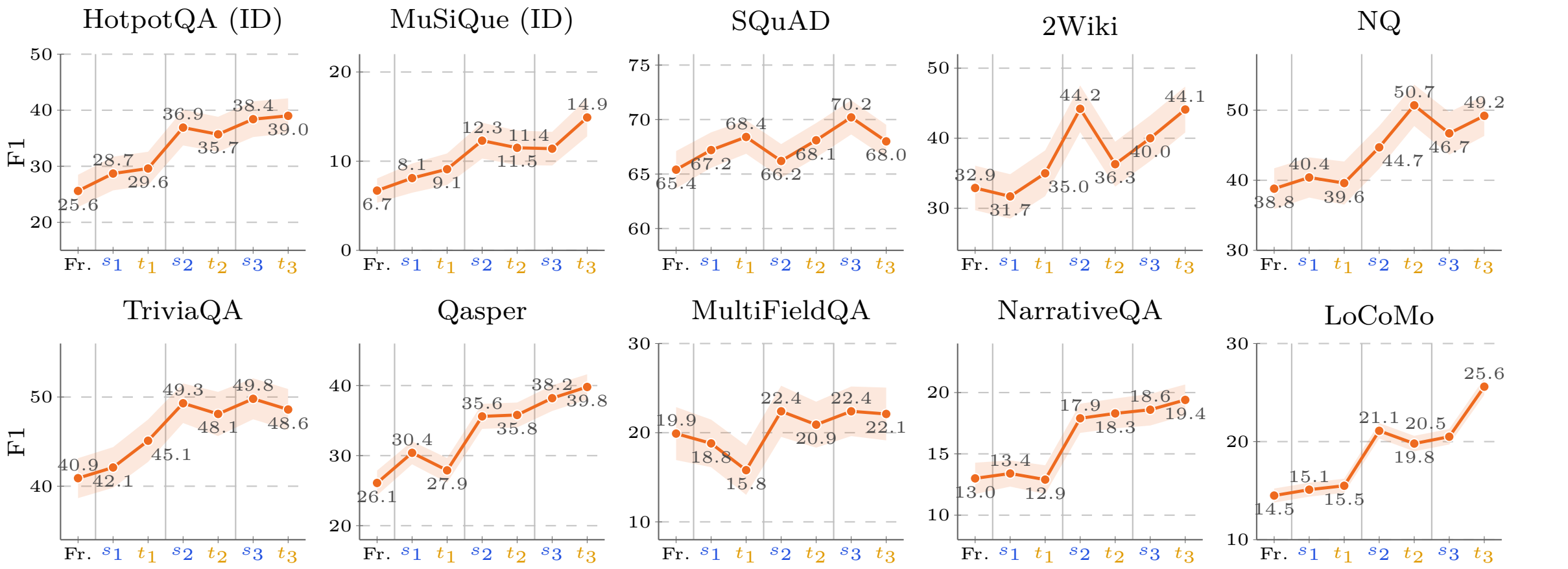
\begin{figure}[!htb]
\centering
\begin{tikzpicture}
\begin{groupplot}[
  group style={group size=5 by 2, horizontal sep=0.55cm, vertical sep=0.95cm},
  scale only axis, width=2.5cm, height=2.0cm,
  xmin=-0.5, xmax=6.5,
  xtick={0,...,6},
  xticklabels={Fr.,\textcolor{stuV}{$s_1$},\textcolor{tchV}{$t_1$},\textcolor{stuV}{$s_2$},\textcolor{tchV}{$t_2$},\textcolor{stuV}{$s_3$},\textcolor{tchV}{$t_3$}},
  axis lines=left, axis line style={-, black!60}, tick style={black!60}, major tick length=2pt,
  tick label style={font=\tiny, inner sep=1pt},
  title style={font=\footnotesize, yshift=-4pt},
  ymajorgrids, grid style={black!22, dashed},
  extra x ticks={0.5,2.5,4.5}, extra x tick labels={},
  extra x tick style={grid=major, major grid style={black!25, solid}, tick style={draw=none}},
  clip=false,
]
\nextgroupplot[title={HotpotQA (ID)}, ymin=15, ymax=50, ytick={20,30,40,50}, ylabel={F1}, ylabel style={font=\scriptsize, yshift=-3pt}]
\addplot[name path=U, draw=none, forget plot] table[x=x, y expr=\thisrow{m}+\thisrow{sd}] {
x  m  sd
0  25.6  2.88
1  28.7  3.01
2  29.6  2.99
3  36.9  3.19
4  35.7  3.13
5  38.4  3.19
6  39.0  3.12
};
\addplot[name path=L, draw=none, forget plot] table[x=x, y expr=\thisrow{m}-\thisrow{sd}] {
x  m  sd
0  25.6  2.88
1  28.7  3.01
2  29.6  2.99
3  36.9  3.19
4  35.7  3.13
5  38.4  3.19
6  39.0  3.12
};
\addplot[lineO, fill opacity=0.15, draw=none, forget plot] fill between[of=U and L];
\addplot[lineO, line width=0.9pt, mark=*, mark size=1.4pt, mark options={fill=lineO, draw=white, line width=0.35pt}] table[x=x, y=m] {
x  m  sd
0  25.6  2.88
1  28.7  3.01
2  29.6  2.99
3  36.9  3.19
4  35.7  3.13
5  38.4  3.19
6  39.0  3.12
};
\node[font=\tiny, text=black!70, anchor=north, yshift=-1.6pt, inner sep=0.5pt] at (axis cs:0,25.6) {25.6};
\node[font=\tiny, text=black!70, anchor=south, yshift=1.6pt, inner sep=0.5pt] at (axis cs:1,28.7) {28.7};
\node[font=\tiny, text=black!70, anchor=north, yshift=-1.6pt, inner sep=0.5pt] at (axis cs:2,29.6) {29.6};
\node[font=\tiny, text=black!70, anchor=south, yshift=1.6pt, inner sep=0.5pt] at (axis cs:3,36.9) {36.9};
\node[font=\tiny, text=black!70, anchor=north, yshift=-1.6pt, inner sep=0.5pt] at (axis cs:4,35.7) {35.7};
\node[font=\tiny, text=black!70, anchor=south, yshift=1.6pt, inner sep=0.5pt] at (axis cs:5,38.4) {38.4};
\node[font=\tiny, text=black!70, anchor=north, yshift=-1.6pt, inner sep=0.5pt] at (axis cs:6,39.0) {39.0};
\nextgroupplot[title={MuSiQue (ID)}, ymin=0, ymax=22, ytick={0,10,20}]
\addplot[name path=U, draw=none, forget plot] table[x=x, y expr=\thisrow{m}+\thisrow{sd}] {
x  m  sd
0  6.7  1.36
1  8.1  1.64
2  9.1  1.77
3  12.3  2.01
4  11.5  1.94
5  11.4  1.89
6  14.9  2.12
};
\addplot[name path=L, draw=none, forget plot] table[x=x, y expr=\thisrow{m}-\thisrow{sd}] {
x  m  sd
0  6.7  1.36
1  8.1  1.64
2  9.1  1.77
3  12.3  2.01
4  11.5  1.94
5  11.4  1.89
6  14.9  2.12
};
\addplot[lineO, fill opacity=0.15, draw=none, forget plot] fill between[of=U and L];
\addplot[lineO, line width=0.9pt, mark=*, mark size=1.4pt, mark options={fill=lineO, draw=white, line width=0.35pt}] table[x=x, y=m] {
x  m  sd
0  6.7  1.36
1  8.1  1.64
2  9.1  1.77
3  12.3  2.01
4  11.5  1.94
5  11.4  1.89
6  14.9  2.12
};
\node[font=\tiny, text=black!70, anchor=north, yshift=-1.6pt, inner sep=0.5pt] at (axis cs:0,6.7) {6.7};
\node[font=\tiny, text=black!70, anchor=south, yshift=1.6pt, inner sep=0.5pt] at (axis cs:1,8.1) {8.1};
\node[font=\tiny, text=black!70, anchor=north, yshift=-1.6pt, inner sep=0.5pt] at (axis cs:2,9.1) {9.1};
\node[font=\tiny, text=black!70, anchor=south, yshift=1.6pt, inner sep=0.5pt] at (axis cs:3,12.3) {12.3};
\node[font=\tiny, text=black!70, anchor=north, yshift=-1.6pt, inner sep=0.5pt] at (axis cs:4,11.5) {11.5};
\node[font=\tiny, text=black!70, anchor=south east, yshift=1.6pt, inner sep=0.5pt] at (axis cs:5,11.4) {11.4};
\node[font=\tiny, text=black!70, anchor=south, yshift=1.6pt, inner sep=0.5pt] at (axis cs:6,14.9) {14.9};
\nextgroupplot[title={SQuAD}, ymin=58, ymax=76, ytick={60,65,70,75}]
\addplot[name path=U, draw=none, forget plot] table[x=x, y expr=\thisrow{m}+\thisrow{sd}] {
x  m  sd
0  65.4  1.71
1  67.2  1.63
2  68.4  1.54
3  66.2  1.55
4  68.1  1.58
5  70.2  1.57
6  68.0  1.52
};
\addplot[name path=L, draw=none, forget plot] table[x=x, y expr=\thisrow{m}-\thisrow{sd}] {
x  m  sd
0  65.4  1.71
1  67.2  1.63
2  68.4  1.54
3  66.2  1.55
4  68.1  1.58
5  70.2  1.57
6  68.0  1.52
};
\addplot[lineO, fill opacity=0.15, draw=none, forget plot] fill between[of=U and L];
\addplot[lineO, line width=0.9pt, mark=*, mark size=1.4pt, mark options={fill=lineO, draw=white, line width=0.35pt}] table[x=x, y=m] {
x  m  sd
0  65.4  1.71
1  67.2  1.63
2  68.4  1.54
3  66.2  1.55
4  68.1  1.58
5  70.2  1.57
6  68.0  1.52
};
\node[font=\tiny, text=black!70, anchor=north, yshift=-1.6pt, inner sep=0.5pt] at (axis cs:0,65.4) {65.4};
\node[font=\tiny, text=black!70, anchor=north, yshift=-1.6pt, inner sep=0.5pt] at (axis cs:1,67.2) {67.2};
\node[font=\tiny, text=black!70, anchor=south, yshift=1.6pt, inner sep=0.5pt] at (axis cs:2,68.4) {68.4};
\node[font=\tiny, text=black!70, anchor=north, yshift=-1.6pt, inner sep=0.5pt] at (axis cs:3,66.2) {66.2};
\node[font=\tiny, text=black!70, anchor=north, yshift=-1.6pt, inner sep=0.5pt] at (axis cs:4,68.1) {68.1};
\node[font=\tiny, text=black!70, anchor=south, yshift=1.6pt, inner sep=0.5pt] at (axis cs:5,70.2) {70.2};
\node[font=\tiny, text=black!70, anchor=north, yshift=-1.6pt, inner sep=0.5pt] at (axis cs:6,68.0) {68.0};
\nextgroupplot[title={2Wiki}, ymin=24, ymax=52, ytick={30,40,50}]
\addplot[name path=U, draw=none, forget plot] table[x=x, y expr=\thisrow{m}+\thisrow{sd}] {
x  m  sd
0  32.9  3.18
1  31.7  3.17
2  35.0  3.27
3  44.2  3.34
4  36.3  3.21
5  40.0  3.26
6  44.1  3.28
};
\addplot[name path=L, draw=none, forget plot] table[x=x, y expr=\thisrow{m}-\thisrow{sd}] {
x  m  sd
0  32.9  3.18
1  31.7  3.17
2  35.0  3.27
3  44.2  3.34
4  36.3  3.21
5  40.0  3.26
6  44.1  3.28
};
\addplot[lineO, fill opacity=0.15, draw=none, forget plot] fill between[of=U and L];
\addplot[lineO, line width=0.9pt, mark=*, mark size=1.4pt, mark options={fill=lineO, draw=white, line width=0.35pt}] table[x=x, y=m] {
x  m  sd
0  32.9  3.18
1  31.7  3.17
2  35.0  3.27
3  44.2  3.34
4  36.3  3.21
5  40.0  3.26
6  44.1  3.28
};
\node[font=\tiny, text=black!70, anchor=south, yshift=1.6pt, inner sep=0.5pt] at (axis cs:0,32.9) {32.9};
\node[font=\tiny, text=black!70, anchor=north, yshift=-1.6pt, inner sep=0.5pt] at (axis cs:1,31.7) {31.7};
\node[font=\tiny, text=black!70, anchor=north west, yshift=-1.6pt, inner sep=0.5pt] at (axis cs:2,35.0) {35.0};
\node[font=\tiny, text=black!70, anchor=south, yshift=1.6pt, inner sep=0.5pt] at (axis cs:3,44.2) {44.2};
\node[font=\tiny, text=black!70, anchor=north, yshift=-1.6pt, inner sep=0.5pt] at (axis cs:4,36.3) {36.3};
\node[font=\tiny, text=black!70, anchor=north, yshift=-1.6pt, inner sep=0.5pt] at (axis cs:5,40.0) {40.0};
\node[font=\tiny, text=black!70, anchor=south, yshift=1.6pt, inner sep=0.5pt] at (axis cs:6,44.1) {44.1};
\nextgroupplot[title={NQ}, ymin=30, ymax=58, ytick={30,40,50}]
\addplot[name path=U, draw=none, forget plot] table[x=x, y expr=\thisrow{m}+\thisrow{sd}] {
x  m  sd
0  38.8  2.91
1  40.4  2.87
2  39.6  3.08
3  44.7  3.05
4  50.7  2.96
5  46.7  3.05
6  49.2  2.87
};
\addplot[name path=L, draw=none, forget plot] table[x=x, y expr=\thisrow{m}-\thisrow{sd}] {
x  m  sd
0  38.8  2.91
1  40.4  2.87
2  39.6  3.08
3  44.7  3.05
4  50.7  2.96
5  46.7  3.05
6  49.2  2.87
};
\addplot[lineO, fill opacity=0.15, draw=none, forget plot] fill between[of=U and L];
\addplot[lineO, line width=0.9pt, mark=*, mark size=1.4pt, mark options={fill=lineO, draw=white, line width=0.35pt}] table[x=x, y=m] {
x  m  sd
0  38.8  2.91
1  40.4  2.87
2  39.6  3.08
3  44.7  3.05
4  50.7  2.96
5  46.7  3.05
6  49.2  2.87
};
\node[font=\tiny, text=black!70, anchor=north, yshift=-1.6pt, inner sep=0.5pt] at (axis cs:0,38.8) {38.8};
\node[font=\tiny, text=black!70, anchor=south, yshift=1.6pt, inner sep=0.5pt] at (axis cs:1,40.4) {40.4};
\node[font=\tiny, text=black!70, anchor=north, yshift=-1.6pt, inner sep=0.5pt] at (axis cs:2,39.6) {39.6};
\node[font=\tiny, text=black!70, anchor=north west, yshift=-1.6pt, inner sep=0.5pt] at (axis cs:3,44.7) {44.7};
\node[font=\tiny, text=black!70, anchor=south, yshift=1.6pt, inner sep=0.5pt] at (axis cs:4,50.7) {50.7};
\node[font=\tiny, text=black!70, anchor=north, yshift=-1.6pt, inner sep=0.5pt] at (axis cs:5,46.7) {46.7};
\node[font=\tiny, text=black!70, anchor=south, yshift=1.6pt, inner sep=0.5pt] at (axis cs:6,49.2) {49.2};
\nextgroupplot[title={TriviaQA}, ymin=34, ymax=56, ytick={40,50}, ylabel={F1}, ylabel style={font=\scriptsize, yshift=-3pt}]
\addplot[name path=U, draw=none, forget plot] table[x=x, y expr=\thisrow{m}+\thisrow{sd}] {
x  m  sd
0  40.9  2.24
1  42.1  2.28
2  45.1  2.38
3  49.3  2.22
4  48.1  2.48
5  49.8  2.30
6  48.6  2.30
};
\addplot[name path=L, draw=none, forget plot] table[x=x, y expr=\thisrow{m}-\thisrow{sd}] {
x  m  sd
0  40.9  2.24
1  42.1  2.28
2  45.1  2.38
3  49.3  2.22
4  48.1  2.48
5  49.8  2.30
6  48.6  2.30
};
\addplot[lineO, fill opacity=0.15, draw=none, forget plot] fill between[of=U and L];
\addplot[lineO, line width=0.9pt, mark=*, mark size=1.4pt, mark options={fill=lineO, draw=white, line width=0.35pt}] table[x=x, y=m] {
x  m  sd
0  40.9  2.24
1  42.1  2.28
2  45.1  2.38
3  49.3  2.22
4  48.1  2.48
5  49.8  2.30
6  48.6  2.30
};
\node[font=\tiny, text=black!70, anchor=south, yshift=1.6pt, inner sep=0.5pt] at (axis cs:0,40.9) {40.9};
\node[font=\tiny, text=black!70, anchor=north, yshift=-1.6pt, inner sep=0.5pt] at (axis cs:1,42.1) {42.1};
\node[font=\tiny, text=black!70, anchor=north west, yshift=-1.6pt, inner sep=0.5pt] at (axis cs:2,45.1) {45.1};
\node[font=\tiny, text=black!70, anchor=south, yshift=1.6pt, inner sep=0.5pt] at (axis cs:3,49.3) {49.3};
\node[font=\tiny, text=black!70, anchor=north, yshift=-1.6pt, inner sep=0.5pt] at (axis cs:4,48.1) {48.1};
\node[font=\tiny, text=black!70, anchor=south, yshift=1.6pt, inner sep=0.5pt] at (axis cs:5,49.8) {49.8};
\node[font=\tiny, text=black!70, anchor=north, yshift=-1.6pt, inner sep=0.5pt] at (axis cs:6,48.6) {48.6};
\nextgroupplot[title={Qasper}, ymin=18, ymax=46, ytick={20,30,40}]
\addplot[name path=U, draw=none, forget plot] table[x=x, y expr=\thisrow{m}+\thisrow{sd}] {
x  m  sd
0  26.1  1.77
1  30.4  1.64
2  27.9  1.76
3  35.6  1.79
4  35.8  1.78
5  38.2  1.83
6  39.8  1.81
};
\addplot[name path=L, draw=none, forget plot] table[x=x, y expr=\thisrow{m}-\thisrow{sd}] {
x  m  sd
0  26.1  1.77
1  30.4  1.64
2  27.9  1.76
3  35.6  1.79
4  35.8  1.78
5  38.2  1.83
6  39.8  1.81
};
\addplot[lineO, fill opacity=0.15, draw=none, forget plot] fill between[of=U and L];
\addplot[lineO, line width=0.9pt, mark=*, mark size=1.4pt, mark options={fill=lineO, draw=white, line width=0.35pt}] table[x=x, y=m] {
x  m  sd
0  26.1  1.77
1  30.4  1.64
2  27.9  1.76
3  35.6  1.79
4  35.8  1.78
5  38.2  1.83
6  39.8  1.81
};
\node[font=\tiny, text=black!70, anchor=north, yshift=-1.6pt, inner sep=0.5pt] at (axis cs:0,26.1) {26.1};
\node[font=\tiny, text=black!70, anchor=south, yshift=1.6pt, inner sep=0.5pt] at (axis cs:1,30.4) {30.4};
\node[font=\tiny, text=black!70, anchor=north, yshift=-1.6pt, inner sep=0.5pt] at (axis cs:2,27.9) {27.9};
\node[font=\tiny, text=black!70, anchor=south, yshift=1.6pt, inner sep=0.5pt] at (axis cs:3,35.6) {35.6};
\node[font=\tiny, text=black!70, anchor=north, yshift=-1.6pt, inner sep=0.5pt] at (axis cs:4,35.8) {35.8};
\node[font=\tiny, text=black!70, anchor=south, yshift=1.6pt, inner sep=0.5pt] at (axis cs:5,38.2) {38.2};
\node[font=\tiny, text=black!70, anchor=north, yshift=-1.6pt, inner sep=0.5pt] at (axis cs:6,39.8) {39.8};
\nextgroupplot[title={MultiFieldQA}, ymin=8, ymax=30, ytick={10,20,30}]
\addplot[name path=U, draw=none, forget plot] table[x=x, y expr=\thisrow{m}+\thisrow{sd}] {
x  m  sd
0  19.9  2.97
1  18.8  2.67
2  15.8  2.77
3  22.4  2.87
4  20.9  2.58
5  22.4  2.78
6  22.1  2.96
};
\addplot[name path=L, draw=none, forget plot] table[x=x, y expr=\thisrow{m}-\thisrow{sd}] {
x  m  sd
0  19.9  2.97
1  18.8  2.67
2  15.8  2.77
3  22.4  2.87
4  20.9  2.58
5  22.4  2.78
6  22.1  2.96
};
\addplot[lineO, fill opacity=0.15, draw=none, forget plot] fill between[of=U and L];
\addplot[lineO, line width=0.9pt, mark=*, mark size=1.4pt, mark options={fill=lineO, draw=white, line width=0.35pt}] table[x=x, y=m] {
x  m  sd
0  19.9  2.97
1  18.8  2.67
2  15.8  2.77
3  22.4  2.87
4  20.9  2.58
5  22.4  2.78
6  22.1  2.96
};
\node[font=\tiny, text=black!70, anchor=south, yshift=1.6pt, inner sep=0.5pt] at (axis cs:0,19.9) {19.9};
\node[font=\tiny, text=black!70, anchor=north, yshift=-1.6pt, inner sep=0.5pt] at (axis cs:1,18.8) {18.8};
\node[font=\tiny, text=black!70, anchor=north, yshift=-1.6pt, inner sep=0.5pt] at (axis cs:2,15.8) {15.8};
\node[font=\tiny, text=black!70, anchor=south, yshift=1.6pt, inner sep=0.5pt] at (axis cs:3,22.4) {22.4};
\node[font=\tiny, text=black!70, anchor=north, yshift=-1.6pt, inner sep=0.5pt] at (axis cs:4,20.9) {20.9};
\node[font=\tiny, text=black!70, anchor=south, yshift=1.6pt, inner sep=0.5pt] at (axis cs:5,22.4) {22.4};
\node[font=\tiny, text=black!70, anchor=north, yshift=-1.6pt, inner sep=0.5pt] at (axis cs:6,22.1) {22.1};
\nextgroupplot[title={NarrativeQA}, ymin=8, ymax=24, ytick={10,15,20}]
\addplot[name path=U, draw=none, forget plot] table[x=x, y expr=\thisrow{m}+\thisrow{sd}] {
x  m  sd
0  13.0  1.28
1  13.4  1.07
2  12.9  1.17
3  17.9  1.19
4  18.3  1.22
5  18.6  1.29
6  19.4  1.26
};
\addplot[name path=L, draw=none, forget plot] table[x=x, y expr=\thisrow{m}-\thisrow{sd}] {
x  m  sd
0  13.0  1.28
1  13.4  1.07
2  12.9  1.17
3  17.9  1.19
4  18.3  1.22
5  18.6  1.29
6  19.4  1.26
};
\addplot[lineO, fill opacity=0.15, draw=none, forget plot] fill between[of=U and L];
\addplot[lineO, line width=0.9pt, mark=*, mark size=1.4pt, mark options={fill=lineO, draw=white, line width=0.35pt}] table[x=x, y=m] {
x  m  sd
0  13.0  1.28
1  13.4  1.07
2  12.9  1.17
3  17.9  1.19
4  18.3  1.22
5  18.6  1.29
6  19.4  1.26
};
\node[font=\tiny, text=black!70, anchor=north, yshift=-1.6pt, inner sep=0.5pt] at (axis cs:0,13.0) {13.0};
\node[font=\tiny, text=black!70, anchor=south, yshift=1.6pt, inner sep=0.5pt] at (axis cs:1,13.4) {13.4};
\node[font=\tiny, text=black!70, anchor=north, yshift=-1.6pt, inner sep=0.5pt] at (axis cs:2,12.9) {12.9};
\node[font=\tiny, text=black!70, anchor=south, yshift=1.6pt, inner sep=0.5pt] at (axis cs:3,17.9) {17.9};
\node[font=\tiny, text=black!70, anchor=north, yshift=-1.6pt, inner sep=0.5pt] at (axis cs:4,18.3) {18.3};
\node[font=\tiny, text=black!70, anchor=south, yshift=1.6pt, inner sep=0.5pt] at (axis cs:5,18.6) {18.6};
\node[font=\tiny, text=black!70, anchor=north, yshift=-1.6pt, inner sep=0.5pt] at (axis cs:6,19.4) {19.4};
\nextgroupplot[title={LoCoMo}, ymin=10, ymax=30, ytick={10,20,30}]
\addplot[name path=U, draw=none, forget plot] table[x=x, y expr=\thisrow{m}+\thisrow{sd}] {
x  m  sd
0  14.5  0.72
1  15.1  0.71
2  15.5  0.72
3  21.1  0.79
4  19.8  0.79
5  20.5  0.79
6  25.6  0.85
};
\addplot[name path=L, draw=none, forget plot] table[x=x, y expr=\thisrow{m}-\thisrow{sd}] {
x  m  sd
0  14.5  0.72
1  15.1  0.71
2  15.5  0.72
3  21.1  0.79
4  19.8  0.79
5  20.5  0.79
6  25.6  0.85
};
\addplot[lineO, fill opacity=0.15, draw=none, forget plot] fill between[of=U and L];
\addplot[lineO, line width=0.9pt, mark=*, mark size=1.4pt, mark options={fill=lineO, draw=white, line width=0.35pt}] table[x=x, y=m] {
x  m  sd
0  14.5  0.72
1  15.1  0.71
2  15.5  0.72
3  21.1  0.79
4  19.8  0.79
5  20.5  0.79
6  25.6  0.85
};
\node[font=\tiny, text=black!70, anchor=north, yshift=-1.6pt, inner sep=0.5pt] at (axis cs:0,14.5) {14.5};
\node[font=\tiny, text=black!70, anchor=south, yshift=1.6pt, inner sep=0.5pt] at (axis cs:1,15.1) {15.1};
\node[font=\tiny, text=black!70, anchor=north, yshift=-1.6pt, inner sep=0.5pt] at (axis cs:2,15.5) {15.5};
\node[font=\tiny, text=black!70, anchor=south, yshift=1.6pt, inner sep=0.5pt] at (axis cs:3,21.1) {21.1};
\node[font=\tiny, text=black!70, anchor=north, yshift=-1.6pt, inner sep=0.5pt] at (axis cs:4,19.8) {19.8};
\node[font=\tiny, text=black!70, anchor=south east, yshift=1.6pt, inner sep=0.5pt] at (axis cs:5,20.5) {20.5};
\node[font=\tiny, text=black!70, anchor=south, yshift=1.6pt, inner sep=0.5pt] at (axis cs:6,25.6) {25.6};
\end{groupplot}
\end{tikzpicture}
\caption{F1 of RoDeL-Seq with Gemma-4-E2B on each dataset after each update. HotpotQA and MuSiQue (ID) are in-domain and the others out-of-domain. Fr.: frozen roles (PRAG-style). Shaded bands show $\pm 1$ standard deviation of a test-set bootstrap.}
\label{fig:persetgemma}
\end{figure}

\begin{figure}[!htb]
\centering
\begin{tikzpicture}
\begin{groupplot}[
  group style={group size=5 by 2, horizontal sep=0.55cm, vertical sep=0.95cm},
  scale only axis, width=2.5cm, height=2.0cm,
  xmin=-0.5, xmax=3.5,
  xtick={0,...,3},
  xticklabels={Fr.,$u_1$,$u_2$,$u_3$},
  axis lines=left, axis line style={-, black!60}, tick style={black!60}, major tick length=2pt,
  tick label style={font=\tiny, inner sep=1pt},
  title style={font=\footnotesize, yshift=-4pt},
  ymajorgrids, grid style={black!22, dashed},
  clip=false,
]
\nextgroupplot[title={HotpotQA (ID)}, ymin=15, ymax=50, ytick={20,30,40,50}, ylabel={F1}, ylabel style={font=\scriptsize, yshift=-3pt}]
\addplot[name path=U, draw=none, forget plot] table[x=x, y expr=\thisrow{m}+\thisrow{sd}] {
x  m  sd
0  25.6  2.88
1  30.1  3.01
2  36.9  3.11
3  33.2  3.01
};
\addplot[name path=L, draw=none, forget plot] table[x=x, y expr=\thisrow{m}-\thisrow{sd}] {
x  m  sd
0  25.6  2.88
1  30.1  3.01
2  36.9  3.11
3  33.2  3.01
};
\addplot[lineO, fill opacity=0.15, draw=none, forget plot] fill between[of=U and L];
\addplot[lineO, line width=0.9pt, mark=*, mark size=1.4pt, mark options={fill=lineO, draw=white, line width=0.35pt}] table[x=x, y=m] {
x  m  sd
0  25.6  2.88
1  30.1  3.01
2  36.9  3.11
3  33.2  3.01
};
\node[font=\tiny, text=black!70, anchor=north, yshift=-1.6pt, inner sep=0.5pt] at (axis cs:0,25.6) {25.6};
\node[font=\tiny, text=black!70, anchor=north west, yshift=-1.6pt, inner sep=0.5pt] at (axis cs:1,30.1) {30.1};
\node[font=\tiny, text=black!70, anchor=south, yshift=1.6pt, inner sep=0.5pt] at (axis cs:2,36.9) {36.9};
\node[font=\tiny, text=black!70, anchor=north, yshift=-1.6pt, inner sep=0.5pt] at (axis cs:3,33.2) {33.2};
\nextgroupplot[title={MuSiQue (ID)}, ymin=0, ymax=22, ytick={0,10,20}]
\addplot[name path=U, draw=none, forget plot] table[x=x, y expr=\thisrow{m}+\thisrow{sd}] {
x  m  sd
0  6.7  1.36
1  8.7  1.65
2  12.9  1.88
3  15.0  2.02
};
\addplot[name path=L, draw=none, forget plot] table[x=x, y expr=\thisrow{m}-\thisrow{sd}] {
x  m  sd
0  6.7  1.36
1  8.7  1.65
2  12.9  1.88
3  15.0  2.02
};
\addplot[lineO, fill opacity=0.15, draw=none, forget plot] fill between[of=U and L];
\addplot[lineO, line width=0.9pt, mark=*, mark size=1.4pt, mark options={fill=lineO, draw=white, line width=0.35pt}] table[x=x, y=m] {
x  m  sd
0  6.7  1.36
1  8.7  1.65
2  12.9  1.88
3  15.0  2.02
};
\node[font=\tiny, text=black!70, anchor=north, yshift=-1.6pt, inner sep=0.5pt] at (axis cs:0,6.7) {6.7};
\node[font=\tiny, text=black!70, anchor=north west, yshift=-1.6pt, inner sep=0.5pt] at (axis cs:1,8.7) {8.7};
\node[font=\tiny, text=black!70, anchor=north west, yshift=-1.6pt, inner sep=0.5pt] at (axis cs:2,12.9) {12.9};
\node[font=\tiny, text=black!70, anchor=south, yshift=1.6pt, inner sep=0.5pt] at (axis cs:3,15.0) {15.0};
\nextgroupplot[title={SQuAD}, ymin=58, ymax=76, ytick={60,65,70,75}]
\addplot[name path=U, draw=none, forget plot] table[x=x, y expr=\thisrow{m}+\thisrow{sd}] {
x  m  sd
0  65.4  1.71
1  66.7  1.73
2  65.3  1.46
3  62.8  1.57
};
\addplot[name path=L, draw=none, forget plot] table[x=x, y expr=\thisrow{m}-\thisrow{sd}] {
x  m  sd
0  65.4  1.71
1  66.7  1.73
2  65.3  1.46
3  62.8  1.57
};
\addplot[lineO, fill opacity=0.15, draw=none, forget plot] fill between[of=U and L];
\addplot[lineO, line width=0.9pt, mark=*, mark size=1.4pt, mark options={fill=lineO, draw=white, line width=0.35pt}] table[x=x, y=m] {
x  m  sd
0  65.4  1.71
1  66.7  1.73
2  65.3  1.46
3  62.8  1.57
};
\node[font=\tiny, text=black!70, anchor=north, yshift=-1.6pt, inner sep=0.5pt] at (axis cs:0,65.4) {65.4};
\node[font=\tiny, text=black!70, anchor=south, yshift=1.6pt, inner sep=0.5pt] at (axis cs:1,66.7) {66.7};
\node[font=\tiny, text=black!70, anchor=south west, yshift=1.6pt, inner sep=0.5pt] at (axis cs:2,65.3) {65.3};
\node[font=\tiny, text=black!70, anchor=north, yshift=-1.6pt, inner sep=0.5pt] at (axis cs:3,62.8) {62.8};
\nextgroupplot[title={2Wiki}, ymin=24, ymax=52, ytick={30,40,50}]
\addplot[name path=U, draw=none, forget plot] table[x=x, y expr=\thisrow{m}+\thisrow{sd}] {
x  m  sd
0  32.9  3.18
1  38.4  3.28
2  46.7  3.30
3  46.5  3.33
};
\addplot[name path=L, draw=none, forget plot] table[x=x, y expr=\thisrow{m}-\thisrow{sd}] {
x  m  sd
0  32.9  3.18
1  38.4  3.28
2  46.7  3.30
3  46.5  3.33
};
\addplot[lineO, fill opacity=0.15, draw=none, forget plot] fill between[of=U and L];
\addplot[lineO, line width=0.9pt, mark=*, mark size=1.4pt, mark options={fill=lineO, draw=white, line width=0.35pt}] table[x=x, y=m] {
x  m  sd
0  32.9  3.18
1  38.4  3.28
2  46.7  3.30
3  46.5  3.33
};
\node[font=\tiny, text=black!70, anchor=north, yshift=-1.6pt, inner sep=0.5pt] at (axis cs:0,32.9) {32.9};
\node[font=\tiny, text=black!70, anchor=north west, yshift=-1.6pt, inner sep=0.5pt] at (axis cs:1,38.4) {38.4};
\node[font=\tiny, text=black!70, anchor=south, yshift=1.6pt, inner sep=0.5pt] at (axis cs:2,46.7) {46.7};
\node[font=\tiny, text=black!70, anchor=north, yshift=-1.6pt, inner sep=0.5pt] at (axis cs:3,46.5) {46.5};
\nextgroupplot[title={NQ}, ymin=30, ymax=58, ytick={30,40,50}]
\addplot[name path=U, draw=none, forget plot] table[x=x, y expr=\thisrow{m}+\thisrow{sd}] {
x  m  sd
0  38.8  2.91
1  41.9  2.94
2  51.4  2.74
3  49.9  2.87
};
\addplot[name path=L, draw=none, forget plot] table[x=x, y expr=\thisrow{m}-\thisrow{sd}] {
x  m  sd
0  38.8  2.91
1  41.9  2.94
2  51.4  2.74
3  49.9  2.87
};
\addplot[lineO, fill opacity=0.15, draw=none, forget plot] fill between[of=U and L];
\addplot[lineO, line width=0.9pt, mark=*, mark size=1.4pt, mark options={fill=lineO, draw=white, line width=0.35pt}] table[x=x, y=m] {
x  m  sd
0  38.8  2.91
1  41.9  2.94
2  51.4  2.74
3  49.9  2.87
};
\node[font=\tiny, text=black!70, anchor=north, yshift=-1.6pt, inner sep=0.5pt] at (axis cs:0,38.8) {38.8};
\node[font=\tiny, text=black!70, anchor=north west, yshift=-1.6pt, inner sep=0.5pt] at (axis cs:1,41.9) {41.9};
\node[font=\tiny, text=black!70, anchor=south, yshift=1.6pt, inner sep=0.5pt] at (axis cs:2,51.4) {51.4};
\node[font=\tiny, text=black!70, anchor=north, yshift=-1.6pt, inner sep=0.5pt] at (axis cs:3,49.9) {49.9};
\nextgroupplot[title={TriviaQA}, ymin=34, ymax=56, ytick={40,50}, ylabel={F1}, ylabel style={font=\scriptsize, yshift=-3pt}]
\addplot[name path=U, draw=none, forget plot] table[x=x, y expr=\thisrow{m}+\thisrow{sd}] {
x  m  sd
0  40.9  2.24
1  43.7  2.27
2  52.2  2.13
3  50.4  2.30
};
\addplot[name path=L, draw=none, forget plot] table[x=x, y expr=\thisrow{m}-\thisrow{sd}] {
x  m  sd
0  40.9  2.24
1  43.7  2.27
2  52.2  2.13
3  50.4  2.30
};
\addplot[lineO, fill opacity=0.15, draw=none, forget plot] fill between[of=U and L];
\addplot[lineO, line width=0.9pt, mark=*, mark size=1.4pt, mark options={fill=lineO, draw=white, line width=0.35pt}] table[x=x, y=m] {
x  m  sd
0  40.9  2.24
1  43.7  2.27
2  52.2  2.13
3  50.4  2.30
};
\node[font=\tiny, text=black!70, anchor=north, yshift=-1.6pt, inner sep=0.5pt] at (axis cs:0,40.9) {40.9};
\node[font=\tiny, text=black!70, anchor=north west, yshift=-1.6pt, inner sep=0.5pt] at (axis cs:1,43.7) {43.7};
\node[font=\tiny, text=black!70, anchor=south, yshift=1.6pt, inner sep=0.5pt] at (axis cs:2,52.2) {52.2};
\node[font=\tiny, text=black!70, anchor=north, yshift=-1.6pt, inner sep=0.5pt] at (axis cs:3,50.4) {50.4};
\nextgroupplot[title={Qasper}, ymin=18, ymax=46, ytick={20,30,40}]
\addplot[name path=U, draw=none, forget plot] table[x=x, y expr=\thisrow{m}+\thisrow{sd}] {
x  m  sd
0  26.1  1.77
1  30.5  1.63
2  41.0  1.84
3  40.2  1.82
};
\addplot[name path=L, draw=none, forget plot] table[x=x, y expr=\thisrow{m}-\thisrow{sd}] {
x  m  sd
0  26.1  1.77
1  30.5  1.63
2  41.0  1.84
3  40.2  1.82
};
\addplot[lineO, fill opacity=0.15, draw=none, forget plot] fill between[of=U and L];
\addplot[lineO, line width=0.9pt, mark=*, mark size=1.4pt, mark options={fill=lineO, draw=white, line width=0.35pt}] table[x=x, y=m] {
x  m  sd
0  26.1  1.77
1  30.5  1.63
2  41.0  1.84
3  40.2  1.82
};
\node[font=\tiny, text=black!70, anchor=north, yshift=-1.6pt, inner sep=0.5pt] at (axis cs:0,26.1) {26.1};
\node[font=\tiny, text=black!70, anchor=north west, yshift=-1.6pt, inner sep=0.5pt] at (axis cs:1,30.5) {30.5};
\node[font=\tiny, text=black!70, anchor=south, yshift=1.6pt, inner sep=0.5pt] at (axis cs:2,41.0) {41.0};
\node[font=\tiny, text=black!70, anchor=north, yshift=-1.6pt, inner sep=0.5pt] at (axis cs:3,40.2) {40.2};
\nextgroupplot[title={MultiFieldQA}, ymin=8, ymax=30, ytick={10,20,30}]
\addplot[name path=U, draw=none, forget plot] table[x=x, y expr=\thisrow{m}+\thisrow{sd}] {
x  m  sd
0  19.9  2.97
1  18.4  2.93
2  23.4  3.05
3  25.2  3.05
};
\addplot[name path=L, draw=none, forget plot] table[x=x, y expr=\thisrow{m}-\thisrow{sd}] {
x  m  sd
0  19.9  2.97
1  18.4  2.93
2  23.4  3.05
3  25.2  3.05
};
\addplot[lineO, fill opacity=0.15, draw=none, forget plot] fill between[of=U and L];
\addplot[lineO, line width=0.9pt, mark=*, mark size=1.4pt, mark options={fill=lineO, draw=white, line width=0.35pt}] table[x=x, y=m] {
x  m  sd
0  19.9  2.97
1  18.4  2.93
2  23.4  3.05
3  25.2  3.05
};
\node[font=\tiny, text=black!70, anchor=south, yshift=1.6pt, inner sep=0.5pt] at (axis cs:0,19.9) {19.9};
\node[font=\tiny, text=black!70, anchor=north, yshift=-1.6pt, inner sep=0.5pt] at (axis cs:1,18.4) {18.4};
\node[font=\tiny, text=black!70, anchor=north west, yshift=-1.6pt, inner sep=0.5pt] at (axis cs:2,23.4) {23.4};
\node[font=\tiny, text=black!70, anchor=south, yshift=1.6pt, inner sep=0.5pt] at (axis cs:3,25.2) {25.2};
\nextgroupplot[title={NarrativeQA}, ymin=8, ymax=24, ytick={10,15,20}]
\addplot[name path=U, draw=none, forget plot] table[x=x, y expr=\thisrow{m}+\thisrow{sd}] {
x  m  sd
0  13.0  1.28
1  15.4  1.29
2  20.5  1.14
3  21.1  1.24
};
\addplot[name path=L, draw=none, forget plot] table[x=x, y expr=\thisrow{m}-\thisrow{sd}] {
x  m  sd
0  13.0  1.28
1  15.4  1.29
2  20.5  1.14
3  21.1  1.24
};
\addplot[lineO, fill opacity=0.15, draw=none, forget plot] fill between[of=U and L];
\addplot[lineO, line width=0.9pt, mark=*, mark size=1.4pt, mark options={fill=lineO, draw=white, line width=0.35pt}] table[x=x, y=m] {
x  m  sd
0  13.0  1.28
1  15.4  1.29
2  20.5  1.14
3  21.1  1.24
};
\node[font=\tiny, text=black!70, anchor=north, yshift=-1.6pt, inner sep=0.5pt] at (axis cs:0,13.0) {13.0};
\node[font=\tiny, text=black!70, anchor=north west, yshift=-1.6pt, inner sep=0.5pt] at (axis cs:1,15.4) {15.4};
\node[font=\tiny, text=black!70, anchor=north west, yshift=-1.6pt, inner sep=0.5pt] at (axis cs:2,20.5) {20.5};
\node[font=\tiny, text=black!70, anchor=south, yshift=1.6pt, inner sep=0.5pt] at (axis cs:3,21.1) {21.1};
\nextgroupplot[title={LoCoMo}, ymin=10, ymax=30, ytick={10,20,30}]
\addplot[name path=U, draw=none, forget plot] table[x=x, y expr=\thisrow{m}+\thisrow{sd}] {
x  m  sd
0  14.5  0.72
1  18.1  0.79
2  26.5  0.85
3  27.9  0.87
};
\addplot[name path=L, draw=none, forget plot] table[x=x, y expr=\thisrow{m}-\thisrow{sd}] {
x  m  sd
0  14.5  0.72
1  18.1  0.79
2  26.5  0.85
3  27.9  0.87
};
\addplot[lineO, fill opacity=0.15, draw=none, forget plot] fill between[of=U and L];
\addplot[lineO, line width=0.9pt, mark=*, mark size=1.4pt, mark options={fill=lineO, draw=white, line width=0.35pt}] table[x=x, y=m] {
x  m  sd
0  14.5  0.72
1  18.1  0.79
2  26.5  0.85
3  27.9  0.87
};
\node[font=\tiny, text=black!70, anchor=north, yshift=-1.6pt, inner sep=0.5pt] at (axis cs:0,14.5) {14.5};
\node[font=\tiny, text=black!70, anchor=north west, yshift=-1.6pt, inner sep=0.5pt] at (axis cs:1,18.1) {18.1};
\node[font=\tiny, text=black!70, anchor=north west, yshift=-1.6pt, inner sep=0.5pt] at (axis cs:2,26.5) {26.5};
\node[font=\tiny, text=black!70, anchor=south, yshift=1.6pt, inner sep=0.5pt] at (axis cs:3,27.9) {27.9};
\end{groupplot}
\end{tikzpicture}
\caption{F1 of RoDeL-Sim with Gemma-4-E2B on each dataset after each simultaneous update. Datasets, Fr., and shaded bands are as in \Cref{fig:persetgemma}.}
\label{fig:persetgemmasim}
\end{figure}
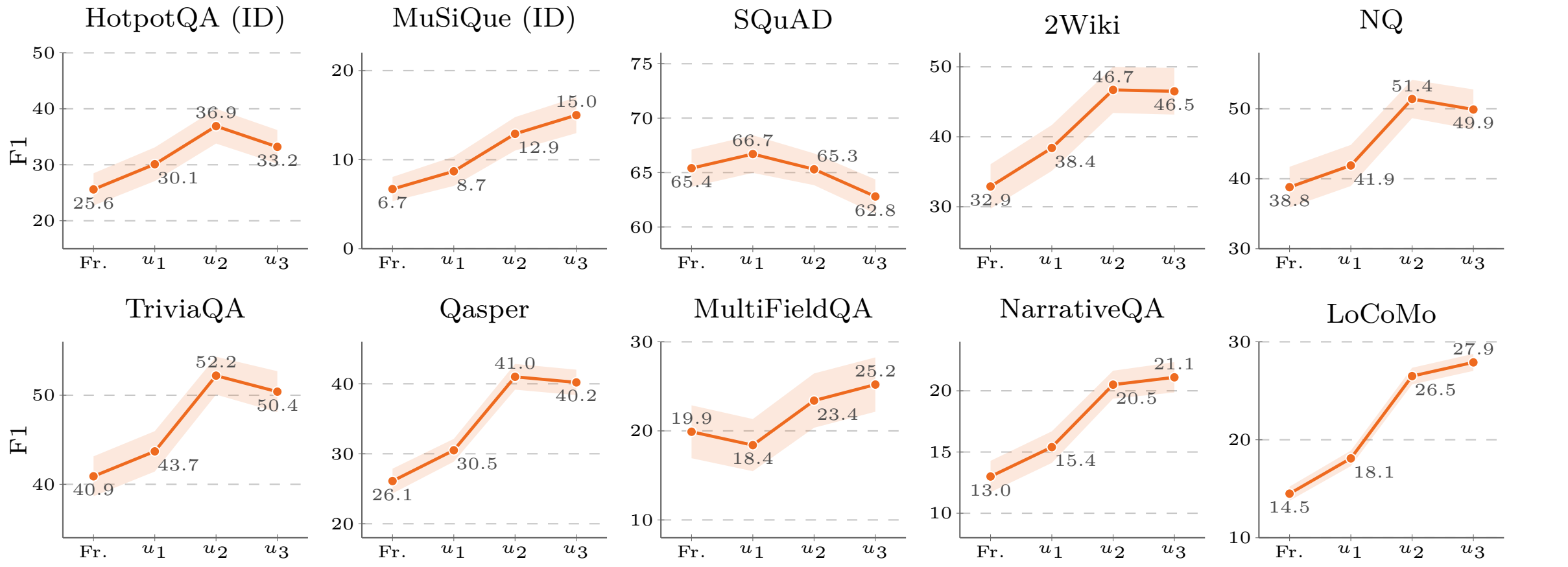

Both variants are also trained with the instruction-tuned Gemma-4-E2B~\citep{gemmateam2026gemma4technicalreport} as the base model, using its text model only. The training documents, synthetic questions, and three rounds of updates are those of the main experiments, and evaluation and significance tests follow \Cref{app:results}. \Cref{fig:avggemma,fig:persetgemma} show the F1 of RoDeL-Seq after each update, on average and on each dataset. \Cref{fig:avggemmasim,fig:persetgemmasim} show the same for RoDeL-Sim.

\begin{figure}[!htb]
\begin{minipage}[t]{\colw}
\centering
\pgfplotstableread{
x  id    idsd  ood   oodsd
0  16.1  1.59  31.4  0.80
1  18.4  1.72  32.4  0.77
2  19.3  1.73  32.5  0.80
3  24.6  1.89  37.7  0.80
4  23.6  1.84  37.3  0.78
5  24.9  1.86  38.3  0.79
6  27.0  1.88  39.6  0.79
}\dyntab
\begin{tikzpicture}
\begin{axis}[
  width=\colw, height=5.2cm,
  xmin=-0.5, xmax=6.5, ymin=8, ymax=51,
  xtick={0,...,6},
  xticklabels={Frozen,\textcolor{stuV}{$s_1$},\textcolor{tchV}{$t_1$},\textcolor{stuV}{$s_2$},\textcolor{tchV}{$t_2$},\textcolor{stuV}{$s_3$},\textcolor{tchV}{$t_3$}},
  ytick={10,20,30,40},
  ylabel={Average F1},
  axis lines=left, axis line style={-, black!60}, tick style={black!60},
  tick label style={font=\small}, label style={font=\small},
  ymajorgrids, grid style={black!22, dashed},
  set layers, clip=false,
  legend style={font=\small, at={(0.5,1.03)}, anchor=south, draw=none, fill=none, legend columns=2, /tikz/every even column/.append style={column sep=10pt}},
  legend cell align=left,
]
\begin{pgfonlayer}{axis background}
\draw[black!25] (axis cs:0.5,8) -- (axis cs:0.5,51);
\draw[black!25] (axis cs:2.5,8) -- (axis cs:2.5,51);
\draw[black!25] (axis cs:4.5,8) -- (axis cs:4.5,51);
\end{pgfonlayer}
\node[font=\scriptsize, text=black!55, anchor=north] at (axis cs:1.5,51) {Round 1};
\node[font=\scriptsize, text=black!55, anchor=north] at (axis cs:3.5,51) {Round 2};
\node[font=\scriptsize, text=black!55, anchor=north] at (axis cs:5.5,51) {Round 3};
\addplot[name path=idU, draw=none, forget plot] table[x=x, y expr=\thisrow{id}+\thisrow{idsd}] {\dyntab};
\addplot[name path=idL, draw=none, forget plot] table[x=x, y expr=\thisrow{id}-\thisrow{idsd}] {\dyntab};
\addplot[lineO, fill opacity=0.15, draw=none, forget plot] fill between[of=idU and idL];
\addplot[name path=oodU, draw=none, forget plot] table[x=x, y expr=\thisrow{ood}+\thisrow{oodsd}] {\dyntab};
\addplot[name path=oodL, draw=none, forget plot] table[x=x, y expr=\thisrow{ood}-\thisrow{oodsd}] {\dyntab};
\addplot[lineO, fill opacity=0.15, draw=none, forget plot] fill between[of=oodU and oodL];
\addplot[lineO, line width=1.2pt, mark=*, mark size=2.3pt, mark options={fill=lineO, draw=white, line width=0.6pt},
  nodes near coords, point meta=explicit symbolic, every node near coord/.append style={font=\scriptsize, text=black!70, anchor=north, yshift=-4pt}]
  table[x=x, y=id, meta=id] {\dyntab};
\addlegendentry{In-domain}
\addplot[lineO, line width=1.2pt, mark=square*, mark size=2.1pt, mark options={fill=lineO, draw=white, line width=0.6pt},
  nodes near coords, point meta=explicit symbolic, every node near coord/.append style={font=\scriptsize, text=black!70, anchor=south, yshift=2pt}]
  table[x=x, y=ood, meta=ood] {\dyntab};
\addlegendentry{Out-of-domain}
\end{axis}
\end{tikzpicture}
\caption{Average F1 of RoDeL-Seq with Gemma-4-E2B after each update. Shaded bands show $\pm 1$ standard deviation of a test-set bootstrap.}
\label{fig:avggemma}
\end{minipage}\hfill
\begin{minipage}[t]{\colw}
\centering
\pgfplotstableread{
x  id    idsd  ood   oodsd
0  16.1  1.59  31.4  0.80
1  19.4  1.73  34.2  0.80
2  24.9  1.82  40.9  0.78
3  24.1  1.81  40.5  0.81
}\simtab
\begin{tikzpicture}
\begin{axis}[
  width=\colw, height=5.2cm,
  xmin=-0.5, xmax=3.5, ymin=8, ymax=51,
  xtick={0,...,3},
  xticklabels={Frozen,$u_1$,$u_2$,$u_3$},
  ytick={10,20,30,40},
  ylabel={Average F1},
  axis lines=left, axis line style={-, black!60}, tick style={black!60},
  tick label style={font=\small}, label style={font=\small},
  ymajorgrids, grid style={black!22, dashed},
  clip=false,
  legend style={font=\small, at={(0.5,1.03)}, anchor=south, draw=none, fill=none, legend columns=2, /tikz/every even column/.append style={column sep=10pt}},
  legend cell align=left,
]
\addplot[name path=idU, draw=none, forget plot] table[x=x, y expr=\thisrow{id}+\thisrow{idsd}] {\simtab};
\addplot[name path=idL, draw=none, forget plot] table[x=x, y expr=\thisrow{id}-\thisrow{idsd}] {\simtab};
\addplot[lineO, fill opacity=0.15, draw=none, forget plot] fill between[of=idU and idL];
\addplot[name path=oodU, draw=none, forget plot] table[x=x, y expr=\thisrow{ood}+\thisrow{oodsd}] {\simtab};
\addplot[name path=oodL, draw=none, forget plot] table[x=x, y expr=\thisrow{ood}-\thisrow{oodsd}] {\simtab};
\addplot[lineO, fill opacity=0.15, draw=none, forget plot] fill between[of=oodU and oodL];
\addplot[lineO, line width=1.2pt, mark=*, mark size=2.3pt, mark options={fill=lineO, draw=white, line width=0.6pt},
  nodes near coords, point meta=explicit symbolic, every node near coord/.append style={font=\scriptsize, text=black!70, anchor=north, yshift=-4pt}]
  table[x=x, y=id, meta=id] {\simtab};
\addlegendentry{In-domain}
\addplot[lineO, line width=1.2pt, mark=square*, mark size=2.1pt, mark options={fill=lineO, draw=white, line width=0.6pt},
  nodes near coords, point meta=explicit symbolic, every node near coord/.append style={font=\scriptsize, text=black!70, anchor=south, yshift=2pt}]
  table[x=x, y=ood, meta=ood] {\simtab};
\addlegendentry{Out-of-domain}
\end{axis}
\end{tikzpicture}
\caption{Average F1 of RoDeL-Sim with Gemma-4-E2B after each simultaneous update. Shaded bands are as in \Cref{fig:avggemma}.}
\label{fig:avggemmasim}
\end{minipage}
\end{figure}

\paragraph{Overall gains.} After $t_3$, RoDeL-Seq scores significantly higher than the frozen roles on both averages and, after Holm correction, on eight of the ten datasets. The in-domain average rises from 16.1 to 27.0, and the out-of-domain average from 31.4 to 39.6. After $u_3$, RoDeL-Sim also scores significantly higher than the frozen roles on both averages: its in-domain average rises from 16.1 to 24.1, and its out-of-domain average from 31.4 to 40.5. Its F1 is higher than that of the frozen roles on nine of the ten datasets, and on SQuAD it falls from 65.4 to 62.8.

\paragraph{Comparison against baselines.} \Cref{tab:main_gemma} compares the two variants with the three baselines trained with Gemma-4-E2B, each at its best point. With RoDeL-Seq, the end of training is the best point on both averages. With RoDeL-Sim, the best point on both averages is after $u_2$ (24.9 in domain and 40.9 out of domain), and $u_3$ changes neither average significantly. RoDeL-Seq has the highest average F1 in domain (27.0), and both variants have a higher out-of-domain average (39.6 and 40.9) than every baseline. The leads of RoDeL-Seq over PRAG-style and PaST-style on both averages have $p<0.05$. For RoDeL-Sim, the out-of-domain leads over all three baselines and the in-domain lead over PRAG-style have $p<0.05$.

\begin{table}[!ht]
\centering
\footnotesize
\setlength{\tabcolsep}{3.2pt}
\caption{F1 of each method at its best point with Gemma-4-E2B. For each method, the same checkpoint has the highest in-domain and the highest out-of-domain average, and all columns use it. The two numbers below each baseline score are its differences from RoDeL-Seq and from RoDeL-Sim, in red when the variant is higher and in green when the baseline is higher; markers as in \Cref{tab:main}. The rows on a gray background are the references of \Cref{tab:main}, with Gemma-4-E2B. Bold: best in each column among the five methods.}
\label{tab:main_gemma}
\begin{tabular}{@{}l rr!{\hspace{3pt}}r@{\hspace{10pt}} rr!{\hspace{3pt}}rrrrrr!{\hspace{3pt}}r@{\hspace{6pt}}}
\toprule
 & \multicolumn{3}{c}{In-domain} & \multicolumn{9}{c}{Out-of-domain} \\
\cmidrule(lr){2-4}\cmidrule(l){5-13}
Method & HotpotQA & MuSiQue & Avg. & SQuAD & 2Wiki & NQ & TriviaQA & Qasper & MFQA & NarQA & LoCoMo & Avg. \\
\midrule
\refrows{2}
No memory & 21.8 & 5.8 & 13.8 & 16.4 & 26.2 & 24.1 & 49.5 & 11.5 & 8.6 & 3.3 & 1.6 & 17.7 \\
In-context & 73.1 & 41.1 & 57.1 & 89.4 & 76.5 & 69.1 & 59.6 & 57.7 & 58.3 & 30.3 & 48.3 & 61.1 \\
\midrule
\textit{PRAG-style} & 25.6 & 6.7 & 16.1 & 65.4 & 32.9 & 38.8 & 40.9 & 26.1 & 19.9 & 13.0 & 14.5 & 31.4 \\[-1.5pt]
\vsrow{Seq} & \dlt{13.4}\dmk{{*\dagger}} & \dlt{8.2}\dmk{{*\dagger}} & \dlt{10.9}\dmk{*} & \dlt{2.6} & \dlt{11.2}\dmk{{*\dagger}} & \dlt{10.4}\dmk{{*\dagger}} & \dlt{7.7}\dmk{{*\dagger}} & \dlt{13.7}\dmk{{*\dagger}} & \dlt{2.2} & \dlt{6.4}\dmk{{*\dagger}} & \dlt{11.1}\dmk{{*\dagger}} & \dlt{8.2}\dmk{*} \\[-1.5pt]
\vsrow{Sim} & \dlt{11.3}\dmk{{*\dagger}} & \dlt{6.2}\dmk{{*\dagger}} & \dlt{8.8}\dmk{*} & \dltp{0.1} & \dlt{13.8}\dmk{{*\dagger}} & \dlt{12.6}\dmk{{*\dagger}} & \dlt{11.3}\dmk{{*\dagger}} & \dlt{14.9}\dmk{{*\dagger}} & \dlt{3.5} & \dlt{7.5}\dmk{{*\dagger}} & \dlt{12.0}\dmk{{*\dagger}} & \dlt{9.5}\dmk{*} \\[2pt]
\textit{SEAL-style} & 38.3 & 12.8 & 25.6 & 58.0 & 41.8 & 48.0 & 46.9 & 37.0 & \textbf{25.6} & 18.1 & \textbf{32.6} & 38.5 \\[-1.5pt]
\vsrow{Seq} & \dlt{0.7} & \dlt{2.1} & \dlt{1.4} & \dlt{10.0}\dmk{{*\dagger}} & \dlt{2.3} & \dlt{1.2} & \dlt{1.7} & \dlt{2.8}\dmk{{*\dagger}} & \dltp{3.5} & \dlt{1.3} & \dltp{7.0}\dmk{*} & \dlt{1.1} \\[-1.5pt]
\vsrow{Sim} & \dltp{1.4} & \dlt{0.1} & \dltp{0.7} & \dlt{7.3}\dmk{{*\dagger}} & \dlt{4.9}\dmk{*} & \dlt{3.4} & \dlt{5.3}\dmk{*} & \dlt{4.0}\dmk{{*\dagger}} & \dltp{2.2} & \dlt{2.4} & \dltp{6.1}\dmk{*} & \dlt{2.4}\dmk{*} \\[2pt]
\textit{PaST-style} & 35.8 & 8.5 & 22.2 & 62.9 & 40.2 & 46.7 & 49.2 & 34.9 & 23.4 & 18.3 & 21.9 & 37.2 \\[-1.5pt]
\vsrow{Seq} & \dlt{3.2} & \dlt{6.4}\dmk{*} & \dlt{4.8}\dmk{*} & \dlt{5.1}\dmk{{*\dagger}} & \dlt{3.9} & \dlt{2.5} & \dltp{0.6} & \dlt{4.9}\dmk{{*\dagger}} & \dltp{1.3} & \dlt{1.1} & \dlt{3.7}\dmk{{*\dagger}} & \dlt{2.4}\dmk{*} \\[-1.5pt]
\vsrow{Sim} & \dlt{1.1} & \dlt{4.4}\dmk{*} & \dlt{2.7} & \dlt{2.4} & \dlt{6.5}\dmk{{*\dagger}} & \dlt{4.7}\dmk{*} & \dlt{3.0} & \dlt{6.1}\dmk{{*\dagger}} & {\scriptsize\textcolor{black!55}{0.0}} & \dlt{2.2} & \dlt{4.6}\dmk{{*\dagger}} & \dlt{3.7}\dmk{*} \\
\midrule
RoDeL-Seq & \textbf{39.0} & \textbf{14.9} & \textbf{27.0} & \textbf{68.0} & 44.1 & 49.2 & 48.6 & 39.8 & 22.1 & 19.4 & 25.6 & 39.6 \\
RoDeL-Sim & 36.9 & 12.9 & 24.9 & 65.3 & \textbf{46.7} & \textbf{51.4} & \textbf{52.2} & \textbf{41.0} & 23.4 & \textbf{20.5} & 26.5 & \textbf{40.9} \\
\bottomrule
\end{tabular}
\end{table}

\end{document}